%% file: thesis.tex
\documentclass[12pt]{ociamthesis}  

\usepackage{times} 
\usepackage{amsmath,amssymb,amsthm} 
\usepackage[round]{natbib} 
\usepackage{multirow} 
\usepackage{graphicx} 
\usepackage{verbatim} 
\usepackage{pifont} 
\usepackage{setspace} 
\usepackage{titlesec} 
\usepackage{afterpage} 
\usepackage{cancel} 
\usepackage{appendix} 
\usepackage{subcaption}
\usepackage{pgfplots}

\usepackage{latexsym}
\usepackage{epstopdf}
\usepackage{longtable}
\usepackage{tabularx}
\usepackage{amsfonts}
\usepackage{enumerate}
\usepackage{mathtools}
\usepackage{tikz}
\usepackage{float}
\usepackage{fancyvrb}
\usepackage{pdflscape}
\usepackage{multirow}
\usepackage{enumitem}
\usepackage{booktabs}
\usepackage{algorithm}
\usepackage{algpseudocode}
\usepackage{xr}
\usepackage{hyperref}
\usepackage{url}
\usepackage{eucal,amsbsy}
\usepackage{CJKutf8}
\usetikzlibrary{positioning, fit, arrows.meta}

\usepackage{bibentry}
\nobibliography*

\pgfplotsset{compat=1.9}

\newcommand{\Break}{\State \textbf{break} }

\let\svthefootnote\thefootnote

\include{mymacros}

\title{Tackling Sequence to Sequence Mapping Problems with Neural Networks}   							

\author{Lei Yu}             
\college{Mansfield College}  

\degree{Doctor of Philosophy}     	
\degreedate{Trinity 2017}		

\DeclareMathOperator{\ReLU}{ReLU}
\DeclareMathOperator{\softmax}{softmax}
\DeclareMathOperator*{\topk}{topk}
\DeclareMathOperator*{\argtopk}{arg\,topk}
\DeclareMathOperator*{\candidate}{getCandidateOutputs}
\newcommand*\colvec[3][]{
	\left[\!
	\begin{array}{c}
		f_a\\
		f_b\\
	\end{array}
	\!\right]
}

\begin{document}

\setcounter{secnumdepth}{3}
\setcounter{tocdepth}{3}

\maketitle                  
\include{dedication}        
\include{acknowledgements}  
\include{abstract}          

\begin{romanpages}          
\tableofcontents            
\listoffigures              
\listoftables              
\end{romanpages}            

\baselineskip=20pt plus1pt

\include{introduction}
\include{chapter_3}

\include{chapter_4}

\include{chapter_5}
\include{chapter_6}
\include{conclusion}

\appendix
\include{appendix_a}

\addcontentsline{toc}{chapter}{Bibliography}

\bibliography{refs}        
\bibliographystyle{plainnat}  

\end{document}

%% file: mymacros.tex
\newcommand{\argmax}{\operatornamewithlimits{argmax}}

\titleformat{\subsubsection}[runin]{\bfseries}{}{}{}[]

\makeatletter
\newenvironment{chapterabstract}{%
    \begin{center}%
      \bfseries Chapter Abstract
    \end{center}\begin{quote}\begin{itshape}}%
   {\end{itshape}\end{quote}\begin{center}%
      \par%
      \noindent\rule[0.5ex]{0.75\linewidth}{1pt}%
     \end{center}}
\makeatother


%% file: dedication.tex
\begin{dedication}
	{\it To my parents Shijie and Lijuan}
\end{dedication}

%% file: acknowledgements.tex
\begin{acknowledgements}
First and foremost, I am deeply grateful to my supervisors, Prof. Stephen Pulman and Prof. Phil Blunsom for their seemingly immense knowledge and extraordinary kindness. Without their guidance and advice throughout my studies, this thesis would have never been accomplished. Stephen brought me to the exciting field of Natural Language Processing (NLP). 
The door to Stephen's office was always open whenever I had trouble in research or personal life. Phil is the key person who guided me to apply statistical machine learning to NLP and led me working on various exciting projects. He not only patiently taught me new things, but also shaped
the way I think in a fundamental way. I was fortunate to have worked with him. I thank my examiners Prof. Shimon Whiteson and Prof. Kyunghyun Cho for proofreading my thesis and for their suggestions on how to improve the presentation of the thesis.  Additionally, I would like to express my sincere gratitude to Prof. Chris Dyer, my manager at DeepMind. I have greatly benefited from his scientific insight and have been influenced by his dedication to work and enthusiasm to science.

I thank my colleagues at the NLP group at Oxford for the sleepless nights we were working together before deadlines, and for all the fun we have had in the last four years: Jan Buys, Pengyu Wang, Yishu Miao, Karl Moritz Hermann, Tom\'{a}\v{s} Ko\v{c}isk\'{y}, Edward Grefenstette,  Nal Kalchbrenner, and 
Paul Baltescu. I thank my colleagues at the language team at DeepMind for their fruitful discussions of my projects: G\'{a}bor Melis, Wang Ling, Lingpeng Kong, Dani Yogatama, and Stephen Clark.

Last but not the least, I would like to thank my family: my parents and to my sister for their love and encouragement. Most importantly, I am grateful to Wenda Li, whose faithful support helped me pass through all the hard times during my PhD. 




\end{acknowledgements}

%% file: abstract.tex
\begin{abstract}

In Natural Language Processing (NLP), it is important to detect the relationship between two sequences or to generate a sequence of tokens given another observed sequence. We call the type of problems on modelling sequence pairs as sequence to sequence (seq2seq) mapping problems.  A lot of research has been devoted to finding ways of tackling these problems, with traditional approaches relying on a combination of hand-crafted features, alignment models, segmentation heuristics, and external linguistic resources. Although great progress has been made, these traditional approaches suffer from various drawbacks, such as complicated pipeline, laborious feature engineering, and the difficulty for domain adaptation. Recently, neural networks emerged as
a promising solution to many problems in NLP, speech recognition, and computer vision. Neural models are powerful because they can be trained end to end, generalise well to unseen examples, and the same framework can be easily adapted to a new domain.

The aim of this thesis is to advance the state-of-the-art in seq2seq mapping problems with neural networks.  We explore solutions from three major aspects: investigating neural models for representing sequences, modelling interactions between sequences, and using unpaired data to boost the performance of neural models. For each aspect, we propose novel models and evaluate their efficacy on various tasks of seq2seq mapping.

Chapter \ref{ch:nn} covers the relevant literature on neural networks. Following
this, in Chapter \ref{ch:sentence_model} we explore the usefulness of distributed sentence models in seq2seq mapping problems by testing them in the task of answer sentence selection. We also empirically compare the performance of distributed sentence models based on different types of neural networks. Chapter \ref{ch:ssnt} presents a neural sequence transduction model that learns to alternate between encoding and decoding segments of the input as it is read. The model not only outperforms the encoder-decoder model significantly on various tasks such as machine translation and sentence summarisation, but also is capable of predicting outputs online during decoding. In Chapter \ref{ch:noisy_channel}, we propose to incorporate abundant unpaired data using the noisy channel model---with the component models parameterised by recurrent neural networks---and present a tractable and effective beam search decoder. 

\end{abstract}

%% file: introduction.tex
\chapter{Introduction}
\label{ch:intro}

The problem of mapping from one sequence to another is an important challenge of natural language processing (NLP). Common applications include the recognition tasks of paraphrase detection, textual entailment, and question answer selection; and the generation tasks of machine translation and text summarisation. The solution to the problem lies in the way to represent sequences properly and the ability to robustly process the interactions between sequences.

Traditionally this type of problem has been tackled by a combination of hand-crafted features, alignment models, segmentation heuristics, and external linguistic resources \citep{,burchardt2007semantic,malakasiotis2007learning,kozareva2006paraphrase,yih2013question}. For example, to determine whether the sentence ``{\it Hallmark remains the largest maker of greeting cards}'' contains the answer to the question ``{\it What company sells the most greeting cards?}'',  a system would first encode the input sentence pairs into vector representations using complex lexical, syntactic, and semantic features. Subsequently, various similarity measures are computed from the obtained representations, and these similarity scores are then fed into classifiers to produce the decision.
In sequence generation tasks like machine translation and text summarisation, apart from feature engineering, extra components---such as language models, alignment models, and reordering models---are required to be incorporated into the system in order to achieve good performance \citep{koehn2009statistical}. 

While these models have achieved encouraging results, they suffer from three major drawbacks. First,  it is laborious to rely on a significant amount of feature engineering and to tune each model within a pipeline separately. Moreover, the external linguistic resources and tools, which are employed by these traditional models for obtaining syntactic and semantic features, are expensive to build, particularly for resource-low languages; and errors could be introduced from them, as these tools themselves are not perfect.
Another limitation of such feature-based models is the difficulty of
adapting to new domains, requiring separate feature
extraction and resource development or identification steps for every domain.

At the same time, neural networks have been widely used and achieved successes in the fields of computer vision \citep{krizhevsky2012imagenet,he2016deep}, speech recognition \citep{graves2013speech,chan2016listen}, and natural language processing \citep{dyer2015transition,bahdanau2014neural}. 
One dominant feature of neural networks is their ability to learn hierarchical representations, which models based on hand-crafted features cannot learn well.
In addition to the promising performance, neural network models can be trained in an end-to-end fashion with minimal domain knowledge and can be adapted to new domains easily.

This thesis investigates and advances the application of neural networks to sequence to sequence (seq2seq) mapping problems in NLP. 

In the remainder of the chapter, we first formalise the seq2seq mapping problem, and then describe the aim and contributions of the thesis. This chapter will be concluded by providing an outline of the thesis, together with a brief introduction of each chapter.

\section{Formalisation of Sequence to Sequence Mapping Problems}
The goal of seq2seq mapping problems is to decide the relationship between two sequences of data ({\it recognition}) or to generate a sequence of output data given a sequence of input data ({\it generation}). In the tasks that we tackle in the thesis, the data are either sequences of words (i.e. sentence pairs) or sequences of characters.  An example of the recognition task is textual entailment, where the model is learned to decide whether the meaning of a sentence can be inferred from the meaning of another. Machine translation is a typical example of the generation task of seq2seq mapping. 

Seq2seq mapping problems are superset of sequence labelling tasks---such as part-of-speech tagging and named entity recognition---in that there is no restriction on the lengths of sequence pairs. Moreover, we assume that the interaction between sequence pairs is not provided, although it may be useful to have this information available in order to make more precise predictions. We formalise the task of seq2seq mapping as follows:

Let $\mathcal{X}$ and $\mathcal{Y}$ be two separate sets of sequences. A sequence pair is represented as $(\boldsymbol{x}, \boldsymbol{y}) \in \mathcal{X} \times \mathcal{Y} $, where $\boldsymbol{x}  = (x_1, x_2, \dots, x_m)$, $\boldsymbol{y} = (y_1, y_2, \dots, y_n)$, and $x_i$ and $y_j$ are drawn from two fixed size vocabularies $\mathcal{V}_x$ and $\mathcal{V}_y$, respectively. We denote $S$ to be a set of training examples.

For the generation task, we have $(\boldsymbol{x}, \boldsymbol{y}) \in S$. The task is then to train a model $f$: $\mathcal{X} \rightarrow \mathcal{Y}$ to predict the output sequence $\hat{\boldsymbol{y}}$ for each input sequence $\boldsymbol{x}$ in the test set that minimises some task-dependent error measure. 

For the recognition task, let $l \in L$ be the label. The training set has triples of $(\boldsymbol{x}, \boldsymbol{y}, l)$ as data points. The task is to learn a classifier over these triples so that it can predict the labels of any additional sequence pairs $(\boldsymbol{x}', \boldsymbol{y}')$.
\section{Contributions}
In this thesis, we investigate the solutions to seq2seq mapping problems from three aspects, namely the representation of sequences, the alignment between sequences, and the utilisation of unpaired data. For each aspect, we propose novel models and evaluate the efficacy of these  models by experimenting on a number of popular and important seq2seq mapping tasks in NLP.  

We now summarise our major contributions. As the first attempt at tackling seq2seq mapping problems,  we consider the task of answer sentence selection in Chapter \ref{ch:sentence_model}. It is a task of selecting sentences containing the answer to a given question from a set of candidates. This chapter contributes to the thesis three-fold. First, by obtaining semantic vectors of sentences via distributed sentence  models, 
we determine the feasibility of using vectors for capturing the semantics of the entire sentences, and further discover how well these sentence vectors contribute to the semantic mappings between sentence pairs. In the architecture of our model, two distributed sentence models work in parallel, encoding questions and answers into their vector representations. These vectors are then used to learn the semantic similarity between them.  Second, we build sentence models based on various types of neural networks, and empirically compare their performance on answer sentence selection. By keeping the model for matching questions and answers consistent across different types of distributed sentence models, we examine to what extent these sentence models with different levels of sophistication affect the overall performance of the system. Third, we contribute to the field by setting a new state of the art on this task. Despite its simplicity, our models achieve superior performance over all prior work based on feature engineering on the answer selection task. 

We next attempt the generation tasks of seq2seq mapping with neural networks. Here (Chapter \ref{ch:ssnt}), we present a novel neural transduction model that aims to address the bottleneck of the vanilla encoder-decoders \citep{sutskever2014sequence,kalchbrenner2013recurrent,cho2014learning}. We introduce a latent segmentation to the encoder-decoder, which determines correspondences between tokens of the input sequence and those of the output sequences and thus enables the model to learn to generate and align simultaneously. We carefully design the parameterisation of the model so that exact inference can be carried out in polynomial time during training and during decoding the model can make incremental predictions. As an additional contribution, we provide detailed mathematical derivations and algorithms for training and decoding. In experiments on four representative seq2seq tasks in NLP---abstractive sentence summarisation, morphological inflection, and Chinese-English and French-English machine translation---we find that our neural transduction model outperforms the vanilla encoder-decoder by a substantial margin, and performs better or on par with the attentional seq2seq model while permitting online predictions.

Models based on neural networks can achieve good performance given that sufficient labelled data is available for training. However, in practice, for a number of tasks, there is only limited amounts of labelled data. In the case of seq2seq mapping tasks, in many domains vastly more unpaired output examples are available than input-output pairs. In Chapter \ref{ch:noisy_channel}, we explore approaches to leveraging unpaired data to improve neural models in seq2seq mapping problems. Our strategy is to formulate the seq2seq mapping problem as a noisy channel decoding problem and use recurrent networks to parameterise the source and channel models. With this formulation, the component models can be trained with paired and unpaired data. While training these models is straightforward, decoding is a computational challenge. We further contribute to the thesis by proposing a tractable and effective beam search decoder. Experiments on several seq2seq mapping tasks confirm the feasibility of this approach: the noisy channel model achieves significantly better performance than the direct model.

\section{Thesis Outline}

The main body of the thesis includes four chapters, with one chapter providing a necessary theoretical background for the thesis, and each of the three remainder chapters corresponding to one of our approaches to solving seq2seq mapping problems.

The context of the three supporting chapters (\ref{ch:sentence_model}, \ref{ch:ssnt}, \ref{ch:noisy_channel}) are based on three papers presented at various conferences and workshops. The work contained in these publications and presented in this thesis is principally the author's, except when stated otherwise in the relevant chapters.

Below we summarise each chapter of this thesis and elaborate on the material contained therein.

\begin{enumerate}[leftmargin=*,label=\bfseries Chapter \arabic*:,itemindent=2.5em]
	\setcounter{enumi}{1}
	
	\item \textbf{Neural Networks}~~~
	
	We provide an overview of the main concepts and algorithms of neural networks. We also introduce important models based on neural networks, including recurrent language models and seq2seq models with/without attention, which are closely related to our work.
	
	\item \textbf{The Role of Distributed Sentence Models}~~~
	
	We focus on the task of answer sentence selection, for which we propose to encode sentences into vector representations via distributed sentence models and measure the semantic relatedness of vector pairs via a bilinear model. We describe our model as well as prior work on this task and empirically compare their performance. We also compare the performance of distributed sentence models based on different types of neural networks. The publications related to the work presented in this chapter are:
	\begin{itemize}
		\item [] \bibentry{Yu:2014}
		\item [] \bibentry{miao:2016}
	\end{itemize}
	
	\item \textbf{Sequence Transduction}~~~
	
	Orthogonal to the attentive seq2seq model \citep{bahdanau2014neural}, we propose a neural transduction model that learns to generate and align simultaneously. Controlled by a latent variable, the model alternates between encoding more of the input sequence and decoding output tokens from the encoded representations. We evaluate the model by experimenting on a number of sequence generation tasks. Apart from obtaining encouraging results, we also find that the alignments predicted by the model are highly intuitive. Furthermore, this chapter includes a discussion of other recent work on employing latent variables in neural networks. The work presented in this chapter is based on the following publication:
	
	\begin{itemize}
		\item [] \bibentry{yu:2016}
	\end{itemize}
	
	\item \textbf{Incorporating Unpaired Data}~~~

	We formulate seq2seq mapping as a noisy channel decoding problem. Under this formulation,  both paired and unpaired data can be leveraged to train the component models. This chapter starts by introducing the noisy channel framework, including its definition, advantages (besides incorporating unpaired data), and its important historical positions in different fields such as speech recognition and machine translation. We subsequently explain why and how  the neural transduction model proposed in Chapter \ref{ch:ssnt} can work as a channel model, but the conventional attentive models cannot. After presenting a decoding algorithm, we examine the efficacy of the neural noisy channel model by experimenting on three seq2seq mapping tasks, abstractive sentence summarisation, machine translation, and morphological inflection. The work presented in this chapter is based on the following publication:
	\begin{itemize}
		\item [] \bibentry{yu:2017}
	\end{itemize}
	
	\item \textbf{Conclusion} ~~~
	
	We shortly summarise the thesis with future work.
\end{enumerate}

%% file: chapter_3.tex
 \chapter{Neural Networks}
\label{ch:nn}

\begin{chapterabstract}
This chapter provides an overview of neural networks. This review begins by describing the architecture of different types of neural networks, including regular fully connected feedforward neural networks, convolutional neural networks, and recurrent neural networks. For each type of neural networks, we will give a formulation of the forward pass (for loss calculation) and the backward pass (for gradient calculation). 
Subsequently, we will introduce several important models built from recurrent neural networks, namely recurrent language models, the encoder-decoder paradigm, and the seq2seq model with attention. These models are fundamental to our work in the thesis. The encoder-decoder paradigm is the foundation of our neural transduction model to be introduced in Chapter \ref{ch:ssnt}. Our neural noisy channel model to be presented in Chapter \ref{ch:noisy_channel} combines the neural transduction model and a recurrent language model in a noisy channel framework. 
\end{chapterabstract}
Artificial Neural networks (ANNs) were originally introduced for mathematically modelling the way that biological brains process information \citep{mcculloch1943logical,rosenblatt1962principles,rumelhart1985learning}. Although  available for decades, not until recently ANNs have experienced a resurgence  due to the significant improvement of computational power, the availability of larger datasets, and the introduction of new algorithms. Models based on neural networks have achieved promising results in many fields, such as computer vision \citep{krizhevsky2012imagenet,xu2015show}, speech recognition \citep{graves2013speech} and natural language processing \citep{bahdanau2014neural}. 
This success is attributed to their ability to learn hierarchical representations, which traditional methods that rely upon hand-engineered features do not have.

The basic elements of an ANN are a number of simple processing units called neurons or nodes and the weighted connections between them. 
A neuron in the biological system can be activated by some outside process or stimuli from neighbouring neurons, produces its own activation, and passes the output to other nonactive neurons. 
Analogous to that, the artificial neuron receives a set of real-valued inputs, processes them with its {\it activation function} and spreads the results to the rest of the network through the weighted connections. 
Neurons are typically organised by layers. Inputs are passed to the input layer, and outputs are produced from the output layer.

Neural networks vary by the type of neurons being used, the way these neurons are connected, and the overall topology. One important property that distinguishes the types of neural networks is whether there are feedback connections, in which outputs of the model are fed back into itself. Networks with no feedback connections, i.e. having the structure of a directed acyclic graph, are called feedforward neural networks or {\it multilayer perceptrons} \citep{rumelhart1985learning} (\S\ref{mlp}) . Multilayer perceptrons (MLPs) are well suited for pattern recognition tasks. In particular, {\it convolutional neural networks} \citep{lecun1989backpropagation,lecun1998gradient} (\S\ref{cnn}), 
a specialised form of MLPs, are excellent models for object recognitions from images. Those models with feedback connections are known as recurrent neural networks (RNNs) (\S\ref{rnn}), which are more suitable for sequence labelling tasks compared to MLPs.  The neural seq2seq models with one RNN working as the encoder and another RNN as the decoder is a standard approach for neural machine translation and many other seq2seq mapping problems. We review seq2seq models with and without an attention mechanism in \S\ref{seq2seq}. CNNs, RNNs, and seq2seq models are important preliminaries for the thesis.

Neural networks can be viewed as a general class of parametric nonlinear functions $f$ from  input variables $\mathbf{x}$ to output variables $\mathbf{y}$: $\mathbf{y} = f(\mathbf{x}\, ;\, \boldsymbol{\theta})$. We train neural networks using the given dataset aiming to determine the parameters $\boldsymbol{\theta}$ that best approximate the true function $f^*$ over all the input-output pairs of the dataset. Training is usually done via gradient descent. This requires the evaluation of derivatives of the objective function with respect to parameters, which can be obtained efficiently using the
technique of error {\it backpropagation}. We present the backpropagation algorithm for MLPs in \S\ref{bp} and the backpropagation through time for RNNs in \S\ref{bptt}.

\section{Multilayer Perceptrons}
\label{mlp}
MLPs are the  fundamental type of neural network. In this section, we provide a brief overview of the building blocks of MLPs, namely the architecture, activation functions, and forward and backward propagation.

\subsection{Forward Propagation}
\begin{figure}
	\centering
	\includegraphics[scale=0.75]{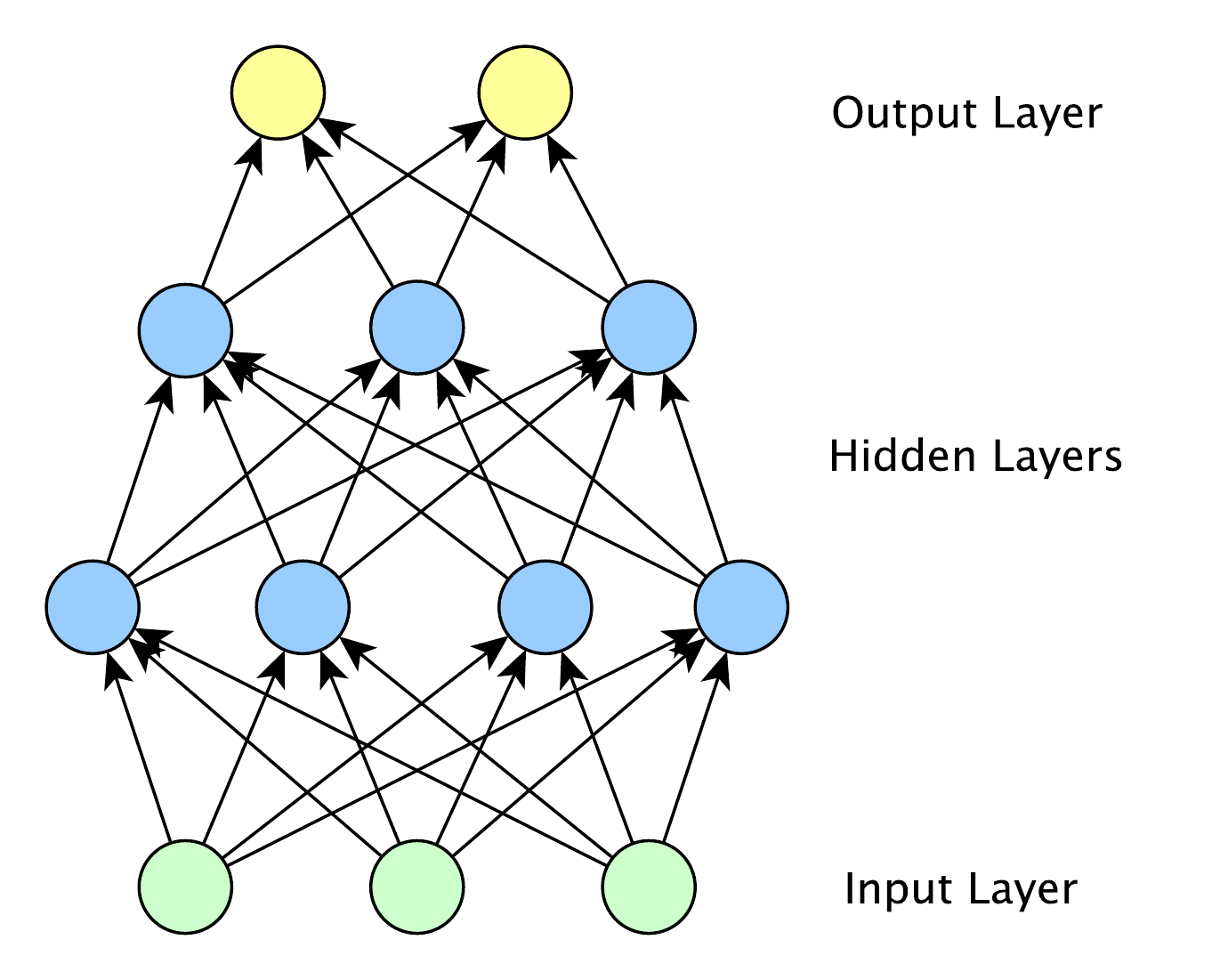}
	\caption[Architecture of feedforward neural networks]{The architecture of fully connected feedforward neural networks. The bottom layer is the input layer, and the topmost layer is the output layer. The layers in between are hidden layers. Activations are passed from lower layers to their adjacent higher layers.}
	\label{mlp_figure}
\end{figure}

As shown in Figure \ref{mlp_figure}, neurons in MLPs are arranged in layers. The input $\mathbf{x}$ to the network is presented to the input layer, whose outputs lead to successive activations of the hidden layers until the output $\mathbf{\hat{y}}$ is generated from the output layer. The output value of each neuron is calculated by applying the activation function $\phi$ to the weighted sum of the outputs of the neurons in the previous layer. Formally, consider an MLP with $L$ layers, $J$ and $I$ units on the $l$-th and ($l+1$)-th layers, respectively. We have
\begin{align}
a_i^{l+1} &= \sum_{j=1}^{J} w_{ij}^{l} b^{l}_j + w_{i0}^{l} \\
b_i^{l+1} &= \phi (a_i^{l+1}),
\end{align}
where $w_{ij}^{l}$ denotes the weight from neuron $j$ in layer $l$ to unit $i$ in layer $l+1$, $w_{i0}^{l}$ is the bias term. The term $b^{l}_j $ is the {\it activation} from the $j$-th neuron of layer $l$. It has the value of $x_j$ if the current layer is the input layer.

Common choices of activation functions include the logistic sigmoid function
\begin{align}
\sigma(x) = \frac{1}{1 + e^{-x}},
\end{align}
the hyperbolic tangent
\begin{align}
\tanh(x) = \frac{e^{2x} - 1}{e^{2x} + 1},
\end{align}
and the rectified linear unit (ReLU)
\begin{align}
\ReLU(x) = \max(0, x).
\end{align}
 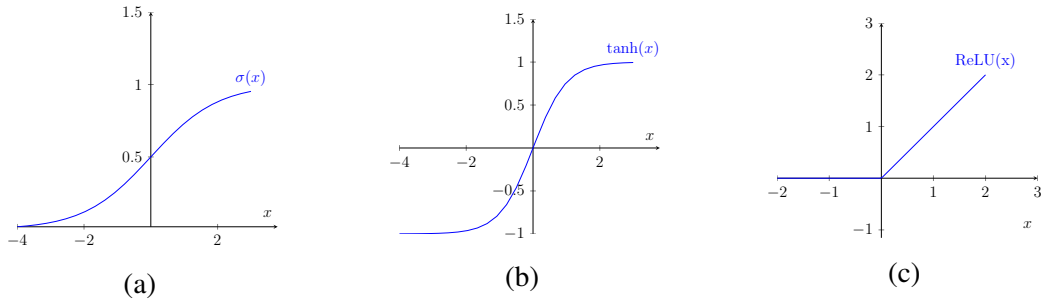
\begin{figure}
 	\begin{subfigure}{.33\textwidth}
 		\centering
 		\begin{tikzpicture}[scale=0.5]
 		\begin{axis}[axis lines=middle,
 		xmax=3.8,ymax=1.5,
 		xlabel = $x$,
 		x label style={at={(1,0.02)}},
 		]
 		\addplot[blue,domain=-4:3]  {1/(1+pow(e,-x))} node [above]{$\sigma(x)$};
 		\end{axis}
 		\end{tikzpicture}
 		\caption{}
 		\label{fig:act1}
 	\end{subfigure}%
 	\begin{subfigure}{.33\textwidth}
 		\centering
 		\begin{tikzpicture}[scale=0.5]
 		\begin{axis}[axis lines=middle,
 		xmax=3.8,ymax=1.5,
 		xlabel = $x$,
 		x label style={at={(1,0.41)}},
 		]
 		\addplot[blue,domain=-4:3]  {(pow(e,2*x)-1)/(pow(e,2*x)+1)} node[above]{$\tanh(x)$};
 		\end{axis}
 		\end{tikzpicture}
 		\caption{}
 		\label{fig:act2}
 	\end{subfigure}%
 	\begin{subfigure}{.33\textwidth}
 		\centering
 		\begin{tikzpicture}[scale=0.5]
 		\begin{axis}[	axis lines=middle,
 		xmax=3,ymax=3,
 		axis equal,
 		xlabel = $x$,
 		x label style={at={(1,0.02)}},
 		]
 		\addplot[blue,domain=-2:2,samples=100]  {max(0,x)} node[above] {$\mathrm{ReLU(x)}$};
 		\end{axis}
 		\end{tikzpicture}
 		\caption{}
 		\label{fig:act3}
 	\end{subfigure}
 	\caption[Plots of different activation functions]{Plots of different activation functions. (a) The sigmoid function, (b) the hyperbolic tangent, (c) the rectified linear unit.}
 	\label{act_func}
 \end{figure}

Figure \ref{act_func} shows the diagrams of these functions. As can be seen, in the $\tanh$ and the sigmoid functions gradients decrease significantly as the absolute values of the inputs increase.  By contrast, $\ReLU$s are similar to linear units --- the gradients passing through $\ReLU$s remain large and consistent whenever the unit is active (corresponding to the points on the right half of the graph). This property makes ReLUs a good choice of the activation function for MLPs \citep[Chapter 6]{Goodfellow-et-al-2016}. However, special consideration must be taken when initialising the parameters of the networks during experiments to make sure $\ReLU$s are initially active for most inputs in the training data. The $\tanh$ and sigmoid functions are more commonly used in recurrent neural networks. 

The number of neurons and the choice of activation functions at the output layer depends upon the task. For binary classification tasks, we apply the logistic sigmoid in the single node of the output layer. For multiclass classification with $K$ distinct classes $(K > 2)$, {\it softmax} function \citep{bridle1990probabilistic} is leveraged to normalise the output activations of the $K$ nodes:
\begin{align}
\hat{y}_k = \frac{e^{a_k}}{\sum_{k'=1}^K  e^{a_{k'}}}.
\end{align}
For regression problems, we have linear output. The natural language processing tasks that we deal with in this thesis have either logistic sigmoid function or softmax as the activation function at the output layer.

\subsection{Backpropagation}
\label{bp}
Neural networks are trained using stochastic gradient descent (SGD) that iteratively reduces the loss function. In most cases, the loss function  (or objective function) follows the principle of maximum likelihood. That is, we aim to minimise the cross-entropy error between the predicted outputs and the true targets (alternatively negative log likelihood). The equation of the gradient update is 
\begin{align}
\boldsymbol{\theta} \leftarrow \boldsymbol{\theta} - \eta \nabla_{\boldsymbol{\theta}} \mathcal{ L}_n,
\end{align}
where $\eta$ denotes the learning rate, $\nabla_{\boldsymbol{\theta}}  \mathcal{ L}_n$ is the gradient of the objective function with respect to the parameters $\boldsymbol{\theta}$ calculating on the $n$-th mini-batch of data points.  Recently, many variants of SGD have been developed to enable faster and better convergence. These include Adagrad \citep{duchi2011adaptive}, RMSprop \citep{rmsprop}, and Adam \citep{DBLP:journals/corr/KingmaB14}, which adapt the learning rate of each parameter separately during learning.

The backpropagation algorithm \citep{rumelhart1985learning} provides an efficient way of calculating the gradient of the loss function w.r.t. parameters in MLPs. The algorithm essentially applies the chain rule repeatedly for the partial derivatives of the parameters at each layer. Here we describe the backpropogation algorithm in a non-vectorised way, i.e. we calculate derivatives of scalars w.r.t. scalars. The first step is to do a forward pass to calculate the activations $b_i^{l}$ of each node in the network and the output $\hat{\mathbf{y}}$ at the output layer. With the predicted output $\hat{\mathbf{y}}$ and the target $\mathbf{y}$ we obtain the loss $\mathcal{ L}$. Subsequently, the backward pass starts by calculating the derivatives of the loss function with respect to the output nodes:
\begin{align}
\delta_k^{L}= \frac{\partial \mathcal{L} }{\partial \hat{y}_k}\phi_{k}'(a_k^{L}).
\end{align}
Here $\phi_k$ is the activation function at the output layer. For the derivative of the parameters in the hidden layers, we introduce the error term $\delta_j^{l}$ that measures how much the $j$-th node at the $l$-th layer is “responsible” for the error in the output. The formal definition of $\delta_j^{l}$ is
\begin{align}
\delta_j^{l} = \frac{\partial \mathcal{L}}{\partial a_j^{l}}.
\end{align}
Based on the chain rule, we can rewrite $\delta_j^{l}$ as
\begin{align}
\delta_j^{l} = \frac{\partial \mathcal{L}}{\partial a_j^{l}} = \sum_{i=1}^I \frac{\partial \mathcal{L}}{\partial a_i^{l+1}} \frac{\partial a_i^{l+1}}{\partial b_j^{l}} \frac{\partial b_j^{l}}{\partial a_j^{l}}.
\end{align}
The use of the summation is due to the fact that the activation of every node in the $l$-th layer contributes to the activation of any node in the $(l + 1)$-th layer. The equation above can be further simplified as:
\begin{align}
\delta_j^{l} = \phi'(a_j^{l}) \sum_{i=1}^I \delta_i^{l+1} w_{ij}^{l}.
\end{align} 
Therefore, the error term of the nodes in the hidden layers can be calculated recursively. Having the values $b_j^{l}$ calculated from the forward pass, and $\delta_j^{l}$ obtained from the backward pass, we arrive at the derivatives of the loss function with respect to the network weights:
\begin{align}
\frac{\partial \mathcal{L}}{\partial w_{ij}^{l}} = \frac{\partial \mathcal{L}}{\partial a_i^{l+1}} \frac{\partial a_i^{l+1}}{\partial w^l_{ij}} = \delta_i^{l+1} b_j^{l}.
\end{align} 
\section{Convolutional Neural Networks}
\label{cnn}
Convolutional Neural Networks (CNNs) are a variant of MLPs that take a biological inspiration from the visual cortex. Compared to regular fully connected MLPs with the same number of hidden units, CNNs are easier to train and contain much fewer parameters.

\subsection{Architecture}
\begin{figure}[h]
	\centering
	\begin{subfigure}[b]{0.55\textwidth}
		\includegraphics[scale=0.75]{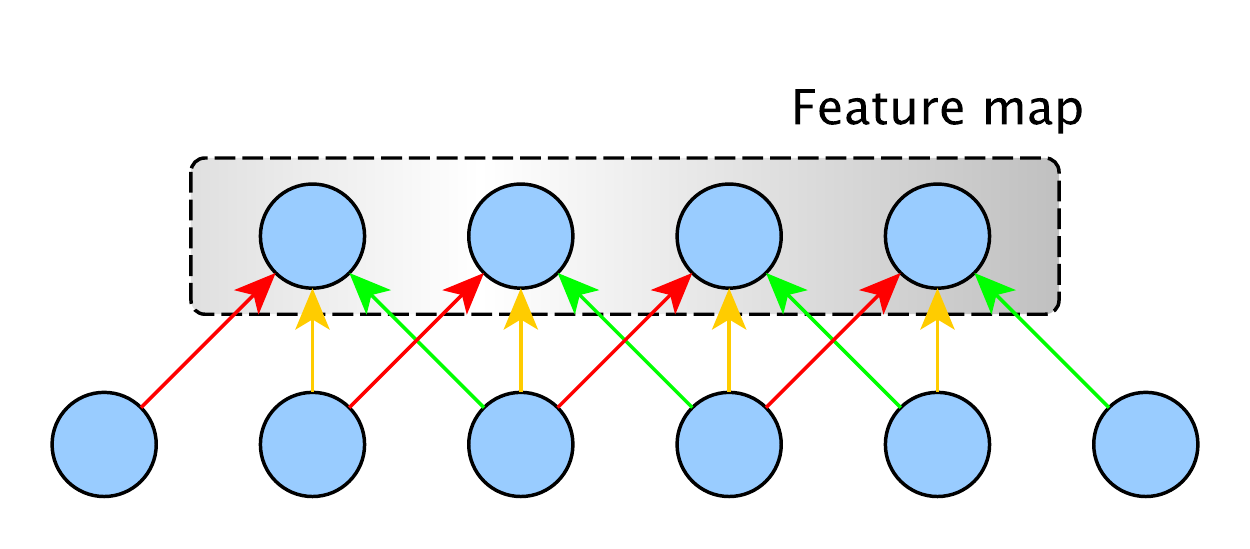}
		\caption{}
		\label{cnn1_figure}
	\end{subfigure}
	
	\begin{subfigure}[b]{0.55\textwidth}
		\includegraphics[scale=0.75]{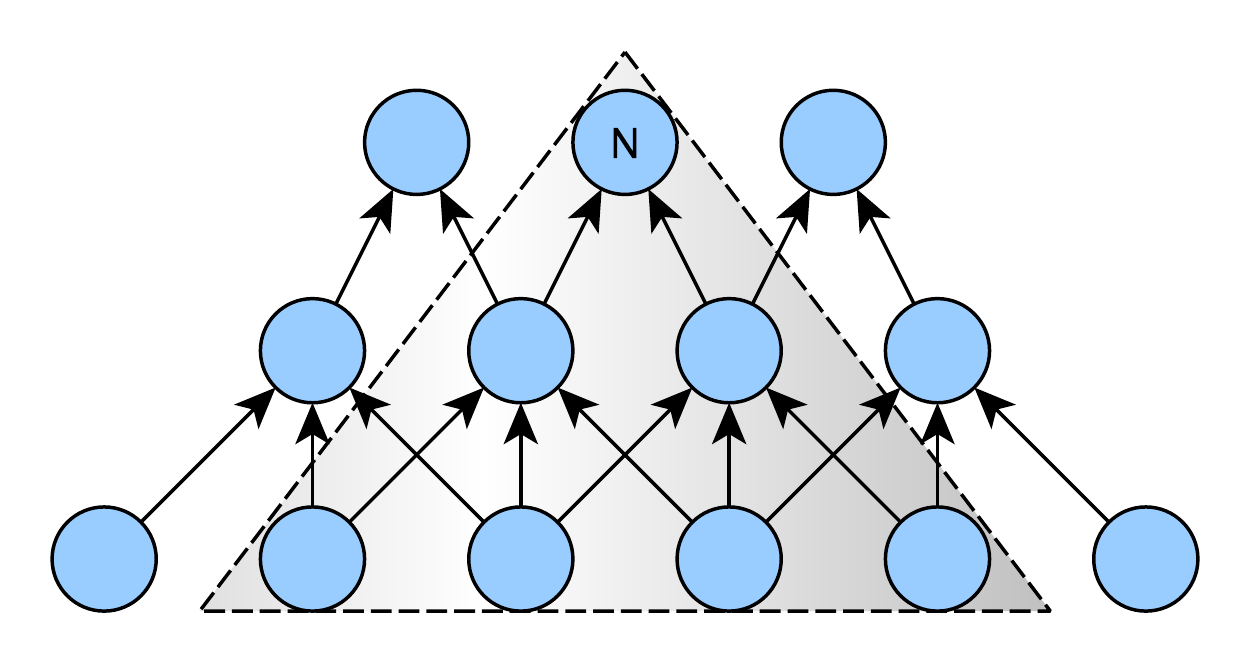}
		\caption{}
		\label{cnn2_figure}
	\end{subfigure}
	\caption[Architecture of convolutional neural networks]{The architecture of convolutional neural networks. Each figure shows a portion of hidden layers. In CNNs, the node at the $l$-th layer is connected to a small number of consecutive nodes in the $(l-1)$-th layer. In (a), weights of edges in the same colour are shared. In (b), the shaded area represents the receptive field of the node $N$.}
	\label{cnn_figure}
\end{figure}

CNNs are comprised of the interleaves of {\it convolutional layers} and {\it pooling layers} optionally followed by fully connected layers. In the convolutional layer, as illustrated in Figure \ref{cnn_figure},  {\it filters} (or {\it kernels}) are applied repeatedly to small local regions of neurons in the previous layer, and their outputs form {\it feature maps}. These local regions are called {\it receptive fields}, which are analogous to regions of the visual field to which the cells in visual cortex are sensitive. The parameters of each filter are shared for any input region, but they are different across different filters. This mechanism of parameter sharing not only reduces the number of parameters but also helps to capture local features of the input that are invariant to their locations. As shown in Figure \ref{cnn2_figure}, by stacking multiple convolutional layers, features captured by deeper layers become increasingly global. To take the task of image object recognition as an example, while the filters at lower layers may capture horizontal edges or curves, those at top layers are able to capture the features of the entire digits or faces \citep{DBLP:conf/eccv/ZeilerF14}.

A pooling function aggregates statistics of features at various locations. For example, the {\it max pooling} \citep{zhou1988computation} operation partitions the feature map into a set of non-overlapping sub-regions; and for each sub-region, it outputs the maximum value. Other pooling functions include average pooling and L2-norm pooling. A nice property of pooling is that it provides a way of getting fixed size output matrix, which typically is required for classification. It also reduces the number of parameters further. Most importantly, pooling is invariant to small translations of the input, which is desirable in many tasks, e.g.  image recognition and audio recognition. That is, the pooled outputs will remain approximately the same even when the image undergoes slight shifts or rotations.

CNN models have also been shown to be effective in natural language processing, mostly as an approach to composing word vectors into sentence vectors. In Chapter \ref{ch:sentence_model}, we present the use of CNNs for sentence modelling in the recognition task of seq2seq mapping.

\section{Recurrent Neural Networks}
The MLPs introduced in the previous sections cannot capture orderings efficiently.
In practice, for many tasks, it is necessary to consider orderings and dependencies. For example, in language modelling, the prediction of the next word extensively relies on the context, i.e. the words before it. Recurrent neural networks (RNNs) are extensions of feedforward neural networks with cyclical connections. The connections enable RNNs to summarise the previous inputs in their internal states which subsequently affect the output predictions. Such property makes them excellent models for sequence modelling. Numerous varieties of RNNs have been proposed, including Elman networks \citep{elman1990finding}, Long Short Term Memory (LSTM) networks \citep{hochreiter1997long}, Gated Recurrent Units (GRUs) \citep{cho2014learning}, and the more recent Neural Turing Machines (NTM) \citep{graves2014neural,graves2016hybrid}.

In the rest of the section, we first review general RNNs and then introduce the LSTM, which is the type of RNN that we use in the thesis.

\label{rnn}
\subsection{Forward Propagation}
\begin{figure}
	\centering
	\begin{subfigure}{0.35\textwidth}
		\includegraphics[scale=0.75]{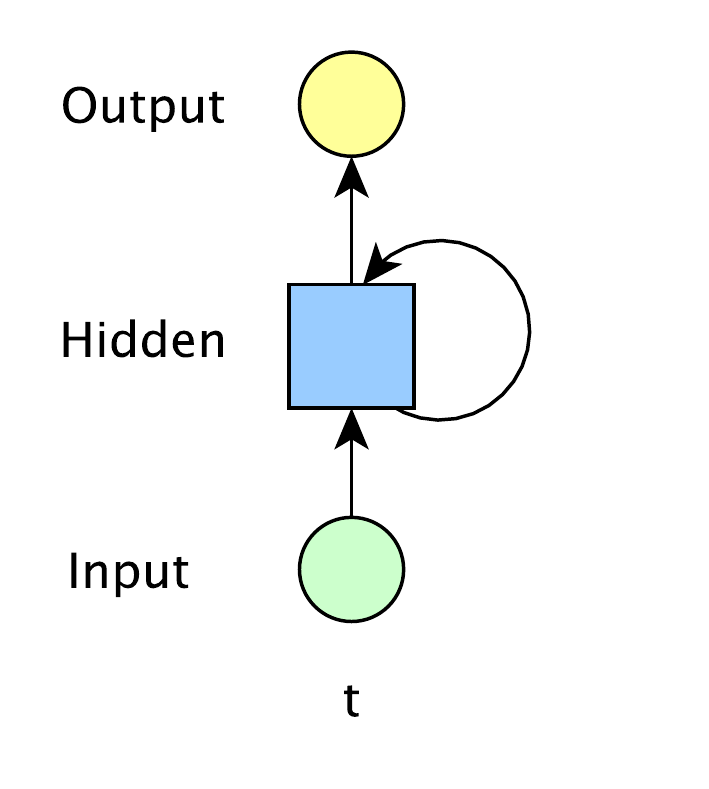}
		\caption{}
		\label{rnn1_fold}
	\end{subfigure}
	\begin{subfigure}{0.55\textwidth}
		\includegraphics[scale=0.75]{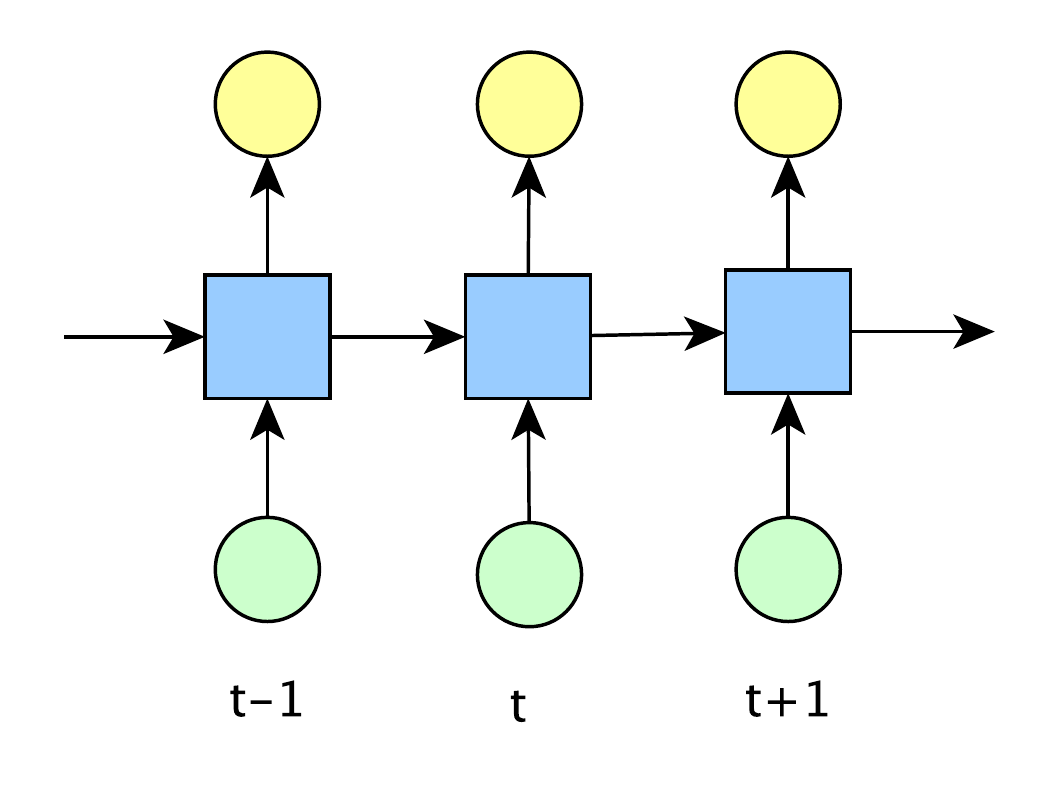}
		\caption{}
		\label{rnn2_unfold}
	\end{subfigure}
	\caption[Architecture of recurrent neural networks]{The architecture of recurrent neural networks: (a) the folded version of an RNN, (b) the unfolded version of an RNN with respect to time. The output layer is optional.}
	\label{nn_figure}
\end{figure}
The notion of time is introduced in RNNs. Figure \ref{rnn1_fold} shows the architecture of an RNN and Figure \ref{rnn2_unfold} is its unfolded version with respect to time. Consider an RNN with $J$ input nodes, $I$ nodes at the hidden layer and $K$ output nodes. Let  the sequence $\mathbf{x}$  of length $T$ be the input to the network. At time $t$, nodes at the hidden layers receive input from current data point $\mathbf{x}_t$ and the hidden layer activations from time $(t-1)$. Their activations are generated by the following equation:
\begin{align}
\label{rnn_eq}
\mathbf{h}_t = \phi (\mathbf{W}^{xh}\mathbf{x}_t + \mathbf{W}^{hh}  \mathbf{h}_{t-1} + \mathbf{b}),
\end{align} 
where $\mathbf{h}_t$\footnote{By convention, $\mathbf{h}$ is used to denote hidden layer activations in RNNs instead of using $\mathbf{b}$ as in MLPs.} denotes the activations of hidden layers at time $t$, $\phi$ is the activation function, $\mathbf{b}$ is the bias term, and $\mathbf{W}^{xh}$ and $\mathbf{W}^{hh}$ are the weight matrices between the input and the hidden layer and between the hidden layer  and itself at adjacent time steps, respectively. Notably, these weight matrices are shared across different timesteps, which significantly reduces the number of parameters needed to learn. The initial state $\mathbf{h}_0$ is usually set to 0. The non-vectorised version of Equation \ref{rnn_eq} is expressed as:\footnote{For the simplicity of notation, $w_{ij}^{xh}$ and $w_{ii'}^{hh}$ are written as $w_{ij}$ and $w_{ii'}$, respectively.}
\begin{align}
a_i^t & = \sum_{j=1}^{J} w_{ij} x_j^t + \sum_{i'=1}^{I} w_{ii'} h_{i'}^{t-1} + w_{i0} \\
h_i^t & = \phi (a_i^t).
\end{align}
Optionally, at each timestep $t$ output is emitted which can either
be discrete or real-valued. In the language generation tasks that we tackle in this thesis, the output values are discrete (usually words in a vocabulary), and therefore we apply the activation function $\softmax$ to obtain the probability distribution over all the output classes:
\begin{align}
\hat{y}_t \sim \softmax(\mathbf{W}^{hy} \mathbf{h}_t + \mathbf{b}).
\end{align}
The matrix $\mathbf{W}^{hy}$ is the weight matrix between the hidden layer and the output layer and $\mathbf{b}$ is the bias term. 
The corresponding non-vectorised expression is
\begin{align}
a_k^t & =  \left(\sum_{i=1}^{I} w_{ki} h_i^t + w_{k0}\right). \\
\hat{y}^t_k &= \frac{e^{a^t_k}}{\sum_{k'=1}^K  e^{a^t_{k'}}}.
\end{align} 
For tasks such as machine translation and language modelling, the number of output classes can range from 50k to 100k, which makes the above matrix-vector multiplication a major bottleneck of the model in terms of memory usage and computational efficiency.  How to address this issue is an important research topic \citep[\textit{inter alia}]{DBLP:conf/aistats/MorinB05,DBLP:conf/icml/MnihT12,DBLP:conf/acl/JeanCMB15} in the field of language generation, but it is out of the scope of this thesis.

The loss of the entire sequence is the sum of the loss (in $\log$ space) at each timestep:
\begin{align}
\mathcal{L} = \sum_{t=1}^{T} \mathcal{ L}(\hat{{y}}_t, {y}_t).
\end{align}

In addition to generation tasks, RNNs are also suitable for composing word vectors into a sentence vector. In this case, the output layer at each timestep is not required, and usually the last hidden state vector $\mathbf{h}_T$ is used as the sentence vector.
\subsection{Backpropagation Through Time}
\label{bptt}
The gradients in RNNs are calculated via backpropagation through time (BPTT) \citep{williams1995gradient,werbos1990backpropagation}. BPTT follows the same procedure as the backpropagation algorithm \citep{rumelhart1985learning} for MLPs. Assume that we have done the forward pass that calculates the activations of the hidden layers and the outputs at each timestep.  We have also obtained the loss.
Similar to MLPs, we define the error term as:
\begin{align}
\delta_i^t = \frac{\partial \mathcal{ L}}{\partial a_i^t}.
\end{align}
The first step of the backward pass is to get the derivatives of the loss with respect to the output nodes:
\begin{align}
\delta_k^t = \frac{\partial \mathcal{ L}}{\partial \hat{y}^t_k}\softmax'(a_k^t).
\end{align}
For the hidden layers, the error $\delta^t$ for nodes at time $t$ is not only propagated from the output layer but also from the hidden layer at time $(t+1)$. Therefore, we have
\begin{align}
	\label{bptt-err}
	\delta_i^t = \phi'(a_i^t) \left( \sum_{k=1}^{K} \delta_k^t w_{ki}  + \sum_{i'=1}^{I} \delta_{i'}^{t+1} w_{i'i} \right).
\end{align}
Starting from the base case $\delta_i^{T+1} = 0$, we calculate $\delta_i^t$ for $t \in [1, T]$ backwards by recursively applying Equation \ref{bptt-err}. Finally, we have the derivatives of the loss function with respect to the parameters:
\begin{align}
	\frac{\partial \mathcal{ L}}{\partial w_{ij}} = \sum_{t=1}^{T} \frac{\partial \mathcal{L}}{\partial a_i^t} \frac{\partial a_i^t}{\partial w_{ij}} = \sum_{t=1}^{T} \delta_i^t h_j^t,
\end{align}
where the summation over $T$ is due to the fact that parameters are shared across timesteps.
\subsection{Deep Recurrent Neural Networks}
\begin{figure}
	\centering
	\includegraphics[scale=0.75]{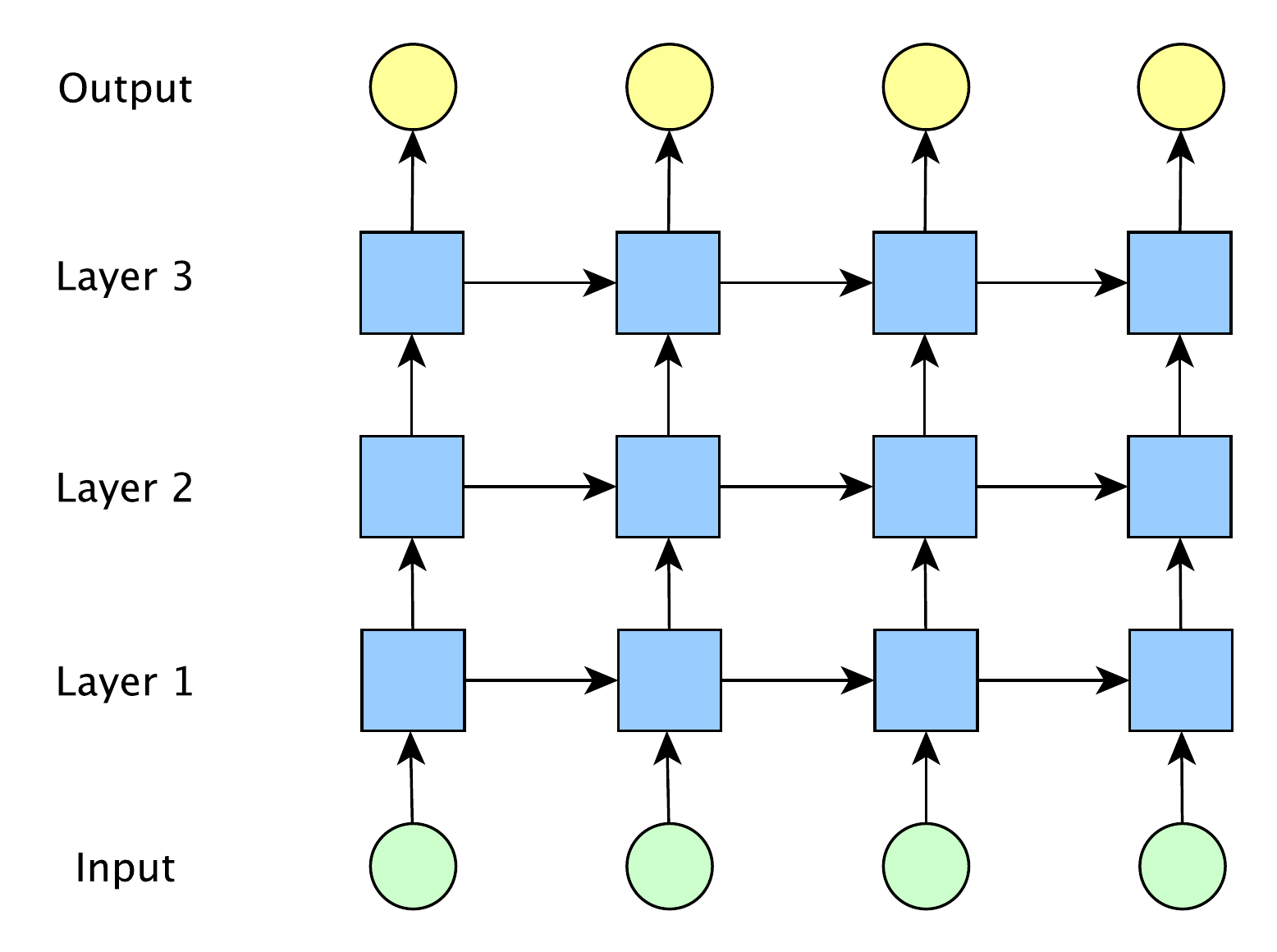}
	\caption[Architecture of deep recurrent neural networks.]{Architecture of deep recurrent neural networks. The hidden state on the intermediate hidden layer $l$ at timestep $t$ takes as input the output of the hidden state in the $(l-1)$-th layer at timestep $t$ and that in the $l$-th layer at timestep $t-1$. }
	\label{fig:deep_rnn}
\end{figure}
In practice, we usually stack RNNs into multiple layers (Figure \ref{fig:deep_rnn}). In deep RNNs, the input layer, output layer, and the lowest hidden layer are the same as those in shallow RNNs. The hidden state in the intermediate layers is constructed from two states: the one in the previous layer and the one in the same layer,
\begin{align}
\mathbf{h}^l_t = \phi (\mathbf{W}^l_{h'h}\mathbf{h}^{l-1}_t + \mathbf{W}^l_{hh}\mathbf{h}^l_{t-1} + \mathbf{b}),
\end{align}
where $\mathbf{h}_t^l$ denotes the hidden state in the $l$-th layer at timestep $t$, $\mathbf{W}^l_{h'h}$ is the weight matrix between the hidden state in the previous layer and that in the current layer, and $\mathbf{W}^l_{hh}$ is the weight matrix between the hidden states in the same layer. Compared to shallow RNNs, deep RNNs are able to create higher-level abstractions and to capture more non-linearities within data.

\subsection{Bidirectional Recurrent Neural Networks}
\begin{figure}
	\centering
	\includegraphics[scale=0.75]{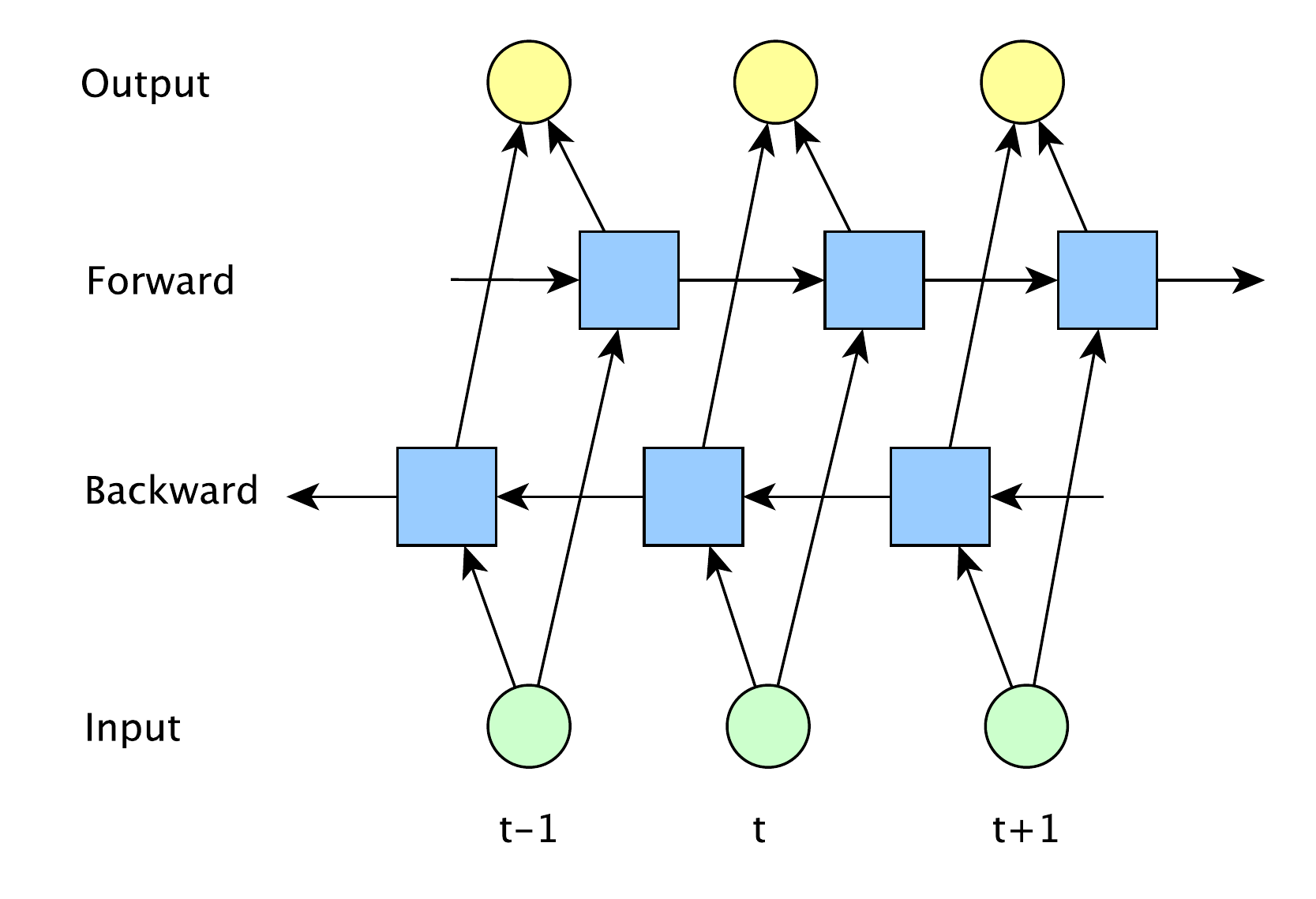}
	\caption[Architecture of bidirectional recurrent neural networks]{The architecture of bidirectional recurrent neural networks. It is constructed by two separate RNNs, one to read sequences in the forward direction and the other to read sequences in the backward direction. At each timestep, the corresponding hidden states of both RNNs are combined to predict the output.}
	\label{fig:brnn}
\end{figure}
One limitation of standard RNNs is that the information accumulated at the current timestep $t$ is only from the past. For many tasks, it is useful to access both the past and future information in order to predict an output accurately. For example, consider the task of part-of-speech tagging for sentences, the part-of-speech tag for the current word depends on both the words preceding it and those following it. Bidirectional recurrent neural networks (BiRNNs) \citep{schuster1997bidirectional} address such limitation. In the architecture of BiRNNs (Figure \ref{fig:brnn}), there are two separate recurrent hidden layers, both of which are connected to the same output layer. The two hidden layers are differentiated from each other in that the first one processes the input sequence in the forward direction as the standard RNNs, whereas the second one passes information in the opposite direction.  The equations corresponding to a BiRNN are
\begin{align}
\mathbf{h}^{\rightarrow}_t &= \phi(\mathbf{W}^{xh} \mathbf{x}_t + \mathbf{W}^{hh} \mathbf{h}^{\rightarrow}_{t-1} + \mathbf{b}^h) \\
\mathbf{h}^{\leftarrow}_t & = \phi(\mathbf{W}^{xs} \mathbf{x}_t + \mathbf{W}^{ss} \mathbf{h}^{\leftarrow}_{t+1} + \mathbf{b}^s) \\
\hat{y}_t &\sim \softmax(\mathbf{W}^{hy} \mathbf{h}^{\rightarrow}_t + \mathbf{W}^{sy} \mathbf{h}^{\leftarrow}_t + \mathbf{b}^y),
\end{align}
where $\mathbf{h}^{\rightarrow}_t$ and $\mathbf{h}^{\leftarrow}_t$ are the hidden state vectors in the forward and backward directions, respectively. The weight matrices between the two separate hidden layers are not shared.

As will be discussed in \S\ref{seq2seq_att}, it is very effective to encode the input sequences with BiRNNs in the attentional seq2seq models, as it allows the model to attend the context surrounding a certain input token. Although BiRNNs are superior to standard RNNs in sequence modelling in general \citep{graves2012supervised,bahdanau2014neural}, they cannot be applied to tasks that require online predictions, as future inputs cannot be accessed.
\subsection{Long Short-Term Memory}

\begin{figure}
\centering
\begin{tikzpicture}[scale=1.5,
prod/.style={circle, draw, inner sep=0pt},
ct/.style={circle, draw, inner sep=5pt, ultra thick, minimum width=10mm},
ft/.style={circle, draw, minimum width=8mm, inner sep=1pt},
filter/.style={circle, draw, minimum width=7mm, inner sep=1pt, path picture={\draw[thick, rounded corners] (path picture bounding box.center)--++(65:2mm)--++(0:1mm);
		\draw[thick, rounded corners] (path picture bounding box.center)--++(245:2mm)--++(180:1mm);}},
mylabel/.style={font=\scriptsize\sffamily},
>=LaTeX
]

\node[ct, label={[mylabel]Cell}] (ct) {$\mathbf c_t$};
\node[filter, right=of ct] (int1) {};
\node[prod, right=of int1] (x1) {$\times$}; 
\node[right=of x1] (ht) {$\mathbf h_t$};
\node[prod, left=of ct] (x2) {$\times$}; 
\node[filter, left=of x2] (int2) {};
\node[prod, below=5mm of ct] (x3) {$\times$}; 
\node[ft, below=5mm of x3, label={[mylabel]right:Forget Gate}] (ft) {$\mathbf f_t$};
\node[ft, above=of x2, label={[mylabel]left:Input Gate}] (it) {$\mathbf i_t$};
\node[ft, above=of x1, label={[mylabel]left:Output Gate}] (ot) {$\mathbf o_t$};

\foreach \i/\j in {int2/x2, x2/ct, ct/int1, int1/x1,
	x1/ht, it/x2, ot/x1, ft/x3}
\draw[->] (\i)--(\j);


\draw[->] (ct) to[bend right=30] (x3);
\draw[->] (x3) to[bend right=30] (ct);

\node[fit=(int2) (it) (ot) (ft), draw, inner sep=0pt] (fit) {};

\draw[<-] (fit.west|-int2) coordinate (aux)--++(135:7mm) node[left]{$\mathbf x_t$};
\draw[<-] (fit.west|-int2) coordinate (aux)--++(225:7mm) node[left]{$\mathbf h_{t-1}$};

\draw[<-] (fit.north-|it) coordinate (aux)--++(135:7mm) node[above]{$\mathbf x_t$};
\draw[<-] (fit.north-|it) coordinate (aux)--++(45:7mm) node[above]{$\mathbf h_{t-1}$};

\draw[<-] (fit.north-|ot) coordinate (aux)--++(135:7mm) node[above]{$\mathbf x_t$};
\draw[<-] (fit.north-|ot) coordinate (aux)--++(45:7mm) node[above]{$\mathbf h_{t-1}$};

\draw[<-] (fit.south-|ft) coordinate (aux)--++(-135:7mm) node[below]{$\mathbf x_t$};
\draw[<-] (fit.south-|ft) coordinate (aux)--++(-45:7mm) node[below]{$\mathbf h_{t-1}$};

\end{tikzpicture}
\caption[Architecture of LSTM cell]{The architecture of LSTM cell (adapted from \cite{graves2013generating}). }
\label{lstm_cell}	
\end{figure}
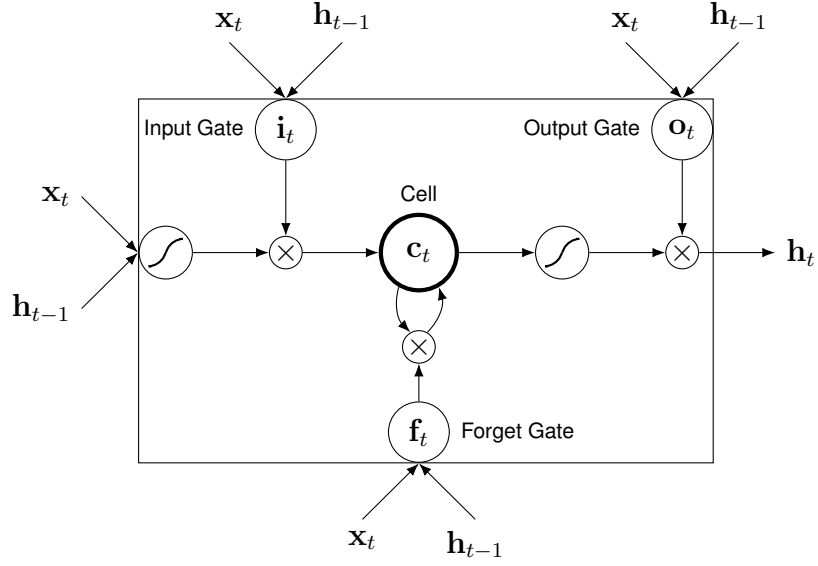

\label{lstm}
While in theory recurrent neural networks are simple and powerful models for sequence modelling, in practice it is challenging to train them well due to the difficulty of learning long-range dependencies. As described by \cite{bengio1994learning}, two of the major reasons for the difficulty are the problems of {\it exploding}  and  {\it vanishing} gradients when backpropagating errors across many timesteps. Exploding gradients refers to the situation where gradients increase exponentially during backpropagation, making learning diverge. Conversely, vanishing gradients happens when gradients decrease exponentially fast towards zero, making learning extremely slow or even stop. 

\cite{pascanu2013difficulty} provide a detailed mathematical derivation of the exact conditions under which the problems of vanishing and exploding gradients may occur. Here, we give a brief informal explanation. Consider a recurrent network with a single input node, a single output node, and a single recurrent hidden node. We pass an input at time one and calculate the loss $\mathcal{ L}(\hat{y}_\tau, y_\tau)$ at timestep $\tau$, with the assumption that input of zero is passed to the timesteps between one and $\tau$. The hidden state at time $\tau$ is calculated as 
\begin{align}
	h_\tau = \phi(w^{hh}(\phi(w^{hh} \cdots \phi(w^{xh}x + w^{hh}h_1)))).
\end{align}
Based on the chain rule, we calculate the gradient of $w^{hh}$ with respect to $\mathcal{ L}(\hat{y}_\tau, y_\tau)$ as
\begin{align}
\frac{\partial \mathcal{ L}(\hat{y}_\tau, y_\tau)}{\partial w^{hh}} = \sum_{t=1}^{\tau}	\frac{\partial \mathcal{ L}(\hat{y}_\tau, y_\tau)}{\partial \hat{y}_\tau} \frac{\partial \hat{y}_\tau}{\partial h_\tau} \frac{\partial h_\tau}{\partial h_t} \frac{\partial h_t}{\partial w^{hh}}. 
\end{align}
In particular, we expand the term $\frac{\partial h_\tau}{\partial h_t}$:
\begin{align}
\frac{\partial h_\tau}{\partial h_t} = \prod_{i=t+1}^{\tau} \frac{\partial h_i}{\partial h_{i-1}} = \prod_{i=t+1}^{\tau} w^{hh} \phi'(w^{hh}h_{i-1}).
\end{align}
Intuitively, if the value of $w^{hh}$ is very small, e.g. $(0 < w^{hh} < 1)$, then $\frac{\partial h_\tau}{\partial h_t}$ will be much smaller as $w^{hh}$ is multiplied by itself $(\tau - t)$ times. In this case, gradient vanishing will happen. On the contrary, if $w^{hh}$ is greater than a threshold, then it may cause gradient explosion.

The problem of exploding gradients for RNNs is relatively easy to alleviate using approaches such as truncated backpropagation through time (TBPTT) \citep{williams1990efficient} and gradient clipping \citep{mikolov2012statistical}. It is more difficult to circumvent the issue of vanishing gradients.  Solutions that have been proposed include adding skip connections through time \citep{lin1996learning,waibel1990readings} and applying leaky units \citep{jaeger2007optimization}. Among them, the Long Short-Term Memory (LSTM) model \citep{hochreiter1997long} is one of the most widely used and effective architecture. The key idea of the LSTM is to replace each ordinary node in the hidden layer of a standard RNN with a {\it memory cell}. Each memory cell contains a node with a self-recurrent connection with the fixed weight of one, ensuring that gradients flow smoothly through time. In addition, there are gating units controlling the reading, writing and resetting of the cells. 

We now describe the architecture of LSTMs in detail. Figure \ref{lstm_cell} illustrates an LSTM memory cell. The input to the memory cell is the activations from the input layer at the current timestep $\mathbf{x}_t$ and the activations from the hidden layer at the previous timestep $\mathbf{h}_{t-1}$. The input node takes the input $(\mathbf{x}_t, \mathbf{h}_{t-1})$ and calculates its own activation in the same way as the standard RNN, i.e. weighted sum is applied followed by an activation function. The core of the memory cell is the {\it cell state} denoted as $\mathbf{c}_t$, whose updates are guarded by {\it gates}.  A gate is a sigmoidal unit that also takes $(\mathbf{x}_t, \mathbf{h}_{t-1})$ as its input and outputs a value between zero and one. It controls the  flow of information in and out of another node. If the value of the gate is zero, then the flow is blocked. 
If the value of the gate is one, then all flow is passed through. There are three gates in a memory cell: an {\it input gate}, an {\it output gate}, and a {\it forget gate}. The input gate controls to what extent the input signal can alter the cell state. The output gate decides what part of the cell state  to output. Finally, the forget gate, which is introduced by \cite{gers2000learning}, determines how much memory to keep.

The following equations describe the LSTM formally:
\begin{align}
	\mathbf{g}_t &= \phi (\mathbf{W}^{gx} \mathbf{x}_t + \mathbf{W}^{gh}\mathbf{h}_{t-1} + \mathbf{b}^g) \\
	\mathbf{i}_t &= \sigma(\mathbf{W}^{ix}\mathbf{x}_t + \mathbf{W}^{ih}\mathbf{h}_{t-1} + \mathbf{b}^i) \\
	\mathbf{f}_t &= \sigma(\mathbf{W}^{fx}\mathbf{x}_t + \mathbf{W}^{fh}\mathbf{h}_{t-1} + \mathbf{b}^f) \\
	\mathbf{o}_t &= \sigma(\mathbf{W}^{ox}\mathbf{x}_t + \mathbf{W}^{oh}\mathbf{h}_{t-1} + \mathbf{b}^o) \\
	\mathbf{c}_t &= \mathbf{g}_t \odot \mathbf{i}_t + \mathbf{c}_{t-1} \odot \mathbf{f}_t \\
	\mathbf{h}_t &= \phi (\mathbf{c}_t) \odot \mathbf{o}_t,
\end{align}
where the operator $\odot$ denotes the element-wise operation, and $\mathbf{i}$, $\mathbf{f}$, and $\mathbf{o}$ denote the input gate, forget gate, and output gate, respectively. $\mathbf{W}$'s and $\mathbf{b}$'s are weight matrices and bias terms, respectively. Following \cite{zaremba2014recurrent}, we adopt $\tanh$ as the activation function $\phi$. 

There are several variations of this LSTM structure. The original form of LSTM only contains input and output gates. Forget gates, as described above, were added later by \cite{gers2000learning}, which have been proved useful in tasks that require the network to forget previous inputs. We can obtain the original form of LSTM by setting $\mathbf{f} = 1$. Another variation is to add peephole connections that allow gates to access the cell state directly. The Gated Recurrent Unit (GRU) is a popular alternative to LSTMs. It combines the forget and input gates into a single {\it update gate}. It also merges the cell state and hidden state, and makes some other changes. We refer the readers to the article \cite{cho2014learning} for a detailed description of a GRU. In this thesis, we only consider LSTMs with forget gates and without peephole connections.
\section{Recurrent Language Model}
\label{sec:rnnlm}
In the remainder of the chapter, we discuss two important applications of RNNs, the recurrent language model (RNNLM) and the encoder-decoder paradigm.\footnote{We use the term encoder-decoder or seq2seq model interchangeably throughout the thesis.} 

A language model (LM) is a probability distribution over sequences of words: $p(\boldsymbol{y})$.  It is a crucial component in text generation tasks such as machine translation and speech recognition, helping to check whether the output sequence is fluent. To calculate the joint probability $p(\boldsymbol{y})$, we decompose it into the probability of each successive word conditioned on the preceding words:
\begin{align}
	p(\boldsymbol{y}) = \prod_{j=1}^{J} p (y_j\ |\ \boldsymbol{y}_1^{j-1})
\end{align}
Traditionally, the probability $p(y_j\ |\ \boldsymbol{y}_1^{j-1})$ is approximated as $p(y_j\ |\ \boldsymbol{y}_{j-n+1}^{j-1})$ based on the Markovian assumption that the prediction of the next word depends only on the previous $n-1$ words. The size of the fixed context window $n - 1$ is usually set around 2 to 5. While $n$-gram models are simple, they cannot model long-range dependencies and the count-based variant do not generalise well to unseen $n$-grams.

\begin{figure}
	\centering
	\includegraphics[scale=0.70]{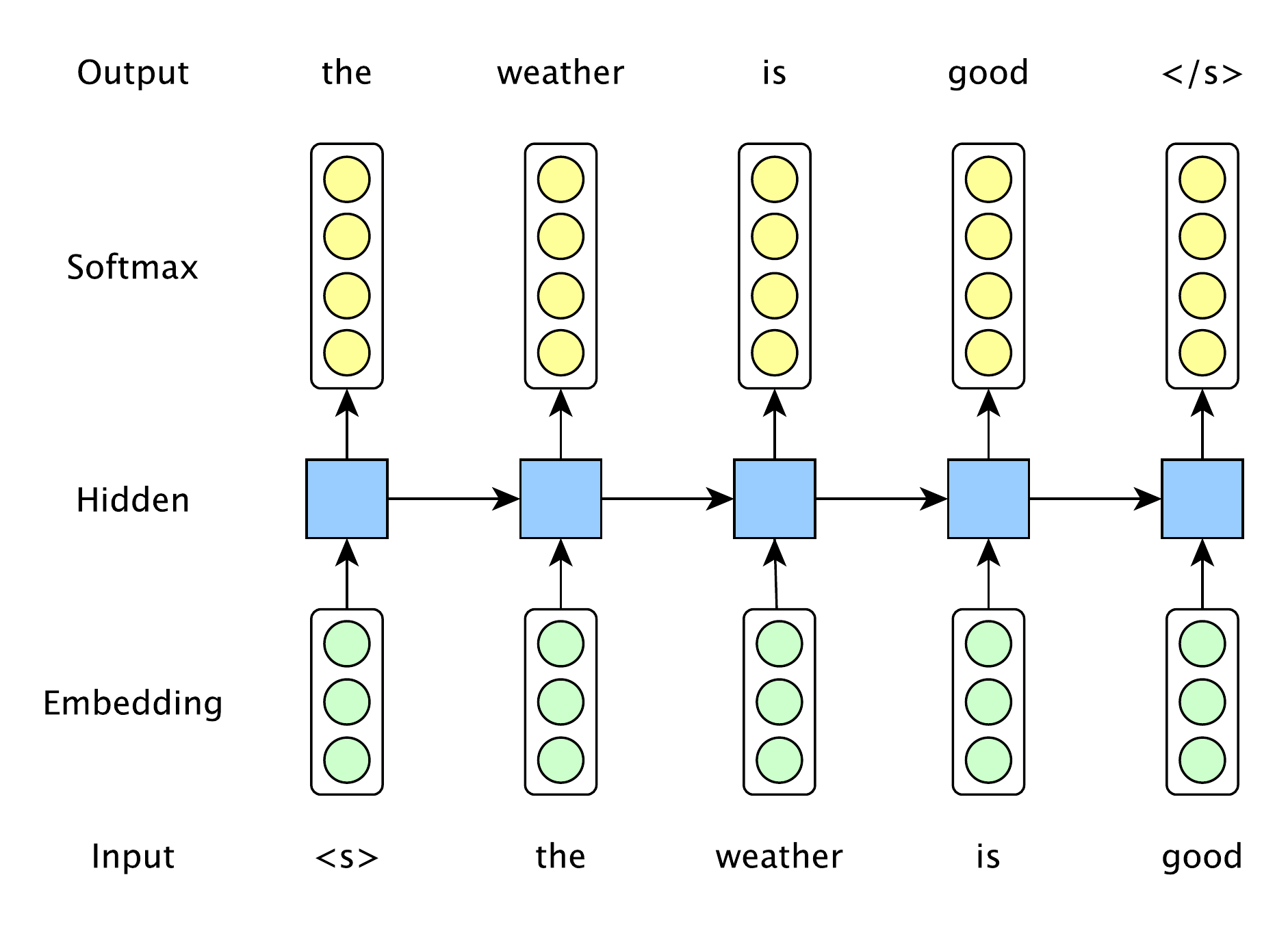}
	\caption[Architecture of Recurrent lanugage model]{The architecture of recurrent language model. }
	\label{lm_figure}
\end{figure}

The RNNLM \citep{mikolov2010recurrent} addresses both the aforementioned issues effectively.  In RNNLMs, an {\it embedding} matrix $\mathbf{W}^e \in \mathbb{R}^{d\times|V|}$ is maintained, which represents the distributed
feature vectors associated with each word in the vocabulary $V$. The matrix is learned together with the rest of parameters in the model. Starting from the special start-of-sentence symbol, for each input word $y_j$, the vector representation $\mathbf{y}_j$  is first retrieved by looking up the embedding matrix, and it is then fed into the input layer of the RNN (usually LSTM) at timestep $j$. Since the output sequence are discrete symbols, the $\softmax$ function is used at the output layer of the RNN. As illustrated in Figure \ref{lm_figure}, the output word generated from timestep $j-1$ is used as the input to timestep $j$. RNNLM has been one of the most successful applications of RNNs. It outperforms $n$-gram LMs by a significant margin, becoming the most commonly used language models.

\section{Sequence to Sequence Model}
\label{seq2seq}
The seq2seq model is a dominant architecture for processing sequence pairs.
First introduced for machine translation \citep{kalchbrenner2013recurrent,sutskever2014sequence,cho2014learning,bahdanau2014neural}, seq2seq models have also been successful in image caption generation \citep{vinyals2015show,xu2015show}, dialog systems  \citep{vinyals2015neural,shang2015neural} and speech recognition \citep{chorowski2014end,hannun2014deep}. 
In this section, we provide the necessary background of the vanilla seq2seq model, and the seq2seq model with attention. This background is highly related to our work described in Chapters \ref{ch:ssnt} and \ref{ch:noisy_channel}.
\subsection{Architecture}
The seq2seq model aims to model the conditional probability directly:
\begin{align}
p(\boldsymbol{y}\ |\ \boldsymbol{x}) = \prod_{j=1}^Jp(y_j\ |\ \boldsymbol{y}_1^{j-1}, \boldsymbol{x} ),
\end{align}
meaning the probability of generating the output sequence $\boldsymbol{y}$ given the input sequence $\boldsymbol{x}$.
The model is composed of an encoder and a decoder. The encoder reads the input sequence one token at a time to obtain a large fixed-size vector. The decoder then generates an output sequence from the vector. Empirically, according to different tasks, the design of the encoder and decoder varies in terms of  the types of neural networks for modelling data, the types of recurrent units for RNNs, and the depth of the networks etc. For instance, for image caption generation, CNNs are employed for encoding images. We usually use RNNs as the encoder if the input sequence is text, although \cite{kalchbrenner2013recurrent} use CNNs to encode source sentences in machine translation. The choice of networks for the decoder is less flexible, since a unidirectional RNN is more suitable for generating sequential data compared to others. 

\begin{figure}
	\centering
	\includegraphics[scale=0.70]{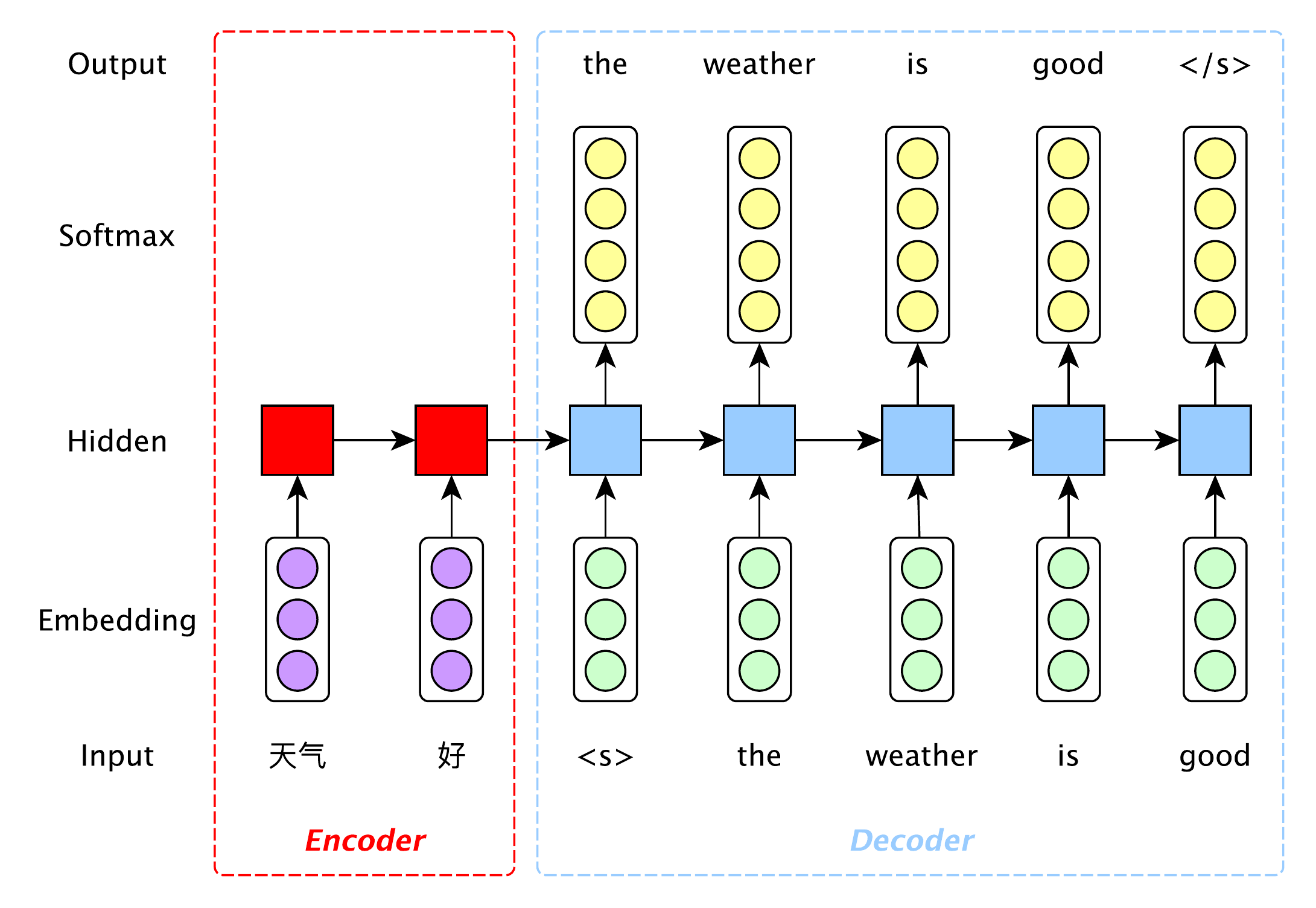}
	\caption[Architecture of the sequence to sequence model]{The architecture of the sequence to sequence model. The encoder RNN encodes the source sequence (Chinese sentence) into a vector, and the decoder RNN takes the said vector as its initial hidden state and generates output (English word) at each timestep. The word embeddings for the source and target sentences may be shared or separated.}
	\label{fig:seq2seq}
\end{figure}

Figure \ref{fig:seq2seq} provides an illustration of the seq2seq model for the Chinese-to-English machine translation. In this example, both the encoder and decoder are modelled by RNNs. Similar to the RNNLM, the representation of each word is first retrieved from the embedding matrix. These word vectors are then fed into RNNs. Unlike RNNLM, usually two separate embedding matrices are maintained for the source and target vocabularies. The RNN encoder takes the source sentence as input without generating output. When the end of the source sentence is reached, a special symbol that indicates the start of the output sentence is sent to the RNN decoder. The RNN decoder takes the last hidden state of the RNN encoder as its initial hidden state,  emitting output at each timestep. The decoder is essentially an RNNLM except that it is conditioned on the input sequence.

During training, the true sources and targets are fed into the encoder and decoder, respectively, and the loss is backpropagated from the outputs of the decoder across the entire seq2seq model. The model is trained by minimising the negative log likelihood of the sequence pairs in the training data $S$:
\begin{align}
	\mathcal{ L} = -\sum_{(\boldsymbol{x}, \boldsymbol{y}) \in S} \log p(\boldsymbol{y}\ |\ \boldsymbol{x} ).
\end{align} 
At the inference time, the approximately most likely output sequence is generated via greedy decoding or a beam search algorithm. In greedy decoding the most likely word predicted at the previous timestep is fed as the input to the next timestep and the decoding is finished when the end-of-the-sentence symbol is generated. Instead of predicting the token with the best score, beam search keeps track of $k$ hypotheses, where $k$ refers to beam size. At each timestep we extend the existing $k$ hypotheses by appending each of them with a new token chosen from the vocabulary $V$. This results in a total of $k|V|$ new hypotheses, from which $k$ best ones are kept in the beam of the current timestep for further extension. Once every hypothesis reaches the end-of-the-sentence token, the hypothesis with the highest score is returned as the output sequence.

\subsection{Attention Mechanism}
\label{seq2seq_att}
\begin{figure}
	\centering
	\includegraphics[scale=0.70]{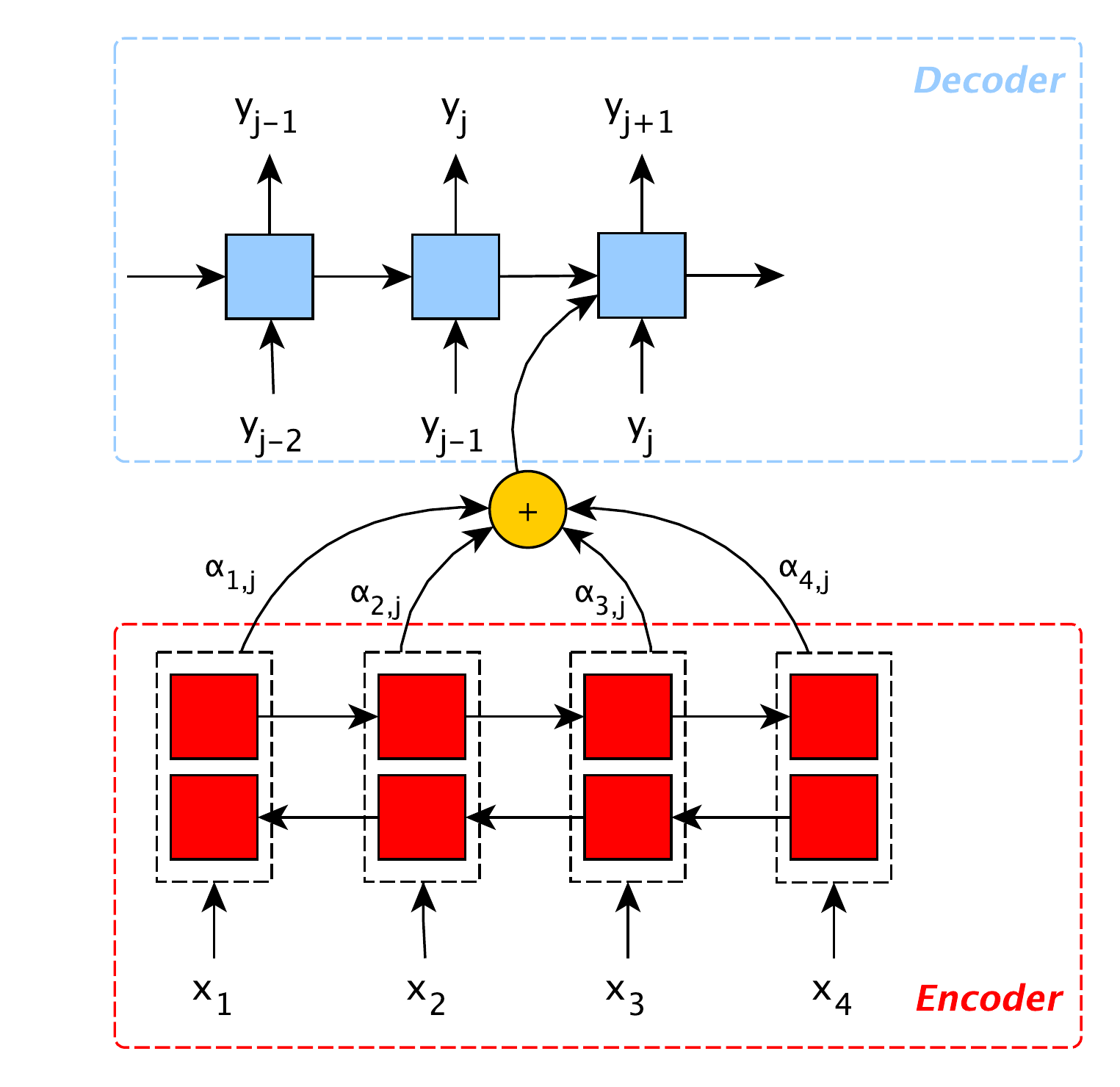}
	\caption[Architecture of sequence to sequence model with attention]{The architecture of the sequence to sequence model with attention (adapted from \cite{bahdanau2014neural}). The encoder is a bidirectional RNN. At timestep $j$, the context vector, which is obtained from the hidden states of the encoder and the previous hidden state of the decoder, is used to predict the output token and is fed into the next recurrent state.}
	\label{fig:seq2seq_att}
\end{figure}

In the vanilla seq2seq model, the encoder compresses the entire input sequence into a fixed-size vector, and the decoder then generates the output sequence based only on this vector. Intuitively, this is problematic since the encoded vector may not be able to capture all the necessary information of the input sequence, especially for long sequences. This will in turn result in output generated in low quality.
 It has been revealed that in machine translation, the quality of translations produced by seq2seq models deteriorates dramatically as the lengths of the source sentence increases \citep{cho2014properties}. In practice, a big network has to be used  in order to achieve good performance, which requires a large amount of computational resources.

To address the aforementioned weakness, \cite{bahdanau2014neural} incorporate an attention mechanism to the encoder-decoder framework. When generating an output token, their model looks at each input position, and decides which part is the most relevant for the current prediction. It then predicts the output based on the {\it context vectors} associated with the relevant part of the input and the previously generated output tokens. As demonstrated by the authors, the attentional seq2seq model is much better at processing long sequences compared to the vanilla seq2seq model. This is mainly because as opposed to encoding everything into a fixed-size vector, the input sequence is stored as a dynamic sized matrix, parts of which are selectively used by the decoder.

We now describe the attentional seq2seq model formally (Figure \ref{fig:seq2seq_att}). The encoder is a BiRNN, where we obtain two sequences of hidden state vectors $(\mathbf{h}^{\rightarrow}_1, \mathbf{h}^{\rightarrow}_2, \dots, \mathbf{h}^{\rightarrow}_I)$ and $(\mathbf{h}^{\leftarrow}_1, \mathbf{h}^{\leftarrow}_2, \dots, \mathbf{h}^{\leftarrow}_I)$ by feeding the input sequence into a forward RNN and a backward RNN, respectively. We concatenate the hidden state vectors from both RNNs at each timestep to build a sequence of {\it annotation vectors} $(\mathbf{h}_1, \dots, \mathbf{h}_I)$:
\begin{align}
	\mathbf{h}_i = [\mathbf{h}^{\leftarrow}_i\ ;\  \mathbf{h}^{\rightarrow}_i].
\end{align}
Each annotation vector encodes the information surrounding the $i$-th input token. The decoder is a unidirectional RNN with an attention mechanism that 
evaluates to what extent each annotation vector contributes to the output prediction at each timestep. The attention score $\alpha_{i,j} \in [0, 1]$ of the $i$-th annotation vector for the $j$-th output token is calculated by feeding each annotation vector $\mathbf{h}_i$ and the previous hidden state of the decoder $\mathbf{s}_{j-1}$ into a feedforward neural network, followed by a $\softmax$ function:
\begin{align}
	e_{ij} &= \text{MLP}(\mathbf{s}_{j-1}, \mathbf{h}_i) \\
	\alpha_{ij} &= \frac{\exp (e_{ij})}{\sum_{i'=1}^{I} \exp e_{i'j}}
\end{align}
We use these scores to derive the context vector $\mathbf{c}_j$ of the $j$-th output token:
\begin{align}
	\mathbf{c}_j = \sum_{i=1}^{I} \alpha_{ij}\mathbf{h}_i.
\end{align}
The context vector contributes to both the emission of the output token and the calculation of the next hidden state of the RNN decoder:
\begin{align}
	\mathbf{s}_j &= \text{RNN}(\mathbf{s}_{j-1}, \mathbf{y}_{j-1}, \mathbf{c}_j) \\
	\mathbf{u}_j & = \tanh (\mathbf{W}^{c}[\mathbf{s}_j \ ;\  \mathbf{c}_j] + \mathbf{b}^c) \\
	p(y_j\ |\ \boldsymbol{y}_1^{j-1} ) & \sim \softmax(\mathbf{W}^s\mathbf{u}_j + \mathbf{b}^s),	
\end{align}
where $\mathbf{y}_{j-1}$ denotes the word representation corresponding to the $(j-1)$-th output token. $\mathbf{W}^{c}$ and $\mathbf{W}^{s}$ are weight matrices to be learned during training and $\mathbf{b}^c$ and $\mathbf{b}^s$ are biases.

The attention mechanism that we have just described is called {\it soft} attention. Because it is a fully differentiable deterministic mechanism that can be incorporated into an existing system, it is trivial to train the model end-to-end  using standard backpropagation.
A common criticism of soft attention is that it is inefficient to calculate a relative score for every input position when predicting each output token \citep{DBLP:conf/emnlp/LuongPM15,DBLP:conf/icml/RaffelLLWE17}. Another criticism is that the context vector summarises the information about the whole input sequence (due to the weighted sum operation), and it may contain noise from irrelevant components, resulting in inaccurate predictions \cite{DBLP:conf/icml/RaffelLLWE17}.

An alternative attention mechanism is {\it hard} attention \citep{xu2015show}, where a particular hidden state of the input sequence is sampled to attend to, instead of using all the hidden states. Hard attention has a strong bias, and if this bias is a good match to the true generating distribution, then it performs better and more efficiently than soft attention.

\section{Summary}
To summarise, in this chapter we have reviewed various types of neural networks and learning algorithms. We have also described two important models based on recurrent neural networks: the recurrent language model and the encoder-decoder paradigm. These models are essential for our work to be described in the remainder of this thesis. Chapter \ref{ch:sentence_model} will investigate the effectiveness of different types of neural networks on modelling sentences. We evaluate them on a seq2seq mapping task named question answer selection. We will leverage recurrent language model in our neural noisy channel model in Chapter \ref{ch:noisy_channel}. In Chapter  \ref{ch:ssnt}, we propose a transduction model that aims to address the bottleneck of the vanilla seq2seq model. Our proposed model is a kind of hard attention mechanism.

%% file: chapter_4.tex
\chapter{The Role of Distributed Sentence Models}
\label{ch:sentence_model}

\begin{chapterabstract}
This chapter presents our first series of work on applying neural networks to seq2seq mapping problems. We focus on question answer selection,
which is the task of identifying sentences that contain the
answer to a given question. This is an important problem in its own right as
well as in the larger context of open domain
question answering. We propose a novel approach to solving this task via
means of distributed representations and learn to match questions with answers by considering their semantic encoding. 
We examine distributed sentence models based on various types of neural networks, aiming to investigate their effectiveness on the seq2seq mapping task.
Experiments on two standard benchmark datasets from TREC and Wikipedia show that the more sophisticated models
work better than the simpler models. In addition to the findings, our model
achieves state of the art performance on the answer sentence selection task.\let\thefootnote\relax\footnote{The material in this chapter was originally presented in \cite{Yu:2014} and \cite{miao:2016}.}
\addtocounter{footnote}{-1}\let\thefootnote\svthefootnote  
\end{chapterabstract}

\section{Introduction}
Having reviewed the main concepts and algorithms of neural networks in Chapter \ref{ch:nn}, in this chapter we describe our first step towards applying neural networks to seq2seq mapping problems. We focus on the task of answer sentence selection (a recognition task), which is the task of selecting a sentence that contains the information required to answer a given question from a set of candidates obtained via some information extraction system. 

Like other recognition tasks of seq2seq mappings, such as paraphrase detection and textual entailment, it is assumed that the relevance between the input sentence pairs (i.e. the answer sentences and the questions) is determined by their semantic similarity. 
Prior work in this field mainly attempted this via syntactic matching of parse trees. This can be achieved with generative models that syntactically transform answers to questions
\citep{wang2007jeopardy,wang2010probabilistic}. Another option is discriminative models over features produced from minimal edit sequences between dependency parse trees \citep{DBLP:conf/naacl/HeilmanS10a,yao2013answer}.
Beyond syntactic information, some prior work has also included semantic
features from resources such as WordNet, with the previous state-of-the-art
model for this task relying on a variety of such lexical semantic resources
\citep
{yih2013question}.

While empirically the inclusion of a large amount of semantic information has
been shown to improve performance, prior approaches rely on a significant amount of feature engineering and require expensive semantic resources
which may be difficult to obtain.

At the same time, distributed approaches that represent words as vectors are more effective in capturing the correlations between words or phrases. In tasks such as sentiment analysis and document classification, neural networks have been leveraged to compose word representations into sentence representations \citep{Hermann:2014:ACLphil,DBLP:conf/acl/KalchbrennerGB14}. Optimal sentence representations can be learned for a given task by training the entire model end to end.

In this work, we apply neural network-based sentence models to the task of answer sentence selection. We build four different distributed
sentence models: (1) a bag-of-words model, (2) a bigram model
based on a convolutional neural network, (3) a model based on a recurrent neural network, and (4) an attentional model.  Assuming a set of pre-trained semantic
word embeddings, we train a supervised model to learn a semantic
matching between question and answer pairs. The underlying distributed models provide the semantic information necessary for this matching function. We also present an enhanced version of this model, which combines the signal of the distributed matching algorithm with two simple word matching features. 

Our main contributions in this chapter are: (1) we show that our model is superior to the traditional approaches based on feature-engineering on the answer sentence selection task --- our model achieves state-of-the-art results on two benchmark datasets without the involvement of human effort or any external linguistic resources; (2), we make experimental comparison between various network architectures on this task, and our findings can be applied to other seq2seq mapping problems.

This chapter is organised as follows: it starts with the background of the answer sentence selection task and compositional distributional semantics; this is followed by model description and experiments; finally, results will be presented and discussed.
\section{Background}
In this section, we first introduce the task of answer sentence selection and discuss the previous work on this task. Subsequently, we review the composition methods for distributional semantics.
\subsection{Answer Sentence Selection}
Question answering can broadly be divided into two categories. One approach
focuses on semantic parsing, where answers are retrieved by turning a question
into a database query and subsequently applying that query to an existing
knowledge base. The other category is open domain question answering, which is
more closely related to the field of information retrieval.

Open domain question answering requires a number of intermediate steps. For
instance, to answer a question such as {\it ``Who wrote the book Harry Potter?''}, a
system would first identify the question type and retrieve relevant documents.
Subsequently, within the retrieved documents, a sentence containing the answer
is selected, and finally the answer ({\it J.K. Rowling}) itself is extracted
from the relevant sentence.
In this chapter, we focus on answer
sentence selection, the task that selects the correct sentences answering a
factual question from a set of candidate sentences. Beyond its role in open
domain question answering, answer sentence selection is also a stand-alone task
with applications in knowledge base construction and information extraction. A feature of this task is that the correct sentence may not
answer the question directly and perhaps it also contains extraneous information,
for example:

\begin{enumerate}
	\item[]\textbf{Q:} When did Amtrak begin operations?
	\item[]\textbf{A:} Amtrak has not turned a profit since it was founded in 1971.
\end{enumerate}

Clearly, answer sentence selection requires both semantic and syntactic
information in order to establish both what information the question seeks to
answer, as well as whether a given candidate contains the required information,
with current state-of-the-art approaches mostly focusing on syntactic matching
between questions and answers. Following the idea that questions can be generated from correct answers by
loose syntactic transformations, \cite{wang2007jeopardy} built a
generative model to match the dependency trees of question answer pairs based on the soft alignment of a quasi-synchronous grammar \citep{smith2006quasi}.
\cite{wang2010probabilistic} proposed another probabilistic
model based on Conditional Random Fields, which models alignment as a set of
tree-edit operations of dependency trees.
\cite{DBLP:conf/naacl/HeilmanS10a} used a tree kernel as a heuristic to search
for the minimal edit sequences between parse trees. Features extracted from
these sequences are then fed into a logistic regression classifier to select the
best candidate.
More recently, \cite{yao2013answer} extended Heilman and Smith's
approach with the difference that they used dynamic programming to find the
optimal tree edit sequences. In addition, they added semantic features obtained from WordNet.
Although some of these approaches use
WordNet relations (e.g. synonym, antonym, hypernym) as explicit features, the
focus of all of this work is primarily on syntactic information
\citep{yih2013question}.

Unlike previous work, \cite{yih2013question} applied rich lexical
semantics to their state-of-the-art QA matching models. These models match
the semantic relations of aligned words in QA pairs by using a combination of
lexical semantic resources such as WordNet with distributed representations
for capturing semantic similarity. This approach results in a series of features
for sentence pairs, which are then fed into a conventional classifier. A
variation of this idea can also be found in
\cite{severynautomatic}, who used an SVM with tree kernels to automatically learn
features from shallow parse trees rather than relying on external resources,
sacrificing semantic information for model simplicity. We combine these
two approaches by proposing a semantically rich
model without the need for feature engineering or extensive human-annotated 
external resources.

\subsection{Compositional Distributional Semantics}
Compared with words in string
form or logical form, distributed representations
of words can capture latent semantic information and thereby exploit
similarities between words. This enables distributed representations to overcome
sparsity problems encountered in atomic representations and further provides
information about how words are semantically related to each other. These
properties have made distributed representations become increasingly
popular in natural language processing.
They have been proved to be successful in applications such
as relation extraction \citep{Grefenstette:1994:EAT:527911}, word
sense disambiguation \citep{DBLP:conf/acl/McCarthyKWC04}, and the measurement of semantic word similarity \citep{mikolov2013efficient}.
Vector representations for words can be obtained in a number of ways, with
many approaches exploiting distributed information from large corpora, for
instance by counting the frequencies of contextual words around a given token.
An alternative method for learning distributed representations comes in the form
of neural language models, where the link to distributed data is somewhat
more obscure
\citep{DBLP:conf/icml/CollobertW08,DBLP:conf/nips/BengioDV00}. A nice side-effect
of the way in which neural language models learn word embeddings is that these
vectors implicitly contain both syntactic and semantic information.

For a number of tasks, vector representations of single words are not sufficient. Instead, distributed representations of phrases and sentences are required in order to provide a deeper language understanding that allows us to address more complex tasks in NLP such as translation, sentiment analysis or information extraction. As sparsity prevents us from directly learning distributed representations at the phrase level, various models of vector composition have been proposed that circumvent this problem by learning higher level representations based on low-level (e.g. word-level) representations. Popular ideas for this include exploiting category theory \citep{ClarkCoeckeSadrzadeh2008}, using parse trees in conjunction with recursive autoencoders \citep{DBLP:conf/emnlp/SocherHMN12,DBLP:conf/acl/HermannB13}, convolutional neural networks \citep{DBLP:conf/icml/CollobertW08,DBLP:conf/acl/KalchbrennerGB14,DBLP:conf/emnlp/Kim14}, and LSTMs \citep{DBLP:conf/emnlp/BowmanAPM15,DBLP:conf/acl/TaiSM15}. 

We compare the performance of four neural network-based models that produce distributed representations of sentence meaning on the answer sentence selection task. The details of these models are described formally in the next section.

\section{Model Description}\label{model}
\begin{figure}
	\begin{center}
		\includegraphics[scale = 0.6]{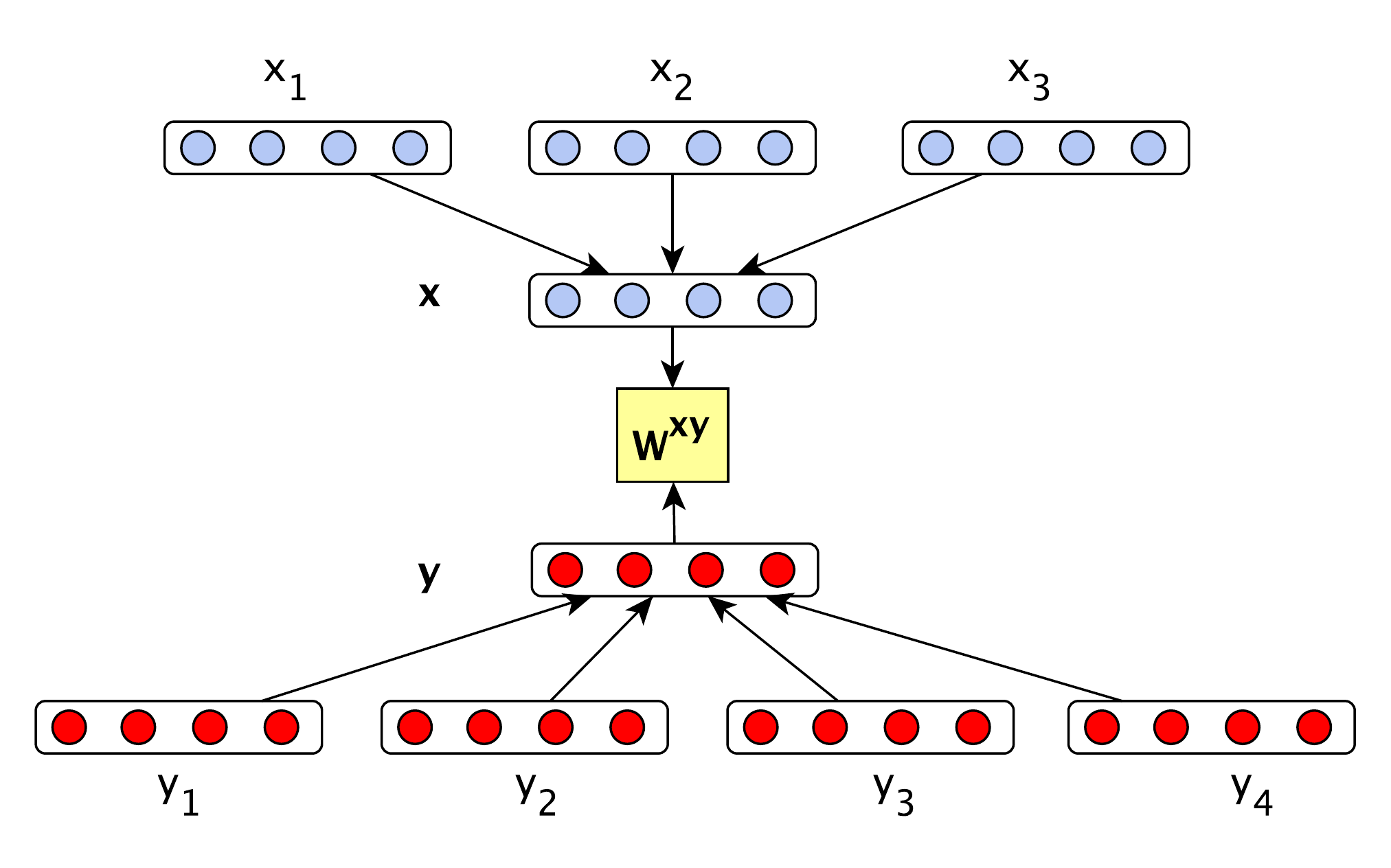}
	\end{center}
	\caption[Model for QA matching]{The model for QA matching.  A bilinear model is employed to calculate the similarity between the vectors for the question and answer. $\mathbf{W}^{xy}$ is the parameter matrix.  Vector representations of the question $\mathbf{x}$ and answer $\mathbf{y}$ are obtained from various distributed sentence models.}
	\label{fig:qa_math}
\end{figure}

Answer sentence selection can be viewed as a binary classification problem.
Assume a set of questions $\mathcal{X}$, where each question $\boldsymbol{x} \in \mathcal{X}$ is
associated with a list of answer sentences $\{\boldsymbol{y}_{1}, \boldsymbol{y}_{2},
\cdots, \boldsymbol{y}_{m}\}$, together with their judgements $\{l_{1}, l_{2},
\cdots, l_{m}\}$, where $l_{i} = 1$ if the answer is correct and $l_{i} =
0$ otherwise. While this could be approached as a multi-labelling task, we
simply treat each data point as a triple $(\boldsymbol{x}, \boldsymbol{y}_{i},
l_{i})$. Thus, our task is to learn a classifier over these triples so
that it can predict the judgements of any additional QA pairs.

Our solution to this problem assumes that correct answers have high
semantic similarity to questions. We model questions and answers as vectors, and evaluate the relatedness of each QA pair in a shared vector space. Formally, following \cite{DBLP:journals/corr/BordesWU14}, given the vector representations of a question $\mathbf{x}$ and an answer $\mathbf{y}$ (both in $\mathbb{R}^{d}$), the probability of the answer being correct is
\begin{equation}
p(l = 1\ |\ \mathbf{x}, \mathbf{y}) = \sigma(\mathbf{x}^T\ \mathbf{W}^{xy}\ \mathbf{y}
+ b),\\[0.5em]
\label{eq:qa_match}
\end{equation}
where the bias term $b$ and the transformation matrix $\mathbf{W}^{xy} \in
\mathbb{R}^{d \times d}$ are model parameters.  This formulation can intuitively
be understood as an expression of the generative approach to open domain
question answering: given a candidate answer sentence, we `generate' a question
through the transformation $\mathbf{x}' = \mathbf{W}^{xy}\ \mathbf{y}$, and
then measure the similarity of the generated question $\mathbf{x}'$ and the
given question $\mathbf{x}$ by their dot product. The sigmoid function
squashes the similarity scores to a probability between $0$ and $1$. The model
is trained by minimising the penalised cross entropy of all labelled data QA pairs:
\begin{equation}
\begin{split}
\mathcal{L} &= -\log \prod_n p(l_n\ |\ \mathbf{x}_n, \mathbf{y}_n) + \frac{\lambda}{2}\|\boldsymbol{\theta}\|^2_F\\
& = -\sum_n l_n \log \sigma(\mathbf{x}_n^T\ \mathbf{W}^{xy}\ \mathbf{y}_n + b) + (1 - l_n)\log (1 - \sigma(\mathbf{x}_n^T\ \mathbf{W}^{xy}\ \mathbf{y}_n + b)) + \frac{\lambda}{2}\|\boldsymbol{\theta}\|^2_F,
\end{split}
\end{equation}
where $\|\boldsymbol{\theta}\|^2_F$ is the Frobenius norm of $\boldsymbol{\theta}$, and $\boldsymbol{\theta}$ includes
$\{\mathbf{W}^{xy},b\}$ as well as any parameters introduced in the sentence
composition  model. Next we describe four methods employed in this work for
projecting sentences into vector space representations.

\subsection{Bag-of-Words Model}

Given word embeddings, the bag-of-words model generates the vector
representation of a sentence by summing over the embeddings of all words in the
sentence --- having previously removed stop words\footnote{a, also, the, an, to, is, am, are, and, or, any, as, at, be, but, by, can, cannot, cant, does, do, did, does, doing, done, each, for, from, has, have, he, her, him, himself, his, i, in, into, inward, is, it, its, itself, me, mine, my, myself, no, not, of, off, on, or, our, ours, ourselves, self, she, should, that, the, their, theirs, them, themselves, these, they, this, those, though, to, too, was, we, were, will, with, would, you, your, yours, yourself, yourselves, /, ?, !, ``, `,  '', ., , , :, ;, -, \_, 's, 're, -LRB-, -RRB-.} from the input. The vector is
then normalised by the length of the sentence.
\begin{equation}
\mathbf{x} = \frac{1}{|\mathbf{x}|}\sum_{i=1}^{|\mathbf{x}|} \mathbf{x}_i.
\end{equation}

\subsection{Bigram Model}

Due to its inability to account for word ordering and other structural
information, the simple bag-of-words model proposed above is unable to capture
more complex semantics of a sentence. To address this issue, we also evaluate a
sentence model based on a convolutional neural network (CNN).

The advantage of this composition model is that it is sensitive to word ordering
and is able to capture features of $n$-grams independent of their positions in
the sentences. Further, the convolutional network can learn to correspond to the
internal syntactic structure of sentences, removing reliance on external
resources such as parse trees \citep{DBLP:conf/acl/KalchbrennerGB14}. 
\begin{figure}
	\begin{center}
		\includegraphics[scale = 0.8]{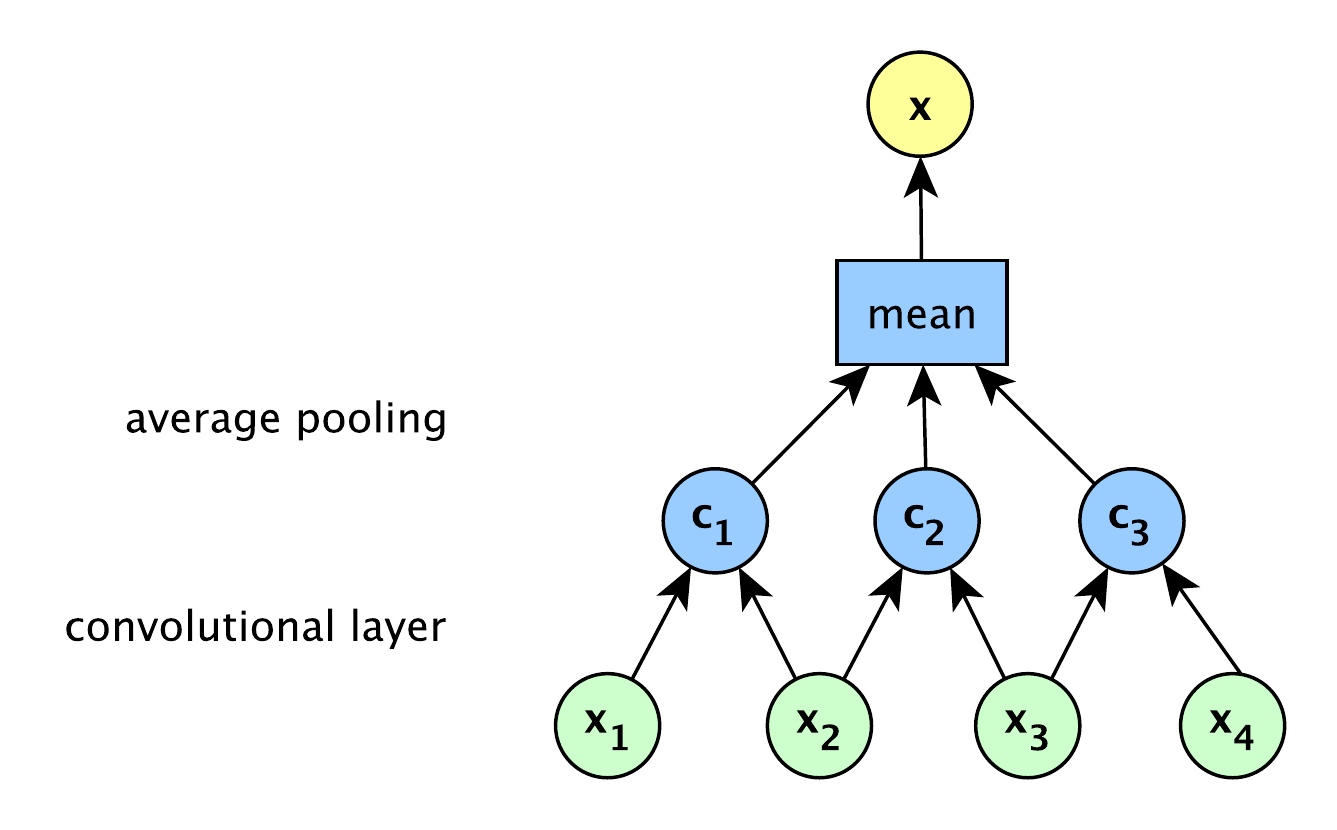}
	\end{center}
	\caption[Architecture of a one-dimensional convolutional neural network]{The architecture of one-dimensional convolutional neural network. A sequence of vector representation of words are fed into a convolutional layer, followed by an average pooling layer.}
	\label{ccn}
\end{figure}
Figure \ref{ccn} illustrates the architecture of the CNN-based sentence model in one dimension. We use bigrams here with one convolutional layer and one pooling
layer. The convolutional vector $\mathbf{w} \in \mathbb{R}^2$, which is shared
by all bigrams, projects every bigram into a feature value $\mathbf{c}_i$,
computed as follows:
\begin{equation}
\mathbf{c}_i = \tanh(\mathbf{w}\cdot\mathbf{x}_{i:i+1} + b).
\label{convolve}
\end{equation}
Since we would like to capture the meaning of the full sentence, we then use
average pooling to combine all bigram features. This produces a full-sentence
representation of the same dimensionality as the initial word embeddings. In
practice, of course, these representations are not a single value, but
$d$-dimensional vectors, and hence $\mathbf{w}$ and each bigram are matrices.
Thus, Equation \ref{convolve} is calculated for each row of $\mathbf{w}$ and the
corresponding row of $\mathbf{x}$. Similarly, average pooling is performed
across each row of the convolved matrix. Formally, our bigram model is
\begin{equation}
\mathbf{x} = \frac{1}{|\mathbf{x}|-1} \sum_{i=1}^{|\mathbf{x}|-1} \tanh (\mathbf{W}^L\ \mathbf{x}_{i} +
\mathbf{W}^R\ \mathbf{x}_{i+1} + \mathbf{b}),
\end{equation}
where $\mathbf{x}_i$ is the vector of the $i$-th word in the sentence, and
$\mathbf{x}$ is the vector representation of the sentence. Both vectors are in
$\mathbb{R}^{d}$. $\mathbf{W}^L$ and $\mathbf{W}^R$ are model
parameters in $\mathbb{R}^{d \times d}$ and $\mathbf{b}$ is the bias. From a linguistic perspective, these
parameters can be considered as informing the ordering of individual word pairs.

\subsection{LSTM}
Since long-range dependencies are common in questions, it should be more effective to use LSTMs to compose word vectors into sentence vectors. In this case,
 word vectors are fed into the LSTM at each timestep, and the last hidden state of the LSTM is taken as the vector representation of the sentence.

\subsection{LSTM with Attention}
\begin{figure}
	\begin{center}
		\includegraphics[scale = 0.7]{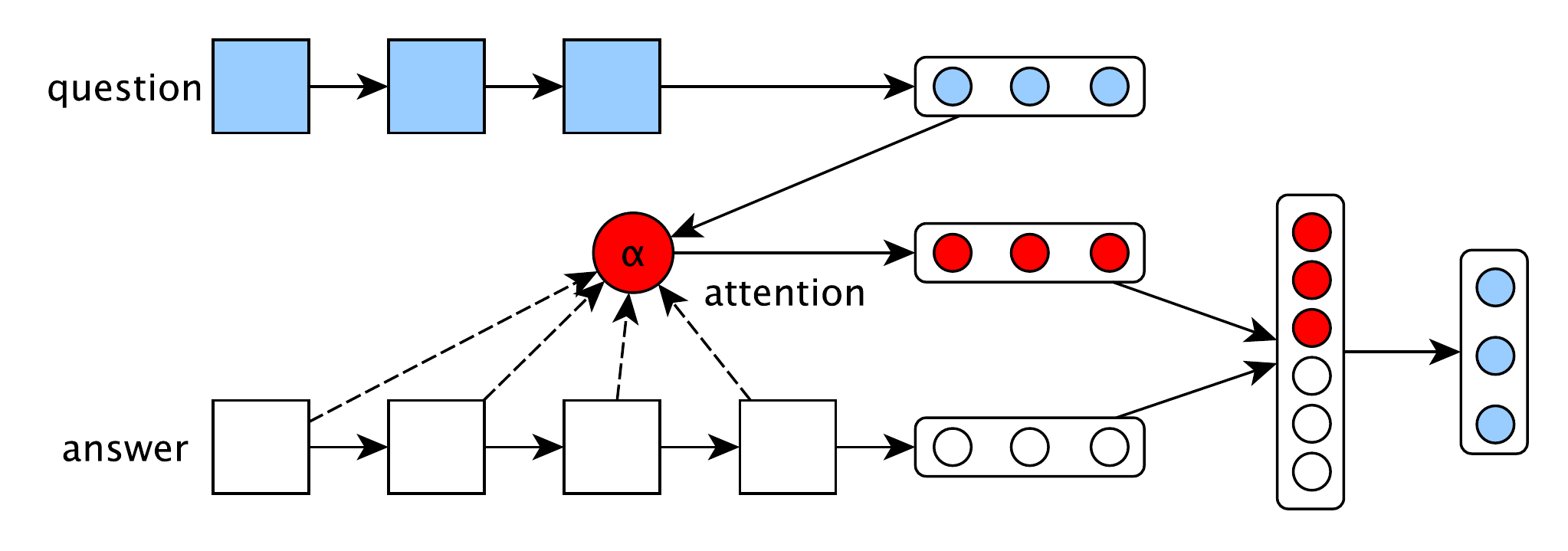}
	\end{center}
	\caption[Attentional model for QA matching]{The attentional model for QA matching. First, the context vector $\mathbf{c}$ (in red) is calculated from the hidden states of the answer LSTM $\{\mathbf{s}_1, \mathbf{s}_2, \dots, \mathbf{s}_{|\boldsymbol{y}|}\}$ and the last hidden state of the question LSTM $\mathbf{h}_{|\boldsymbol{x}|}$.  The context vector is then combined with $\mathbf{s}_{|\boldsymbol{y}|}$, resulting in the vector $\mathbf{y}$.  Finally, the vectors $\mathbf{h}_{|\boldsymbol{x}|}$ and $\mathbf{y}$ are sent to Equation \ref{eq:qa_match} to calculate their relevance score.}
	\label{qa_att}
\end{figure}

Inspired by \cite{rocktaschel2015reasoning}, we incorporate a deterministic attention mechanism in the LSTM model.  We expect the model to attend to the key answer phrases, which provides stronger evidence for the relatedness between the question and the candidate answer sentence. Let $\{\mathbf{h}_1, \mathbf{h}_2, \dots, \mathbf{h}_{|\boldsymbol{x}|}\}$ and $\{\mathbf{s}_1, \mathbf{s}_2, \dots, \mathbf{s}_{|\boldsymbol{y}|}\}$ denote the list of hidden states corresponding to the LSTMs processing the question and answer, respectively. The attentional QA matching model is defined as:
\begin{align}
\alpha_i & \propto \exp(\mathbf{W}^{\alpha} \tanh(\mathbf{W}^h\mathbf{h}_{|\boldsymbol{x}|}  + \mathbf{W}^s\mathbf{s}_i)) \\
\mathbf{c} &= \sum_{i=1}^{|\boldsymbol{y}|} \mathbf{s}_i \alpha_i \\
\mathbf{y} &= \tanh (\mathbf{W}^c \mathbf{c} + \mathbf{W}^y \mathbf{s}_{|\boldsymbol{y}|}) \\
\mathbf{x} &= \mathbf{\mathbf{h}_{|\boldsymbol{x}|}}.
\end{align}
where $\alpha_i$ is the normalised attention score at answer token $i$, and the context vector $\mathbf{c}$ is the weighted sum of all the state outputs $\mathbf{s}_i$. Different from the attentional seq2seq model introduced in Chapter \ref{ch:nn}, which calculates attention scores when generating every output token, here the model calculates attention scores only at the last timestep of the question LSTM.  The vector representation of the answer is obtained by combining the last hidden state of the answer LSTM $\mathbf{s}_{|\boldsymbol{y}|}$ with the context vector. The vectors $\mathbf{x}$ and $\mathbf{y}$ are used for calculating the relevance score in Equation \ref{eq:qa_match}. 

\section{Experiments}
We evaluated the four models presented in the previous section on two standard answer selection datasets. We briefly introduce the datasets before describing our experimental setup. Finally, we report our results and compare them with previous work.

\subsection{TREC Answer Selection Dataset}

The answer sentence selection dataset contains a set of factoid questions, with
a list of answer sentences corresponding to each question. 
\cite{wang2007jeopardy} created this dataset from Text REtrieval Conference
(TREC) QA track ($8$-$13$) data, with candidate answers automatically selected
from each question's document pool. Answer candidates are chosen using a
combination of overlapping non-stop word counts and pattern matching.
Subsequently, the correctness of the candidate answers are judged manually for
parts of the dataset. Table~\ref{data-table} summarises the answer selection dataset, and describes
the train/dev/test split of the data. The training set is noisy as parts of answers are labelled automatically using pattern matching.

The task is to rank the candidate answers based on their relatedness to the
question, and is thus measured in Mean Average Precision (MAP) and Mean
Reciprocal Rank (MRR), which are standard metrics in Information Retrieval and
Question Answering. Whereas MRR measures the rank of any correct answer,  
MAP examines the ranks of all the correct answers. In general, MRR is slightly higher than MAP on the same list of ranked outputs, except that they are the
same in the case where each question has exactly one correct answer.
The scores are calculated using the official
\texttt{trec\_eval} evaluation scripts.
\begin{table}[t]\centering
	\begin{tabular}{@{}llrrrl@{}}
		\toprule
		Source & Data & \# Questions & \# QA Pairs & \% Correct & Judgement
		\\
		\midrule
		\multirow{3}{*}{TREC QA}& Train  & 1,229 & 53,417 & 12.0 & automatic \\
		& Dev		   & 82	  & 1,148  & 19.3 & manual \\
		& Test	   & 100  & 1,517  & 18.7 & manual \\
		\midrule
		\multirow{3}{*}{WikiQA}& Train & 2,118 & 20,360 & 5.1 & manual \\
		& Dev	& 296 & 2,733 & 5.1 & manual \\
		& Test & 633 & 6,165 & 4.8 & manual	 \\
		\bottomrule
	\end{tabular}
	\caption[Summary of the answer sentence selection dataset]{Summary of the answer sentence selection dataset. Judgement denotes
		whether correctness was determined automatically or by human annotators.}
	\label{data-table}
\end{table}

\subsection{WikiQA Answer Selection Dataset}
The questions of WikiQA \citep{yang2015wikiqa} dataset are sampled from Bing query logs. The candidate answer sentences are obtained by (1) find the most relevant Wikipedia page for each question based on the user clicks, (2) make all the sentences in the summary paragraph of the page as the candidate answer sentences. The judgement of each answer sentence is provided by crowdsourcing workers. Compared to TREC QA, WikiQA is
less noisy and less biased towards lexical overlap\footnote{\cite{yang2015wikiqa}  provide a detailed explanation of the differences between the two datasets.}. The statistics of the WikiQA dataset is provided in Table \ref{data-table}. Same as the TREC QA dataset, the metrics for the dataset are MAP and MRR scores.

\subsection{Experimental Setup}
For the TREC QA dataset, we use word embeddings ($d = 50$) that are computed using Collobert and
Weston's neural language model \citep{DBLP:conf/icml/CollobertW08} and provided
by  \cite{DBLP:conf/acl/TurianRB10}. For the WikiQA dataset, the word embeddings are obtained by running the \texttt{word2vec} tool on the English Wikipedia dump and the {\it AQUAINT}\footnote{\url{https://catalog.ldc.upenn.edu/LDC2002T31}} corpus.
Even though our objective
function would allow us to learn word embeddings directly, we fix those
representations in light of the small size of the QA answer selection datasets.
LSTMs have the configuration of 3 layers, 50 hidden units and 40\% dropout after the embedding layer.
All hyperparameters are optimised via grid
search on the MAP score on the development data. We use the AdaGrad algorithm
\citep{DBLP:journals/jmlr/DuchiHS11} for training.

One weakness of the distributed approach is that --- unlike symbolic
approaches --- distributed representations are not very well equipped for dealing
with cardinal numbers and proper nouns, especially considering the small dataset.
As these are important artefacts in any
question answering task, we mitigate this issue by using a second feature that
counts co-occurring words in question-answer pairs. We integrate this feature
with our distributed models by training a logistic regression classifier with
three features: word co-occurrence count, word co-occurrence count weighted by
IDF value and QA matching probability as provided by the distributed model.
Notice that here we did not add more
$n$-gram features, since we would like to see how much the distributed models
contributed in this task, rather than $n$-gram overlapping. L-BFGS was used to
train the logistic regression classifier, with L2 regulariser of $0.01$.

\subsection{Results}
\label{sec:qa_results}

\begin{table}[t]\centering
	
	\begin{tabular}{@{}lrrrrr@{}}
		\toprule
		\multirow{2}{*}{Model} & \multicolumn{2}{c}{TREC QA} && \multicolumn{2}{c}{WikiQA}  \\
		\cmidrule{2-3}  \cmidrule{5-6}
		& MAP & MRR  && MAP & MRR \\
		\midrule
		Bag of words (BoW) 			& 0.5470 & 0.6329  && 0.5879 &  0.5964\\
		Bigram-CNN	 		& 0.5693 & 0.6613 && 0.6186 & 0.6271\\
		LSTM						& 0.6436 & 0.7235 && 0.6552 & 0.6747\\
		LSTM + attention	&  0.6451 & 0.7316 && 0.6639 &  0.6828\\
		BoW + count 	& 0.6934 & 0.7677 && 0.6121 & 0.6218 \\
		Bigram-CNN + count	& 0.7113 & 0.7846 && 0.6564 & 0.6676\\
		LSTM + count & 0.7228 &  0.7986 && 0.6820 &0.6988 \\
		LSTM + attention + count & \bfseries{0.7289} & \bfseries{0.8072} &&  \bfseries{0.6855} & \bfseries{0.7041} \\
		\bottomrule
	\end{tabular}
	\caption[Results for the answer selection task]{Results of our models on the TREC QA and WikiQA datasets. \textit{count} indicates whether the overlapping word count features are also used.}
	\label{our-table}
\end{table}

\begin{table}[t]\centering
	\begin{tabular}{@{}llrr@{}}
		\toprule
		\multicolumn{2}{@{}l}{System} & MAP & MRR \\
		\midrule
		\multicolumn{2}{@{}l}{\textbf{Baselines}} &&  \\
		& Random 		& 0.3965 & 0.4929 \\
		& Word Count  & 0.5707 & 0.6266 \\
		& Weighted Word Count & 0.5961 & 0.6515 \\
		\midrule
		\multicolumn{2}{@{}l}{\textbf{Published Models}} &&  \\
		& Wang et al. (2007)  		& 0.6029 	& 0.6852 \\
		& Heilman and Smith (2010) 	& 0.6091   	& 0.6917 \\
		& Wang and Manning (2010)   & 0.5951    & 0.6951 \\
		& Yao et al. (2013) 			& 0.6307	& 0.7477 \\
		& Severyn and Moschitti (2013)& 0.6781  	& 0.7358 \\
		& Yih et al. (2013)	& \bf{0.7092}	& \bfseries{0.7700} \\
		\midrule
		\multicolumn{2}{@{}l}{\textbf{Our Models}} &&  \\
		& BoW + count & 0.6934 & 0.7677 \\
		& Bigram-CNN + count	& 0.7113 & 0.7846 \\
		& LSTM + count & 0.7228 &  0.7986 \\
		& LSTM + attention + count & \bfseries{0.7289} & \bfseries{0.8072} \\
		\bottomrule
	\end{tabular}
	\caption[Overview of results on the TREC QA answer selection dataset]{Overview of results on the TREC QA dataset. Baseline models
		are taken from Table 2 in \cite{yih2013question}. We also include
		the results of our models, most of which outperform the
		state of the art.}
	\label{result-table}
\end{table}

\begin{table}[t]\centering
	\begin{tabular}{@{}llrr@{}}
		\toprule
		\multicolumn{2}{@{}l}{System} & MAP & MRR \\
		\midrule
		\multicolumn{2}{@{}l}{\textbf{Baselines}} &&  \\
		& Word Count  & 0.4891  &  0.4924 \\
		& Weighted Word Count &  0.5099 & 0.5132 \\
		\midrule
		\multicolumn{2}{@{}l}{\textbf{Published Models}} &&  \\
		& Yih et al. (2013)  		&  0.5993	& 0.6086 \\
		& Paragraph vector 	&  0.5110   	& 0.5160 \\
		& Paragraph vector + count   & 0.5976   & 0.6058 \\
		\midrule
		\multicolumn{2}{@{}l}{\textbf{Our Models}} &&  \\
		& BoW + count & 0.6121 & 0.6218 \\
		& Bigram-CNN + count & 0.6564 & 0.6676  \\
		& LSTM + count & 0.6820 & 0.6988 \\
		& LSTM + attention + count &  \bfseries{0.6855} & \bfseries{0.7041}\\
		\bottomrule
	\end{tabular}
	\caption[Overview of results on the WikiQA answer selection dataset]{Overview of results on the WikiQA dataset. Results of baseline models and prior work 
	are taken from \cite{yang2015wikiqa}. }
	\label{wiki-result-table}
\end{table}

Table \ref{our-table} summarises the results of our models.  As can be seen, on both datasets, the bigram-CNN model performs better than the bag-of-words model and the LSTM model outperforms the bigram-CNN model by a significant margin --- approximately 7\% on MAP and 6\% on MRR on TREC QA. The LSTM model with attention achieves the best results, although the difference between the attentional model and the vanilla LSTM is not large. The addition of the IDF-weighted word count features  improve performance for all models by $8 \%$ -- $15\%$ on the TREC QA dataset and $2\%$ on the WikiQA dataset. The smaller improvement on the WikiQA dataset confirms Yang et al's (\citeyear{yang2015wikiqa}) statement that the WikiQA dataset is less biased towards word overlapping compared to TREC QA.

Tables \ref{result-table} and \ref{wiki-result-table} survey published results on the TREC QA and WikiQA datasets, respectively, and place our models in the context of the state-of-the-art results (during the time when our work was published). Table \ref{result-table} also includes three baseline models provided in \cite{yih2013question}. The
first model randomly assigns scores to each answer. The second model counts the number of words co-occurring in each QA pair, with another version of that baseline weighting these word counts by IDF values. 
As can be seen in both tables, our models (bigram-CNN + count, LSTM + count, and LSTM + attention + count)
outperform all baselines and prior work on both MAP and MRR. Considering the lack of complexity of our models compared to those of previous work, these results are very promising and indicate the soundness of our approach to the QA answer selection task. 

Since our work was published, various extensions of our models have been proposed, with applications on answer sentence selection as well as other recognition tasks of seq2seq mappings. For example, \cite{severyn2015disi} leverages a deep CNN as the sentence model within the same QA mapping framework as ours, though the deep CNN performs on par with our bigram-CNN on this answer selection task.  
Rather than doing semantic matching only on the vector representation of sentences,  \cite{yin2015multi} and \cite{yin2015abcnn} propose to do the matching on multiple levels: word level, phrase level, and sentence level. The ACLwiki website\footnote{\url{https://www.aclweb.org/aclwiki/index.php?title=Question_Answering_(State_of_the_art)}} provides a full list of work on the TREC QA dataset. 
\section{Discussion}

As already stated in the background section of this chapter, most prior work
focuses on syntactic analysis, with semantic aspects mainly being incorporated
through a number of manually engineered features and external resources such as
WordNet. Interestingly, however, the best performing published model is also the
only piece of prior work that is primarily focused on semantics
\citep{yih2013question}. In their model, Yih et al. match aligned words between
questions and answers and extract features from word pairs. The word-level
features are then aggregated to represent sentences, which are used for
classification. They combine a group of word matching features with semantic
features obtained from a wide range of linguistic resources including WordNet,
polarity-inducing latent semantic analysis (PILSA) model \citep{yih2012polarity}
and different vector space models.

When considering the heavy reliance of resources of those models in comparison to the simplicity of our approach, the relative performance of our models is highly encouraging.
As we only use two non-distributed features---question-answer pair word
matching and word matching weighted by IDF values---it is plausible to regard
the distributional aspect of our models as a replacement for the numerous
lexical semantic resources and features utilised in Yih et al.'s work.
In the context of these results, it is also worth noting that---unlike the model of Yih et al.---our
models can directly be applied
across languages, as we are not relying on any external resources beyond some
large corpus on which to train our initial word embeddings.

Earlier in this chapter we argued that methods based purely on vector
representations may not be sufficient for solving complex problems such as
paraphrase detection and question answering because of their weakness in dealing with
certain aspects of language such as numbers and---to a lesser extent---proper
nouns. For example, the mismatching of numbers are crucial for rejecting a pair of
`paraphrases' or an answer. Surface-form matching is particularly important in
our experiment because we did not learn word embeddings from the answer
selection dataset and the given word dictionaries may not cover all the words in
the dataset.\footnote{Approximately $5 \%$ of words in the TREC answer selection
	dataset are not covered in Collobert and Weston's embeddings.} Most of the
non-covered words are proper nouns, which are then assigned the \texttt{UNKNOWN}
token. While these are likely to be crucial for judging the relevance of an
answer candidate, they cannot be incorporated into the distributional aspect of
our model.

However, it is also important to then establish the opposite fact, namely that
distributional semantics improve over purely word counting model.  When
reviewing the baseline results (Table \ref{result-table}) relative to the
performance of our models, it is also evident, that adding distributional
semantics as a feature improves over models based purely on co-occurrence counts
and word matching. In fact, this addition boosts both MAP and MRR scores by
$10\% - 20\%$ in the TREC QA dataset.  We analysed this effect by considering sentences where
our combined model (bigram + count) performs better than the counting baseline.
Here are two examples, where the model of co-occurrence word count failed to identify the
correct answer, but the combined model prevailed:
\begin{enumerate}
	\item
	\begin{itemize}[leftmargin=5mm]
		\item[\textbf{Q:}] When  did James Dean {\it die}?
		\item[\textbf{A1:}] In $\langle$num$\rangle$, actor James Dean was {\it
			killed} in a two-car collision near Cholame, Calif. ({\bf correct})
		\item[\textbf{A2:}] In $\langle$num$\rangle$, the studio asked him to
		become a technical adviser on Elia Kazan's ``East of Eden,'' starring
		James Dean. ({\bf incorrect})
	\end{itemize}
	\item
	\begin{itemize}[leftmargin=5mm]
		\item[{\bf Q:}] How many members are there in the {\it singing group} the
		Wiggles?
		\item[{\bf A1:}] The Wiggles are four effervescent {\it performers} from 
		the Sydney area: Anthony Field, Murry Cook, Jeff Fatt and Greg Page. ({\bf correct})
		\item[{\bf A2:}] Let's now give a welcome to the Wiggles, a goofy new
		import from Australia. ({\bf incorrect})
	\end{itemize}
\end{enumerate}
In both cases, the baseline model cannot tell the difference between the two
candidate answers since they have the same number of matched words to the
question. However, for the first example the combined model assigns higher score
to the first answer since the word {\it die} is semantically close to {\it
	killed}. Similarly, for the second example, the word {\it performers} in the
first sentence is related to {\it singing group} in the question, and hence the
first one gets higher score.

In Section \ref{sec:qa_results}, we have shown that on both answer selection datasets, the bigram-CNN model performs better than the bag-of-words model, and the bigram-CNN model is in turn outperformed by the LSTM model significantly. This indicates the models that are better at modelling sentences are also dominant in the seq2seq mapping problems. Examples 3 and 4 are two examples where the LSTM model identified the answer correctly, but the bigram-CNN model failed:

\begin{enumerate}
	\setcounter{enumi}{2}
	\item
	\begin{itemize}[leftmargin=5mm]
		\item[\textbf{Q:}] What type of business is walmart?
		\item[\textbf{A1:}] Wal-mart stores, branded as walmart, is an American multinational retail corporation that runs chains of large discount department stores and warehouse stores. ({\bf correct})
		\item[\textbf{A2:}] Walmart remains a family-owned business, as the company is controlled by the Walton family, who own a 48 percent stake in walmart. ({\bf incorrect})
	\end{itemize}
	\item
	\begin{itemize}[leftmargin=5mm]
		\item[{\bf Q:}] What order is the moth?
		\item[{\bf A:}] A {\it moth} is an insect related to the butterfly, both being of the {\it order} lepidoptera.
	\end{itemize}
\end{enumerate}
The judgement of Example 3 requires a good understanding of the sentence; and the answer in Example 4 contains long range dependencies, which can be captured by LSTMs. 

Empirically, the addition of the attention mechanism helps to improve the MAP and MRR scores over the vanilla LSTM model, which is in line with our expectation. We also analysed example QA pairs and found that the model is indeed able to attend to the key answer phrase of the given candidate sentence. For example, in the example below, the attention mechanism assigns high scores to the phrase {\it blue color of}.
\begin{enumerate} 
	\setcounter{enumi}{4}
	\item
	\begin{itemize}[leftmargin=5mm]
		\item[{\bf Q:}] What does a liquid oxygen plant look like?
		\item[{\bf A:}] The {\it blue color of} liquid oxygen in a dewar flask
	\end{itemize}
\end{enumerate}

\section{Summary}
In this chapter, we have empirically studied the effectiveness of applying distributed sentence models to answer sentence selection. In our proposed model, we project questions and answers into vectors and learn a semantic matching function between QA pairs. Within the uniform QA matching framework, we have compared the performance of different types of distributed sentence models, namely the bag-of-words model, the bigram-CNN model, the LSTM model, and the LSTM with attention. Experimental results and qualitative analysis demonstrate that the more sophisticated sentence models also generate better results on the seq2seq matching task (i.e. LSTM with attention $>$ LSTM $>$ bigram-CNN $>$ BoW). Our work is the first to apply distributed sentence models to the task of answer sentence selection, and outperforms the models from previous work based on feature engineering and external hand-coded semantic resources.

There are many avenues for future work related to the research presented in this chapter. An obvious extension is to investigate more complex representations of sentences, taking into consideration the syntactic structure of sentences. Subsequent work in this theme includes the Tree-LSTM \citep{DBLP:conf/acl/TaiSM15}, in which syntactic parse trees are built explicitly in the model, and the work by \cite{yogatama2016learning}, in which tree structures are learned for downstream tasks. Another extension is to leverage abundant unlabelled data or relevant paired data to train distributed sentence models in an unsupervised or semi-supervised way, which hopefully enables these models to learn semantics better. An example idea is to leverage the paraphrase dataset \citep{ganitkevitch2013ppdb}, and do multitask learning on the tasks of answer selection and paraphrase detection.

%% file: chapter_5.tex
\chapter{Sequence Transduction}
\label{ch:ssnt}

\begin{chapterabstract}
The focus of this chapter is the generation tasks of seq2seq mapping.
We propose a neural seq2seq model that learns to alternate between encoding and decoding segments of the input as it is read. By independently tracking the encoding and decoding representations our algorithm permits exact polynomial marginalisation of the latent segmentation during training, and during decoding beam search is employed to find the best alignment path together with the predicted output sequence. Orthogonal to the previous attentive models, this model tackles the bottleneck of vanilla encoder-decoders that have to read and memorise the entire input sequence in their fixed-length hidden states before producing any output. It also has the advantage of permitting online prediction. Experiments on a range of tasks show significant gains over the baseline encoder-decoders\let\thefootnote\relax\footnote{The material in this chapter was originally presented in \cite{yu:2016}. We reran some experiments after publishing the paper, and so the results may be different from those presented in the original paper.}. 
\addtocounter{footnote}{-1}\let\thefootnote\svthefootnote
\end{chapterabstract}

\section{Introduction}
In the previous chapter, we employ a simple bilinear model for scoring the relatedness of sentence vectors, aiming to compare the efficacy of different distributional sentence models in seq2seq mapping. In this chapter, we address seq2seq mapping problems from a different angle: modelling the relationship between two sequences. Instead of working on recognition tasks, we focus on generation tasks here. 

In Chapter \ref{ch:nn}, we have reviewed the encoder-decoder paradigm, where an input sequence is encoded into a fixed-size vector and an output sequence is then decoded from said vector \citep{sutskever2014sequence,cho2014learning}. This architecture is appealing, as it makes it possible to tackle the problem of seq2seq mapping by training a large neural network in an end-to-end fashion. However, it is difficult for a fixed-length vector to memorise all the necessary information of an input sequence, especially for long sequences. Often a very large encoding needs to be employed in order to capture the longest sequences, which invariably wastes capacity and computation for short sequences. 
While the attention mechanism of \cite{bahdanau2014neural} goes some way to address this issue, it still requires the full input to be seen before any output can be produced.

In this chapter, we propose an architecture to tackle the limitations of the vanilla encoder-decoder model, a segment to segment neural transduction model (SSNT) that learns to generate and align simultaneously.
SSNT is inspired by the HMM word alignment model proposed for statistical machine translation \citep{vogel1996hmm,tillmann1997dp}; we impose a monotone restriction on the alignments but incorporate recurrent dependencies on the input which enable rich locally non-monotone alignments to be captured. 
That is, the model can deal with word reorderings by reading more input tokens until it thinks the phrase is complete and then starting to generate output tokens corresponding to the phrase. 
Our model is similar to the sequence transduction model of \cite{graves2012sequence}, but we propose alignment distributions which are parameterised separately, making the model more flexible and allowing online inference. 

SSNT introduces a latent segmentation which determines correspondences between tokens of the input sequence and those of the output sequence. The aligned hidden states of the encoder and decoder are used to predict the next output token and to calculate the transition probability of the alignment. We carefully design the input and output RNNs such that they independently update their respective hidden states. This enables us to derive an exact dynamic programme to marginalise out the hidden segmentation during training and an efficient beam search to generate online the best alignment path together with the output sequence during decoding. Unlike previous recurrent segmentation models that only capture dependencies in the input \citep{graves2006connectionist,kong2015segmental}, our segmentation model is able to capture unbounded dependencies in both the input and output sequences while still permitting polynomial inference. 

While attentive models treat the attention weights as the output of a deterministic function, SSNT assigns attention weights to a sequential latent variable which can be marginalised out.
SSNT is general and could be incorporated into any RNN-based encoder-decoder architecture, such as Neural Turing Machines \citep{graves2014neural}, memory networks \citep{weston2014memory,kumar2015ask} or stack-based networks \citep{grefenstette2015learning}, enabling such models to process data online.

We conduct experiments on four different transduction tasks, abstractive sentence summarisation, Chinese-English and French-English machine translation (seq2seq mapping at word level), and morphological inflection generation (seq2seq mapping at character level). We evaluate our proposed algorithms in both the online setting, where the input is encoded with a unidirectional LSTM, and the offline setting, where the whole input is available such that it can be encoded with a bidirectional network.
The experimental results demonstrate the effectiveness of SSNT --- it consistently outperforms the baseline encoder-decoder approach while requiring significantly smaller hidden layers, thus showing that the segmentation model is able to learn to break one large transduction task into a series of smaller encodings and decodings.
When bidirectional encodings are used the segmentation model outperforms an attention-based benchmark.
Qualitative analysis shows that the alignments found by our model are highly intuitive and demonstrates that the model learns to read ahead the required number of tokens before producing output.

The rest of this chapter is organised as follows: we first motivate our work by  comparing it with the previous related work. We then formally describe SSNT in \S \ref{ssnt_model} and  provide the forward-backward algorithm, which enables us to calculate the loss and gradient efficiently (\S \ref{ssnt_train_decode}). This is followed by the description of a decoding algorithm. After presenting experiments in \S \ref{ssnt_exp}, we analyse the quality of the alignment found by SSNT, and show examples which SSNT processes well/poorly (\S \ref{ssnt_analysis}). Finally, we conclude this chapter by highlighting the advantages and limitations of our model.
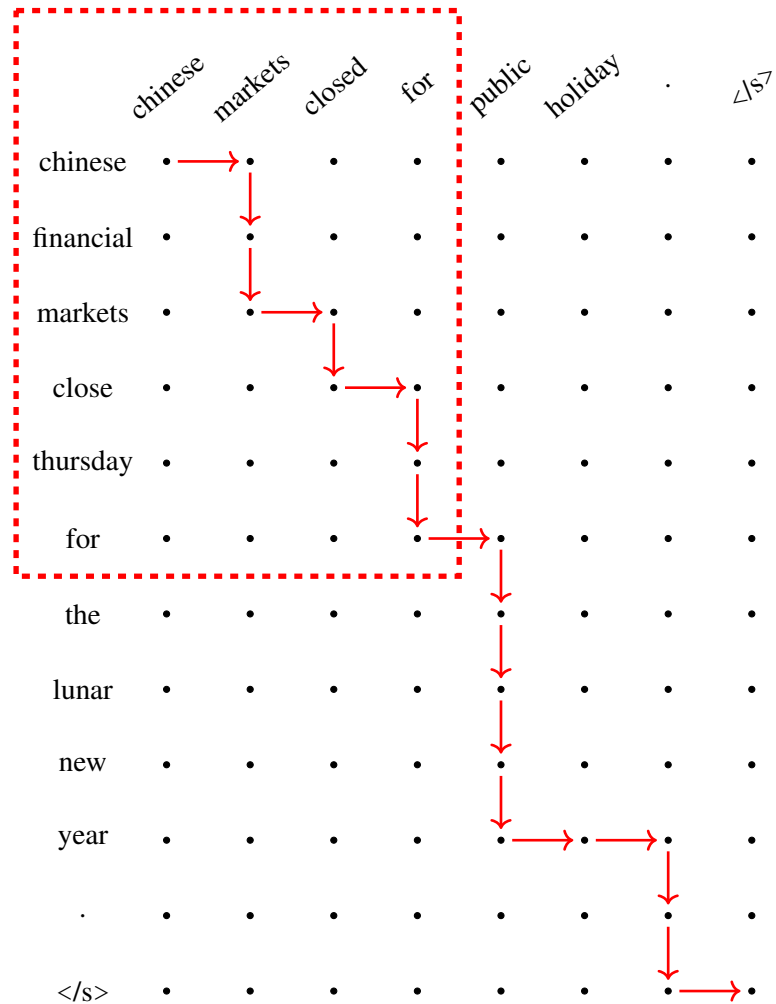
\begin{figure}
	\centering
	\begin{tikzpicture}
	\def \xscale {1.1}
	\def \yscale {1.0}
	
	\foreach \x in {0,...,7}
	\foreach \y in {0,...,11} 
	{\draw [fill]  (\xscale*\x, \y*\yscale) circle [radius=0.04] ;
		\node   (\x\y) at (\xscale*\x,\y*\yscale) {};} 
	
	\foreach \c [count=\x from 0] in {
		{\texttt{<}/s\texttt{>}},{.},{year},{new},{lunar},{the},{for},{thursday},{close},{markets}, {financial},{chinese}} 
	{
		\node at (-\xscale,\x*\yscale) {\small \c};	
	}

	\foreach \c [count=\x from 0] in {
		{chinese},{markets},{closed},{for}, {public},{holiday}, {.},{\texttt{<}/s\texttt{>}}} 
	{
		\node [rotate=40] at (\xscale*\x,12*\yscale) {\small \c};	
	} 
	
	\draw [->,red,line width=1] 
	(011) edge (111) 
	(111) edge (110)
	(110) edge (19)
	(19) edge (29)
	(29) edge (28)
	(28) edge (38)
	(38) edge (37)
	(37) edge (36)
	(36) edge (46)
	(46) edge (45)
	(45) edge (44)
	(44) edge (43)
	(43) edge (42)
	(42) edge (52)
	(52) edge (62)
	(62) edge (61)
	(61) edge (60)
	(60) edge (70);
	
	\draw  [dashed, red,line width=2] (-1.8*\xscale,13*\yscale) -- (3.5*\xscale,13*\yscale) -- (3.5*\xscale,5.5*\yscale) -- (-1.8*\xscale,5.5*\yscale) -- cycle;
	
	\end{tikzpicture}
	\caption[Example output of the neural transduction model]{Example output of our recurrent segmentation model on the task of abstractive sentence summarisation. The path highlighted is the alignment found by the model during decoding. Horizontal moves correspond to generating an output token while vertical moves denote reading a token from the input. The input and output are independently encoded with LSTMs, while generation is conditioned on the concatenation of these encodings at a given grid cell. This architecture permits polynomial time marginalisation of all possible paths while still capturing unbounded dependencies in the input and output. If unidirectional LSTM is used to encode the source sentence, then the model is capable of doing incremental predictions. That is, the tokens that are generated are conditioned only on the prefix of the input. As highlighted by the red rectangle, the generated partial output `chinese markets closed for' is only based on the phrase `chinese financial markets close thursday for' in the input sentence.
	}
	\label{example_graph}
\end{figure}

\section{Related Work}
\label{ssnt_related_work}
Our work is inspired by the seminal HMM alignment model \citep{vogel1996hmm,tillmann1997dp} proposed for machine translation. In contrast to that work, when predicting a target word we additionally condition on all previously generated words, which is enabled by the recurrent neural models. This means that the model also functions as a conditional language model. It can therefore be applied directly, while traditional models have to be combined with a language model through a noisy channel in order to be effective. 
Additionally, instead of EM training on the most likely alignments at each iteration, our model is trained with direct gradient descent, marginalising over all the alignments. Further, the latent segmentation variable in our model is different from those alignment variables in traditional alignment models. Those models have to predict its correspondence immediately while reading a token, and a monotone restriction on alignments would make them impossible to tackle sequences with reorderings. In contrast, under the same constraint, our model can still capture non-monotone alignments by reading sufficient content before generating a segment.

Latent variables have been employed in neural network-based models for sequence labelling tasks in the past. Examples include connectionist temporal classification (CTC) \citep{graves2006connectionist} for speech recognition and the more recent segmental recurrent neural networks (SRNNs) \citep{kong2015segmental}, with applications on handwriting recognition and part-of-speech tagging. Weighted finite-state transducers (WFSTs) have also been augmented to encode input sequences with bidirectional LSTMs~\citep{rastogi2016weighting}, permitting exact inference over all possible output strings.  
While these models have been shown to achieve appealing performance on different applications, they have common limitations in terms of modelling dependencies between labels. It is not possible for CTCs to model explicit dependencies. SRNNs and neural WFSTs model fixed-length dependencies, making it difficult to carry out effective inference as the dependencies become longer. 

Our model shares the property of the sequence transduction model of \cite{graves2012sequence} in being able to model unbounded dependencies between output tokens via an output RNN.
This property makes it possible to apply our model to tasks such as summarisation and machine translation that require the tokens in the output sequence to be modelled highly dependently. 
\cite{graves2012sequence} models the joint distribution over outputs and alignments by inserting null symbols (representing shift operations) into the output sequence. During training the model uses dynamic programming to marginalise over permutations of the null symbols, while beam search is employed during decoding.
In contrast, our model defines a separate latent alignment variable, which adds flexibility to the way the alignment distribution can be defined (as a geometric distribution or parameterised by a neural network) and how the alignments can be constrained, without redefining the dynamic program. In addition to marginalising during training, our decoding algorithm also makes use of dynamic programming, allowing us to use either no beam or small beam sizes.

Similar to our work, the model of \cite{alkhoulialignment} is decomposed into the alignment model and the model of word predictions. The two models are trained separately and combined during decoding, with subsequent refinements using a Viterbi-EM approximation. In contrast, the latent and observed components of the models are trained jointly using a dynamic program to exactly marginalise the unobserved variables in our model.

Our work is also related to the attention-based models first introduced for machine translation~\citep{bahdanau2014neural}.
\cite{DBLP:conf/emnlp/LuongPM15} proposed two alternative attention mechanisms: a global method that attends all words in the input sentence, and a local one that points to parts of the input words.  Another variation on this theme are pointer networks \citep{vinyals2015pointer}, where the outputs are pointers to elements of the variable-length input, predicted by the attention distribution.

Although our model shares the same idea of joint training and aligning with the attention-based models, our design has fundamental differences and advantages.  While attention-based models treat the attention weights as the output of a deterministic function (soft-alignment), in our model the attention weights correspond to a hidden variable, that can be marginalised out using dynamic programming. Further, our model's inherent online nature permits it the flexibility to use its capacity to choose how much input to encode before decoding each segment.

Another trend of work that is related to our model is the investigation of making online prediction for machine translation \citep{gu:2016,grissom:2014,sankaran:2010} and speech recognition \citep{hwang:2016,jaitly2015online}.

In the next section, we will formally describe our neural transduction model.

\section{Model}
\label{ssnt_model}

\begin{figure}
	\centering
	\begin{subfigure}[t]{0.4\textwidth}
		\begin{tikzpicture}[>=stealth,shorten >=1pt,auto]
		\def \xscale {1.35}
		\def \yscale {1.35}
		
		\foreach \x in {0,...,4}
		\foreach \y in {0,...,4} 
		{\draw [fill]  (\xscale*\x, \y*\yscale) circle [radius=0.04] ;
			\node   (\x\y) at (\xscale*\x,\y*\yscale) {};} 
		
		\foreach \x in {0,...,3}
		\foreach \y in {0,...,4} 
		{  \pgfmathtruncatemacro{\xnext}{\x+1};
			\draw [->,dashed] 
			(\x\y) edge (\xnext\y); } 
		
		\foreach \x in {0,...,4}
		\foreach \y in {1,...,4} 
		{  \pgfmathtruncatemacro{\yprev}{\y-1};
			\draw [->,dashed] 
			(\x\y) edge (\x\yprev); }

		\foreach \c [count=\y from 0] in {
			{\texttt{<}/s\texttt{>}},{$x_4$},{$x_3$},{$x_2$}, {$x_1$}} 
		{
			\node at (-0.5*\xscale,\y*\yscale) {\c};	
		} 
		
		\foreach \c [count=\x from 0] in {
			{\texttt{<}s\texttt{>}},{$y_1$},{$y_2$},{$y_3$}, {$y_4$}} 
		{
			\node at (\x*\xscale,4.5*\yscale) {\c};	
		} 
		
		\draw [->,red,line width=0.4mm]
		(04) edge (03)
		(03) edge (02)
		(02) edge (01)
		(01) edge (00)
		(00) edge (10)
		(10) edge (20)
		(20) edge (30)
		(30) edge (40)
		;
		
		\end{tikzpicture}
		\caption{}
		\label{ssnt_alignment1}
	\end{subfigure}\hfill
	\begin{subfigure}[t]{0.5\textwidth}
			\begin{tikzpicture}[>=stealth,shorten >=1pt,auto]
			\def \xscale {1.35}
			\def \yscale {1.35}
			
			\foreach \x in {0,...,4}
			\foreach \y in {0,...,4} 
			{\draw [fill]  (\xscale*\x, \y*\yscale) circle [radius=0.04] ;
				\node   (\x\y) at (\xscale*\x,\y*\yscale) {};} 
			
			\foreach \x in {0,...,3}
			\foreach \y in {0,...,4} 
			{  \pgfmathtruncatemacro{\xnext}{\x+1};
				\draw [->,dashed] 
				(\x\y) edge (\xnext\y); } 
			
			\foreach \x in {0,...,4}
			\foreach \y in {1,...,4} 
			{  \pgfmathtruncatemacro{\yprev}{\y-1};
				\draw [->,dashed] 
				(\x\y) edge (\x\yprev); }

			\foreach \c [count=\y from 0] in {
				{\texttt{<}/s\texttt{>}},{$x_4$},{$x_3$},{$x_2$}, {$x_1$}} 
			{
				\node at (-0.5*\xscale,\y*\yscale) {\c};	
			} 
			
			\foreach \c [count=\x from 0] in {
				{\texttt{<}s\texttt{>}},{$y_1$},{$y_2$},{$y_3$}, {$y_4$}} 
			{
				\node at (\x*\xscale,4.5*\yscale) {\c};	
			} 
			
			\draw [->,red,line width=0.4mm]
			(04) edge (14)
			(14) edge (13)
			(13) edge (12)
			(12) edge (22)
			(22) edge (21)
			(21) edge (31)
			(31) edge (30)
			(30) edge (40)
			;
			
			\end{tikzpicture}
		\caption{}
		\label{ssnt_alignment2}
	\end{subfigure}
	\caption[Possible alignment paths of the transduction model]{Possible alignment paths for a given input-output pair $(\boldsymbol{x}, \boldsymbol{y})$. Arrows represent transitions. Forward arrows correspond to the \textsc{emit} operation, and downward arrows correspond to the \textsc{shift} operation. Any path from the top left node to the bottom right node is a possible alignment. During decoding, the model will find the best alignment path, which is illustrated with the red lines.}  
	\label{ssnt_alignment}
\end{figure}
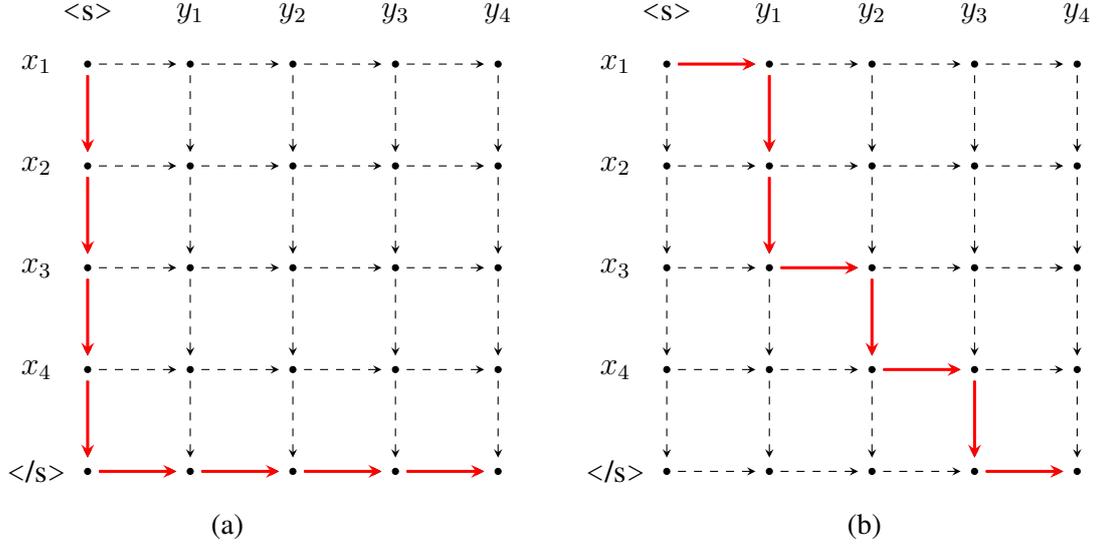

Let $\boldsymbol x_1^I$ be the input sequence of length $I$ and $\boldsymbol{y}_1^J$ the output sequence of length $J$. Let $y_j$ denote the $j$-th token of $\boldsymbol{y}$. Our goal is to model the conditional distribution
\begin{equation}
p(\boldsymbol{y}\ |\ \boldsymbol{x}) = \prod_{j=1}^{J} p(y_j\ |\ \boldsymbol{y}_1^{j-1}, \boldsymbol{x}).
\end{equation}
We introduce a hidden alignment variable $\boldsymbol{z}_1^J$ which indicates when each token of the output sequence is to be generated as the input sequence is being read. 
Each $z_j = i$ denotes that the output token at position $j$ ($y_j$) is generated when the input sequence up through position position $i \in \{1, \dots, I\}$ has been read. Then $p(\boldsymbol{y}\ |\ \boldsymbol{x})$ is calculated by marginalising over all the hidden alignments,
\begin{align}
\label{con_prob}
\begin{split}
p(\boldsymbol{y} \mid \boldsymbol{x}) & = \sum_{\boldsymbol{z}} p(\boldsymbol{y}, \boldsymbol{z} \mid \boldsymbol{x}) \\
p(\boldsymbol{y}, \boldsymbol{z} \mid \boldsymbol{x}) & \approx   \prod_{j=1}^{J} \underbrace{p(z_j \mid z_{j-1},
	\boldsymbol{x}_{1}^{z_j},
	\boldsymbol{y}_1^{j-1})}_{\text{alignment probability}} \underbrace{p(y_j \mid \boldsymbol{x}_{1}^{z_j},
	\boldsymbol{y}_1^{j-1})}_{\text{word probability}}. 
\end{split}
\end{align}

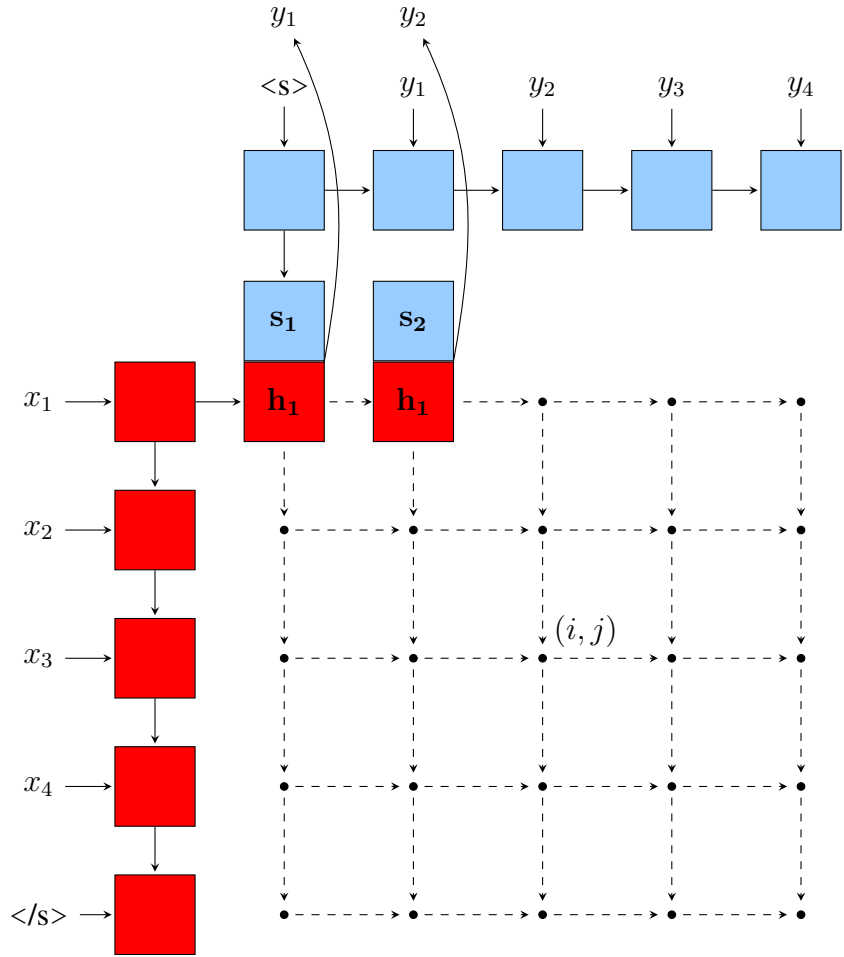
\begin{figure}[h]
	\centering
	\begin{tikzpicture}[>=stealth,shorten >=1pt,auto,scale=1.0]
	\def \xscale {1.7}
	\def \yscale {1.7}
	\def \retangleSize {30}
	
	\definecolor{color1}{RGB}{255,0,0}
	\definecolor{color2}{RGB}{153,204,255}
	
	\tikzstyle{retangle_style1}=[draw,black,fill=color1,inner sep=0pt,minimum width=\retangleSize,minimum height=\retangleSize]
	\tikzstyle{retangle_style2}=[draw,black,fill=color2,inner sep=0pt,minimum width=\retangleSize,minimum height=\retangleSize]
	
	\node (x5) at (1.1*\xscale,0)  {\texttt{<}/s\texttt{>}}; 
	\node (x4) at (1.1*\xscale,1*\yscale)  {$x_4$}; 
	\node (x3) at (1.1*\xscale,2*\yscale)  {$x_3$}; 
	\node (x2) at (1.1*\xscale,3*\yscale)  {$x_2$}; 
	\node (x1) at (1.1*\xscale,4*\yscale)  {$x_1$};

	\foreach \x in {0,...,4}
	\foreach \y in {0,...,4} 
	{\draw [fill]  (\xscale*3+\xscale*\x, \y*\yscale) circle [radius=0.05] ;
		\node   (\x\y) at (\xscale*3+\xscale*\x,\y*\yscale) {};}
	
	\node (x12)[retangle_style1] at (\xscale*2,4*\yscale) {};
	
	\node (x13up)[retangle_style2] at (\xscale*3,4.63*\yscale) {$\bf s_1$};
	\node (x13below)[retangle_style1] at (\xscale*3,4*\yscale) {$\bf h_1$};
	
	\node (x14up)[retangle_style2] at (\xscale*4,4.63*\yscale) {$\bf s_2$};
	\node (x14below)[retangle_style1] at (\xscale*4,4*\yscale) {$\bf h_1$};
	
	\node (x22)[retangle_style1] at (\xscale*2,3*\yscale) {};
	
	\node (x32)[retangle_style1] at (\xscale*2,2*\yscale) {};
	
	\node (x42)[retangle_style1] at (\xscale*2,1*\yscale) {};
	
	\node (x52)[retangle_style1] at (\xscale*2,0*\yscale) {};
	
	\node (y1) at (\xscale*3,\yscale*6.45) {\texttt{<}s\texttt{>}};
	\node (y11)[retangle_style2] at (\xscale*3,\yscale*5.65) {};
	
	\node (y2) at (\xscale*4,\yscale*6.45) {$y_1$};
	\node (y21)[retangle_style2] at (\xscale*4,\yscale*5.65) {};
	
	\node (y3) at (\xscale*5,\yscale*6.45) {$y_2$};
	\node (y31)[retangle_style2] at (\xscale*5,\yscale*5.65) {};
	
	\node (y4) at (\xscale*6,\yscale*6.45) {$y_3$};
	\node (y41)[retangle_style2] at (\xscale*6,\yscale*5.65) {};
	
	\node (y5) at (\xscale*7,\yscale*6.45) {$y_4$};
	\node (y51)[retangle_style2] at (\xscale*7,\yscale*5.65) {};

	\draw [->] 
	(x1) edge (x12)
	(x12) edge (x13below.west)
	(x2) edge (x22)
	(x12) edge (x22)
	(x3) edge (x32)
	(x22) edge (x32)
	(x4) edge (x42)
	(x32) edge (x42)
	(x5) edge (x52)
	(x42) edge (x52)
	;
	
	\draw [->] 
	(y11) edge (y21)
	(y11) edge (x13up)
	(y21) edge (y31)
	(y31) edge (y41)
	(y41) edge (y51)
	(y1) edge (y11)
	(y2) edge (y21)
	(y3) edge (y31)
	(y4) edge (y41)
	(y5) edge (y51)
	;
	
	\node (between) at (\xscale*3.29,\yscale*4.2) {};
	\draw [<-]
	node at (\xscale*3,\yscale*7.0) {$y_1$}  edge[in=80,out=295] (between);
	
	\node (between) at (\xscale*4.29,\yscale*4.2) {};
	\draw [<-]
	node at (\xscale*4,\yscale*7.0) {$y_2$}  edge[in=80,out=295] (between);

	\draw [->,dashed]
	(\xscale*3.35,\yscale*4) -- (\xscale*3.7,\yscale*4);
	
	\draw [->,dashed]
	node at (\xscale*3,\yscale*3.7) {} edge (03)
	node at (\xscale*4,\yscale*3.7) {} edge (13)
	node at (\xscale*4.3,\yscale*4) {} edge (24)
	;
	\foreach \x in {0,...,3}
	\foreach \y in {0,...,4} 
	{  
		\ifthenelse{ \x=0 \AND \y=4 \OR \x=1 \AND \y=4}{}{
			\pgfmathtruncatemacro{\xnext}{\x+1};
			\draw [->,dashed] (\x\y) edge (\xnext\y);
		}
	} 
	
	\foreach \x in {0,...,4}
	\foreach \y in {1,...,4} 
	{  
		\ifthenelse{ \x=0 \AND \y=4 \OR \x=1 \AND \y=4}{}{
			\pgfmathtruncatemacro{\ynext}{\y-1};
			\draw [->,dashed] (\x\y) edge (\x\ynext);
		}
	}

	\node at (22) [above right] {$(i,j)$};
	
	\end{tikzpicture}
	\caption[Structure of the neural transduction model]{The structure of our model. $(x_1, x_2, x_3, x_4)$ and $(y_1, y_2, y_3, y_4)$ denote the input and output sequences, respectively. The points, e.g. $(i,j)$, in the grid represent an alignment between $x_i$ and $y_j$. For each column $j$, the concatenation of the hidden states $[\mathbf{h}_i, \mathbf{s}_j]$ is used to predict $y_{j}$.}
	\label{model_structure}
\end{figure}


Figure \ref{ssnt_alignment} illustrates the model graphically. Each path from the top left node to the bottom right node in the graph corresponds to an alignment. 
The path highlighted in Figure  \ref{ssnt_alignment2} is a possible alignment path indicating that the model learns to alternate between reading and writing small segments. The path highlighted in Figure \ref{ssnt_alignment1} is another possible path, which in theory can be obtained in the extreme case where there are dramatic word reorderings, i.e. the last token of the input sequence is aligned to the first token of the output sequence. In this case, the model falls back into the vanilla encoder-decoder: it consumes all the input tokens before starting to generate output tokens.
We constrain the alignments to be monotone $z_{j+1} \geq z_{j}$, i.e. only forward and downward transitions are permitted at each point in the grid.
This constraint enables the model to learn to perform online generation. 
Another constraint on the alignments is to ensure that the entire input sequence is consumed before last output word is emitted, i.e. all valid alignment paths have to end in the bottom right corner of the grid. This constraint biases away from predicting outputs without explaining them using the input sequence.

The probability contributed by an alignment is obtained by accumulating the probability of word predictions at each point on the path and the transition probability between points.
The transition probabilities and the word output probabilities are modelled by neural networks, which are described in detail in the following sub-sections.

\subsection{Probabilities of Output Word Predictions}
The input sentence $\boldsymbol{x}$ is encoded with a Recurrent Neural Network (RNN), in particular an LSTM \citep{hochreiter1997long}. The encoder can either be a unidirectional or bidirectional LSTM. If a unidirectional encoder is used the model is able to read input and generate output symbols online. The hidden state vectors are computed as 
\begin{align}
\mathbf{h}_i^\rightarrow &= \text{RNN}(\mathbf{h}_{i-1}^\rightarrow, v^{(enc)}(x_i)), \\
\mathbf{h}_i^\leftarrow &= \text{RNN}(\mathbf{h}_{i+1}^\leftarrow ,v^{(enc)}(x_i)) ,
\end{align}
where $v^{(enc)}(x_i)$ denotes the vector representation of the token $x_i$, and $\mathbf{h}_i^\rightarrow$ and $\mathbf{h}_i^\leftarrow$ are the forward and backward hidden states, respectively. For a bidirectional encoder, they are concatenated as $\mathbf{h}_i = [\mathbf{h}_i^\rightarrow; \mathbf{h}_i^\leftarrow]$; and for unidirectional encoder $\mathbf{h}_i = \mathbf{h}_i^\rightarrow$. The hidden state $\mathbf{s}_j$ of the RNN for the output sequence $\boldsymbol{y}$ is computed as 
\begin{equation}
\mathbf{s}_j = \text{RNN}(\mathbf{s}_{j-1}, v^{(dec)}(y_{j-1})), 
\end{equation}
where $v^{(dec                                                                                                                                                                            )}(y_{j-1})$ is the encoded vector of the previously generated output word $y_{j-1}$. Note that here we let $\mathbf{s}_j$ denote the hidden state that excludes $y_j$, i.e., the encoding of the prefix $\boldsymbol{y}_1^{j-1}$.

To calculate the probability of the next word, we concatenate the aligned hidden state vectors $\mathbf{s}_j$ and $\mathbf{h}_{z_j}$ and feed the result into a softmax layer,
\begin{equation}
\label{word_pred}
\ p(y_j = l\ |\ \boldsymbol{y}_1^{j-1}, \boldsymbol{x}_1^{z_j}) 
=  \text{softmax} (\mathbf{W}^w[\mathbf{h}_{z_j};\mathbf{s}_j] + \mathbf{b}^w)_l.
\end{equation}
The model thus depends on the current alignment position $z_j$, which determines how far into $\boldsymbol{x}$ it has read.
The word output distribution in \cite{graves2012sequence} is parameterised in similar way.

Figure \ref{model_structure} illustrates the model structure. 
Note that the hidden states of the input and output decoders are kept independent to permit tractable inference, while the output distributions are conditionally dependent on both.

\subsection{Transition Probabilities}


We now discuss how the sequence of $z_j$'s are generated. First, to ensure the model's capability of doing incremental predictions, we should be careful in modelling this distribution so as to avoid conditioning on the entire input sequence. To illustrate why one might induce a dependency on the entire input sequence in this model, it is useful to compare to a standard attention model. Attention models operate by computing a score using a representation of alignment candidate (in our case, the candidates would be every unread token remaining in the input). If we followed this strategy, it would be necessary to observe the full input sequence when making the first alignment decision.

We instead model the alignment transition from timestep $j$ to $j+1$ by decomposing it into a sequence of conditionally independent \textsc{shift} and \textsc{emit} operations that progressively decide whether to read another token or stop reading. That is, at input position $i$, the model decides to \textsc{emit}, i.e., to set $z_j=i$ and predict the next output token $y_j$ from the word model, or it decides to \textsc{shift}, i.e., to read one more input token and increment the input position $i \gets i+1$. 
While the multinomial distribution is an alternative for parameterising alignments, the \textsc{shift}/\textsc{emit} parameterisation does not place an upper limit on the jump size, as a multinomial distribution would, and biases the model towards shorter jump sizes, which a multinomial model would have to learn.

We describe two methods for modelling the alignment transition probability.

\textbf{Transition Probability Model 1} ~  The first approach is independent of the input or output words. To parameterise the alignment distribution in terms of \textsc{shift} and \textsc{emit} operations we use a geometric distribution,
\begin{equation}
p(z_j\ |\ z_{j-1}) = (1-e)^{z_j - z_{j-1}} e,
\end{equation}
where $e$ is the emission probability. This transition probability only has one parameter $e$, which can be estimated directly by maximum likelihood as
\begin{equation}
e = \frac{\sum_n J_n}{\sum_n I_n + \sum_n J_n},
\end{equation}
where $I_n$ and $J_n$ are the lengths of the input and output sequences of training example $n$, respectively.

\textbf{Transition Probability Model 2} ~ For the second method we model the transition probability with a neural network. The probability $p(e_{i,j} = \textsc{emit} \mid \boldsymbol{x}_1^{i}, \boldsymbol{y}_1^{j-1})$ is calculated using the encoder and decoder states defined above as:
\begin{align}
p(e_{i,j} = \textsc{emit} \mid \boldsymbol{x}_{1}^{i}, \boldsymbol{y}_1^{j-1}) = \sigma(\text{MLP}(\mathbf{W}^t[\mathbf{h}_{i};\mathbf{s}_j] + b^t)).
\end{align}
The probability of \textsc{shift} is simply $1-p(e_{i,j} = \textsc{emit})$. In this formulation, the probabilities of aligning $z_j$ to each alignment candidate $i$ can be computed by reading just $\boldsymbol{x}_1^i$ (rather than the entire sequence). The probabilities are also independent of the contents of the suffix $\boldsymbol{x}_{i+1}^{I}$.

Using the probabilities of the auxiliary $e_{i,j}$ variables, the alignment probabilities needed in Eq.~\ref{con_prob} are computed as:
\begin{align*}
p(z_j = i \mid z_{j-1}, \boldsymbol{y}_1^{j-1}, \boldsymbol{x}_{1}^{i}) &= \begin{cases}
0 & \text{if }i < z_{j-1} \\
p(e_{i,j} = \textsc{emit}) & \text{if }i=z_{j-1} \\
\left(\prod_{i'=z_{j-1}}^{i-1} p(e_{i',j} = \textsc{shift}) \right) p(e_{i,j} = \textsc{emit}) & \text{if }i>z_{j-1}
\end{cases}
\end{align*}


\section{Inference Alogirthms}
\label{ssnt_train_decode}
Since there are an exponential number of possible alignments, it is computationally intractable to explicitly calculate every $p(\boldsymbol{y}, \boldsymbol{z}\ |\ \boldsymbol{x})$ and then sum them to get the conditional probability $p(\boldsymbol{y}\ |\ \boldsymbol{x})$. As described in Eq. \ref{word_pred}, the probability of generating each $y_j$ depends only on the current output position's alignment ($z_j$), the current output prefix ($\boldsymbol{y}_1^{j-1}$), and the input prefix up to the current alignment ($\boldsymbol{x}_1^{z_j}$). It does \emph{not} depend on the history of the alignment decisions. Likewise, the alignment decisions at each position are also conditionally independent of the history of alignment decisions given the decision at timestep $j-1$. Because of these  independence assumptions, $\boldsymbol{z}$ can be marginalised using a $O(|\boldsymbol{x}|^2 \cdot |\boldsymbol{y}|)$ time dynamic-programming algorithm where each fills in a chart with computing the marginal probabilities. The gradients of this objective with respect to the component probability models can be computed using automatic differentiation or using a secondary dynamic program that computes `backward' probabilities. In this section, we provide the algorithms for calculating the objective and the gradients.
\subsection{Forward Algorithm}
For an input $\boldsymbol{x}$ and output $\boldsymbol{y}$, 
the forward variable $\alpha(i,j) = p(z_j=i, \boldsymbol{y}_1^j\ |\ \boldsymbol{x}_1^{z_j})$.
The value of $\alpha(i,j)$ is computed by summing over the probabilities of every path that could lead to this cell. Formally, $\alpha(i,j)$ is defined as follows:

For  $i \in [1, I]$:
\begin{equation}
\alpha(i, 1)  = p(z_1 = i ) p(y_1\ |\ \boldsymbol{x}_1^{i}, y_0).
\end{equation}

For $j \in [2, J-1]$, $i \in [1, I]$ and $(i=I, j=J)$:
\begin{align}
\alpha(i,j) = &\ p(y_j\ |\ \boldsymbol{x}_1^{i}, \boldsymbol{y}_1^{j-1}) \cdot \sum_{k=1}^{i}\alpha(k, j-1)p(z_j = i\ |\ z_{j-1} = k, \boldsymbol{x}_1^{i}, \boldsymbol{y}_1^{j-1}). 
\end{align}

For $j = J$, $i \in [1, I-1]$:
\begin{align}
\alpha(i, J) = 0.
\end{align}

The last case corresponds to the constraint that the full input is consumed when the final output symbol is generated.

\subsection{Backward Algorithm}
The backward variables, defined as $\beta(i,j) = p(\boldsymbol{y}^J_{j+1}\ |\ z_j=i,\boldsymbol{y}_1^j, \boldsymbol{x})$, are computed as:

For $i \in [1, I]$:
\begin{equation}
\beta(i, J) = 1.
\end{equation}

For $j \in [1, J-1]$, $i \in [1, I]$:
\begin{align}
\beta(i,j) = \sum_{k=i}^{I} \beta(k, j+1)  p(z_{j+1} = k\ |\ z_j = i, \boldsymbol{x}_1^k, \boldsymbol{y}_1^j) p(y_{j+1}\ |\ \boldsymbol{x}_1^k, \boldsymbol{y}_1^j). 
\end{align}

\subsection{Training Objective and Gradients}
\label{sec:ssnt_loss}
During training we estimate the parameters by minimising the negative log likelihood of the training set $S$:
\begin{equation}
\begin{split}
\mathcal{L}(\boldsymbol{\theta}) &= - \sum_{(\boldsymbol{x}, \boldsymbol{y}) \in S} \log p(\boldsymbol{y}\ |\ \boldsymbol{x}; \boldsymbol{\theta})\\
&= - \sum_{(\boldsymbol{x}, \boldsymbol{y}) \in S} \log \alpha(I, J).\\ 
\end{split}
\end{equation}

In Appendix \ref{appendix_a}, we present two methods of deriving the gradients in detail. Here, we briefly present one of the methods.    Let $\boldsymbol{\theta}_j$ be the neural network parameters with respect to the model output at position $j$. 
The gradient is computed as:
\begin{equation}
\frac {\partial \log p(\boldsymbol{y}\ |\ \boldsymbol{x}; \boldsymbol{\theta})} 
{\partial \boldsymbol{\theta}} 
= \sum_{j=1}^J \sum_{i=1}^I 
\frac {\partial \log p(\boldsymbol{y}\ |\ \boldsymbol{x}; \boldsymbol{\theta})} 
{\partial \alpha(i, j)}
\frac {\partial \alpha(i, j)} {\partial \boldsymbol{\theta}_j}.
\end{equation}
The derivative with respect to the forward weights is
\begin{equation}
\frac {\partial \log p(\boldsymbol{y}\ |\ \boldsymbol{x}; \boldsymbol{\theta})} 
{\partial \alpha(i, j)} 
= \frac {\beta(i, j)} {p(\boldsymbol{y}\ |\ \boldsymbol{x}; \boldsymbol{\theta})}.
\end{equation}  
The derivative of the forward weights with respect to the model parameters at position $j$ is
\begin{align} 
\frac{\partial \alpha(i, j)} {\partial \boldsymbol{\theta}_j} 
= & \frac {\partial p(y_j\ |\ \boldsymbol{y}_1^{j-1}, \boldsymbol{x}_1^i )} {\partial \boldsymbol{\theta}_j} 
\frac{\alpha(i, j)} {p(y_j\ |\ \boldsymbol{y}_1^{j-1}, \boldsymbol{x}_1^i)} \\ \nonumber 
& + p(y_j\ |\  \boldsymbol{y}_1^{j-1}, \boldsymbol{x}_1^i) \sum_{k=1}^i \alpha(k, j-1) 
\frac {\partial}{\partial \boldsymbol{\theta}_j} p(z_j = i\ |\ z_{j-1} = k, \boldsymbol{x}_1^i, \boldsymbol{y}_1^{j-1}). 
\end{align}  
For the geometric distribution transition probability model, we have
\begin{align}
 \frac {\partial} {\partial \boldsymbol{\theta}_j} p(z_j = i\ |\ z_{j-1} = k, \boldsymbol{x}_1^i, \boldsymbol{y}_1^{j-1}) = 0.
\end{align}

\section{Decoding}

\begin{algorithm}                    
	\caption{DP search algorithm}        
	\label{decode} 
	\begin{algorithmic}                    
		\State \textbf{Input: } source sentence $\boldsymbol{x}$
		\State \textbf{Output: } best output sentence $\hat{\boldsymbol{y}}$
		\State \textbf{Initialization: } $Q \in \mathbb{R}^{I \times J_\text{max}}$, bp $\in \mathbb{N}^{I \times J_\text{max}}$,  $W \in \mathbb{N}^{I \times J_\text{max}}$, $I_\text{end} \gets 0$, $J_\text{end} \gets 0$.
		\For{$i \in [1, I]$}
		\State $Q[i,1] \gets \max_{y \in \mathcal{V}}p(z_1 = i) $$p(y\ |\ \text{START}, \boldsymbol{x}_1^i)$
		\State $bp[i,1] \gets 0$
		\State $W[i,1] \gets \argmax_{y \in \mathcal{V}}p(z_1 = i)$$p(y\ |\ \text{START}, \boldsymbol{x}_1^i)$
		\EndFor
		\For{$j\in[2, J_\text{max}]$}
		\For{$i \in [1, I]$}
		\State $Q[i,j] \gets \max_{y \in \mathcal{V}, k \in [1, i]} Q[k,j-1] p(z_j = i\ |\ z_{j-1} = k, \boldsymbol{x}_1^i, \boldsymbol{y}_1^{j-1})p(y\ |\ \boldsymbol{y}_1^{j-1}, \boldsymbol{x}_1^i)$
		\State $bp[i,j] , W[i,j] \gets \argmax_{y \in \mathcal{V}, k \in [1, i]}  \cdot$
		\State $\quad \quad \quad \quad \quad \quad \quad \quad \quad Q[k,j-1] p(z_j = i\ |\ z_{j-1} = k, \boldsymbol{x}_1^i, \boldsymbol{y}_1^{j-1})p(y\ |\ \boldsymbol{y}_1^{j-1}, \boldsymbol{x}_1^i)$
		\EndFor
		\State $I_\text{end} \gets \argmax_{i \in [1, I]} Q[i, j]$
		\If{$W[I_\text{end}, j] = \text{EOS}$ }
		\State $J_\text{end} \gets j$
		\Break
		\EndIf
		\EndFor 
		\If{$J_\text{end} = 0$}
			\State $I_\text{end} = I$
			\State $J_\text{end} = \argmax_{j \in [1, J]}Q[I, j]$
		\EndIf
		\State
		\Return a sequence of words stored in $W$ by following backpointers starting from $(I_\text{end}, J_\text{end})$.
	\end{algorithmic}
\end{algorithm}

For decoding, we aim to find the best output sequence $\hat{\boldsymbol{y}}$ for a given input sequence $\boldsymbol{x}$:
\begin{equation}
\hat{\boldsymbol{y}} = \argmax_{\boldsymbol{y}} p(\boldsymbol{y}\ |\ \boldsymbol{x}).
\end{equation} 
Marginalising the latent variable during search is computationally hard \citep{simaan:1996}, and we approximate the search problem as \citep{jelinek1976continuous,brown:1993}
\begin{align}
\hat{\boldsymbol{y}} = \arg\max_{\boldsymbol{y}} \max_{\boldsymbol{z}} p(\boldsymbol{y}, \boldsymbol{z}\ |\ \boldsymbol{x}).
\end{align}

The search algorithm is based on dynamic programming \citep{tillmann1997dp}. The main idea is to create a path probability matrix $Q$, and fill in each cell $Q[i,j]$ by recursively taking the most probable path that could lead to this cell. 
We present the greedy search algorithm in Algorithm \ref{decode}. 
We also implemented a beam search that tracks the $k$ best partial sequences at position $(i,j)$.
The notation bp refers to backpointers, $W$ stores words to be predicted, $\mathcal{V}$ denotes the output vocabulary, $J_{\text{max}}$ is the maximum length of the output sequences that the model is allowed to predict, $(I_{\text{end}}, J_{\text{end}})$ is the starting point for backtracking.

\section{Experiments}
\label{ssnt_exp}
We evaluate the effectiveness of our model on four representative natural language processing tasks, sentence compression, Chinese-English machine translation, French-English machine translation and morphological inflection. 
The primary aim of this evaluation is to assess whether our proposed architecture is able to outperform the baseline encoder-decoder model by overcoming its encoding bottleneck. We further benchmark our results against an attention model in order to determine whether our alternative alignment strategy is able to provide similar benefits while processing the input online.

\subsection{Abstractive Sentence Summarisation}

Sentence summarisation is the task of generating a condensed version of a sentence while preserving the majority of its meaning. In abstractive sentence summarisation, summaries are generated from the given vocabulary without the constraint of copying words in the input sentence. \cite{DBLP:conf/emnlp/RushCW15} compiled a dataset for this task from the annotated Gigaword dataset \citep{graff2003english,napoles2012annotated}, where sentence-summary pairs are obtained by pairing the headline of each article with its first sentence. 
An example data point is
\begin{itemize}
\item \textbf{Source}: {\it Vietnam will accelerate the export of industrial goods mainly by developing auxiliary industries, and helping enterprises sharpen competitive edges, according to the ministry of industry on thursday.}
\item \textbf{Target}: {\it Vietnam to boost industrial goods export}
\end{itemize}
The data is preprocessed by tokenising, lower casing, replacing digits with `\#', and replacing
of word types seen less than 5 times with UNK. The average lengths of the source sentences and target sentences are 31.2 and 8.3, respectively.  \cite{DBLP:conf/emnlp/RushCW15} use the splits of 3.8m/190k/381k for training, validation and testing. 
In previous work on this dataset, \cite{DBLP:conf/emnlp/RushCW15} proposed an attention-based model with feed-forward neural networks, and \cite{chopra} proposed an attention-based recurrent encoder-decoder, similar to one of our baselines.

Due to computational constraints we place the following restrictions on the training and validation set: 

\begin{enumerate}
	\item The maximum lengths for the input sentences and summaries are 50 and 25, respectively.
	\item For each sentence-summary pair, the product of the input and output lengths should be no greater than 500. 
\end{enumerate}
We use the filtered 172k pairs for validation and sample 1m pairs for training. While this training set is smaller than that used in previous work (and therefore our results cannot be compared directly against reported results), it serves our purpose for evaluating our algorithm against seq2seq and attention-based approaches under identical data conditions.
Following from previous work \citep{DBLP:conf/emnlp/RushCW15,chopra,gulcehre2016pointing}, we report results on a randomly sampled test set of 2000 sentence-summary pairs.  The quality of the generated summaries is evaluated by three versions of ROUGE for different match lengths, namely ROUGE-1 (unigrams), ROUGE-2 (bigrams), and ROUGE-L (longest-common substring).

\begin{table}[t]\centering
	\begin{tabular}{@{}lccc@{}}
		\toprule
		Model &  ROUGE-1 & ROUGE-2  & ROUGE-L \\
		\midrule
		seq2seq 			& 25.16 & 9.09 & 23.06 \\
		attention	  & 30.31 & 14.18 & 28.48 \\
		\midrule
		uniSSNT	           & 27.76 & 11.25 & 25.68 \\
		biSSNT &  28.28 & 11.73 & 26.13\\
		uniSSNT+ & 31.15 & 14.27 & 28.91 \\
		biSSNT+   & \bfseries{31.66} & \bfseries{14.69} & \bfseries{29.32} \\
		\bottomrule
	\end{tabular}
	\caption[ROUGE F1 on the sentence summarisation test set]{ROUGE F1 scores on the sentence summarisation test set. Seq2seq refers to the vanilla encoder-decoder and attention denotes the attention-based model. SSNT denotes our model with alignment transition probability modelled as geometric distribution. SSNT+ refers to our model with transition probability modelled using neural networks. The prefixes uni- and bi- denote using unidirectional and bidirectional encoder LSTMs, respectively.}
	\label{test_rg}
\end{table}

\begin{table}[t]\centering
	\begin{tabular}{@{}lcc@{}}
		\toprule
		Model &  Configuration & Perplexity  \\
		\midrule
		\multirow{4}{*}{seq2seq} 			& $H=128, L=1$ & 48.5  \\
		& $H=256,L=1$ & 35.6 \\
		& $H=256, L=2$ & 32.1 \\
		& $H=256, L=3$ & 31.0 \\
		\midrule
		\multirow{2}{*}{biSSNT+}          & $H=128, L=1$ & 21.2 \\
		& $H=256, L=1$ & \bfseries{20.1} \\
		\bottomrule
	\end{tabular}
	\caption[Perplexity on the sentence summarisation validation set]{Perplexity on the validation set with 172k sentence-summary pairs.}
	\label{perp}
\end{table}

For training, we use Adam \citep{DBLP:journals/corr/KingmaB14} for optimisation, with an initial learning rate of 0.001. The mini-batch size is set to 32. The number of hidden units $H$ is set to 256 for both our model and the baseline models, and dropout of 0.2 is applied to the input and output of LSTMs. All hyperparameters were optimised via grid search on the perplexity of the validation set. We use greedy decoding to generate summaries.

Table \ref{test_rg} displays the ROUGE-F1 scores of our models on the test set, together with baseline models, including the attention-based model.
Our models achieve significantly better results than the vanilla encoder-decoder  and outperform the attention-based model. The fact that SSNT+ performs better than SSNT is in line with our expectations, as the neural network-parameterised alignment model is more expressive than that modelled by geometric distribution. The uniSSNT+ model that is capable of generating outputs online works comparable to biSSNT+.

To make further comparison, we experimented with different sizes of hidden units and adding more layers to the baseline encoder-decoder. Table \ref{perp} lists the configurations of different models and their corresponding perplexities on the validation set. We can see that the vanilla encoder-decoder tends to get better results by adding more hidden units and stacking more layers. This is due to the limitation of compressing information into a fixed-size vector. It has to use larger vectors and deeper structure in order to memorise more information. In contrast, our model can do well with smaller networks. In fact, even with 1 layer and 128 hidden units, our model works much better than the vanilla encoder-decoder with 3 layers and 256 hidden units per layer.

\subsection{Morphological Inflection}
Morphological inflection is the task of generating a target (inflected form) word from a source word (base form), given a morphological attribute, e.g. number, tense, and person etc.. It is useful for alleviating data sparsity issues in translating morphologically rich languages.
The transformation from a base form to an inflected form usually includes concatenating the base form with a prefix or a suffix and substituting some characters. For example, the inflected form of a Finnish stem {\it el\"{a}keik\"{a}} (retirement age) is  {\it el\"{a}keiitt\"{a}} when the case is abessive and the number is plural.  

In our experiments, we use the same dataset as \cite{faruqui2015morphological}. This dataset was originally created by \cite{durrett2013supervised} from Wiktionary, containing  inflections for German nouns (de-N), German verbs (de-V),  Spanish verbs (es-V), Finnish noun and adjective (fi-NA), and Finnish verbs (fi-V). It was further expanded by \cite{nicolai2015inflection} by adding Dutch verbs (nl-V) and French verbs (fr-V). 
The number of inflection types for each language ranges from 8 to 57. The number of base forms, i.e. the number of instances in each dataset, ranges from 2000 to 11200.
The predefined split is 200/200 for dev and test sets, and the rest of the data for training. 

Our model is trained separately for each type of inflection, the same setting as the factored model described in \cite{faruqui2015morphological}. The model is trained to predict the character sequence of the inflected form given that of the stem. The output is evaluated by accuracies of string matching. For all the experiments on this task, we use 128 hidden units for the LSTMs and apply dropout of 0.5 on the input and output of the LSTMs. We use Adam \citep{DBLP:journals/corr/KingmaB14} for optimisation with the initial learning rate of 0.001. During decoding, beam search is employed with the beam size of 30.

\begin{table}[t]\centering
	\begin{tabular}{@{}lr@{}}
		\toprule
		{{Model}} &  {Avg. accuracy}\\
		\midrule
		seq2seq	& 79.08 \\
		seq2seq w/ sttention	 & 95.64  \\
		adapted-seq2seq (FTND16)	  & \bfseries{96.20}  \\
		\midrule
		uniSSNT+ & 89.09 \\
		biSSNT+	 &   95.88\\
		\bottomrule
	\end{tabular}
	\caption[Average accuracy over all the morphological inflection datasets]{Average accuracy over all the morphological inflection datasets. The baseline results for seq2seq variants are taken from \protect\citep{faruqui2015morphological}. }
	\label{avg_acc}
\end{table}

Table \ref{avg_acc} gives the average accuracy of the uniSSNT+, biSSNT+, vanilla encoder-decoder, and attention-based models. The model with the best previous average result --- denoted as adapted-seq2seq (FTND16) \citep{faruqui2015morphological} --- is also included for comparison.  Our biSSNT+ model outperforms the vanilla encoder-decoder by a large margin and almost matches the state-of-the-art result on this task. As mentioned earlier, a characteristic of these datasets is that the stems and their corresponding inflected forms mostly overlap. Compare to the vanilla encoder-decoder, our model is better at copying and finding correspondences between prefix, stem and suffix segments.

\begin{table}[t]\centering
	\begin{tabular}{@{}lcccc@{}}
		\toprule
		Dataset & DDN13 & NCK15 & FTND16 & biSSNT+ \\
		\midrule
		de-N	& 88.31 & \bfseries{88.60}  & 88.12 & 87.69\\
		de-V	 & 94.76 & 97.50  & \bfseries{97.72} & 94.98 \\
		es-V	  & 99.61 & 99.80  & \bfseries{99.81} & 99.69\\
		fi-NA	 & 92.14 & 93.00  & 95.44 & \bfseries{95.48}\\
		fi-V     & 97.23 & 98.10 & 97.81 & \bfseries{98.17}\\
		fr-V    & 98.80 & \bfseries{99.20}  & 98.82 & 98.97 \\
		nl-V    & 90.50 & 96.10 & 96.71 & 96.19\\ 
		\midrule
		Avg.    & 94.47 & 96.04  & \bfseries{96.20} & 95.88 \\
		\bottomrule
	\end{tabular}
	\caption[Results on morohological inflection]{Comparison of the performance of our model (biSSNT+) against the previous state-of-the-art on each morphological inflection dataset. DDN13, NCK15, and FTND16 denote the models of \cite{durrett2013supervised}, \cite{nicolai2015inflection}, and \cite{faruqui2015morphological}, respectively. }
	\label{sep_acc}
\end{table}

Table \ref{sep_acc} compares the results of biSSNT+ and previous models on each individual dataset. DDN13 and NCK15 denote the models of \cite{durrett2013supervised} and \cite{nicolai2015inflection}, respectively. Both models tackle the task by feature engineering. FTND16 \citep{faruqui2015morphological} adapted the vanilla encoder-decoder by feeding the $i$-th character of the encoded string as an extra input into the $i$-th position of the decoder. It can be considered as a special case of our model by forcing a fixed diagonal alignment between input and output sequences. 
Our model achieves comparable results to these models on all the datasets. Notably, it outperforms other models on the datasets of Finnish noun and adjective, and Finnish verbs, whose stems and inflected forms are the longest.

\subsection{Chinese-English Machine Translation}
\label{sec:zh-en-mt}
 In this and the next sections, we examine the performance of our models in machine translation, which has been the most widely used task for testing seq2seq models \citep{cho2014learning,kalchbrenner2013recurrent,sutskever2014sequence,bahdanau2014neural}. We have demonstrated that SSNT is very effective on the tasks of abstractive sentence summarisation and morphological inflection, where the alignment between sequence pairs are largely monotonic. We experiment on a Chinese-English machine translation task here in order to see the performance of our model when word orders diverge. 
 
  We use parallel data with 184k sentence pairs from the FBIS corpus, LDC2003E14. The training data is preprocessed by lowercasing the English sentences, replacing digits with `\#' token, and replacing tokens appearing less than 5 times with an UNK token. This results in vocabulary sizes of 30k and 20k for Chinese sentences and English sentences, respectively. The maximum input and output lengths are restricted to 50.
 
 For the experimental setup,  uniSSNT and biSSNT are trained using Adam \citep{DBLP:journals/corr/KingmaB14} with the initial learning rate of 0.001.  The size of LSTM hidden units is 512. Dropout of 0.5 is set on the input and output of LSTMs. Gradients are clipped if the norm exceeds 5. To set benchmarks, we train the vanilla and attentional seq2seq models using the same parallel data. While the setup for the attentional model is the same as SSNT, the setup for the vanilla seq2seq model is slightly different: it contains 2 layers,\footnote{Hyperparameters tuning shows that this setup for the vanilla seq2seq model works best.} with 512 hidden units per layer. For decoding, we leverage beam search with beam size 10.

\begin{table}[t]\centering
\begin{tabular}{@{}lrr@{}}
\toprule
{{Model}} &  {Dev Perplexity}  &  {Test BLEU} \\
\midrule
seq2seq	 w/o attention & 36.4  & 11.19  \\
seq2seq w/ attention	 & \bfseries{19.3} & \bfseries{25.27} \\
\midrule
uniSSNT+ & 23.9 & 18.25 \\
biSSNT+	 & 20.4  & 23.33 \\
\bottomrule
\end{tabular}
\caption[Results of Chinese to English translation]{Perplexities (on the development set) and BLEU scores (on the test set) from different models for the Chinese to English machine translation task. }
\label{zh-en-bleu-ppl}
\end{table}

Table \ref{zh-en-bleu-ppl} lists the perplexities and BLEU scores from different models.
We can see that still SSNT outperforms the vanilla seq2seq model significantly. However, unlike sentence summarisation, it is not better than the attentional seq2seq model on the Chinese to English machine translation task. 

\subsection{French-English Machine Translation}
We have seen in the previous section that although both the attentional seq2seq model and SSNT work relatively well, SSNT is not superior to the attentional seq2seq model on the Chinese-English machine translation task. We conjecture that this is not because SSNT is not appropriate for machine translation, but because word orders in Chinese-English sentence pairs diverge significantly, in which case SSNT may be less dominant due to the monotonic assumption. We validate our hypothesis by experimenting on the French-English machine translation, where the alignment between input and output sentences are largely monotonic. 

The parallel dataset for training and evaluating the models is from the MT track of IWSLT 2016.\footnote{\url{http://workshop2016.iwslt.org/}} The dataset is collected from transcripts and translations of TED talks. We preprocessed the dataset by lowercasing, replacing the infrequent words with `UNK', and filtering out sentences longer than 50. The train/dev split is 210k/65k. We used the TED test files of 2014 to evaluate the translations, which contains 1300 French sentences.

The experimental setup is the same as that in Section \ref{sec:zh-en-mt}. Table \ref{tb:fr-en-bleu} provides the BLEU scores from different models. The fact that biSSNT+ surpasses the other models indicates that SSNT is suitable for machine translation, and it biases towards sentence pairs with monotonic alignment.  

\begin{table}[t]\centering
\begin{tabular}{@{}lr@{}}
\toprule
{{Model}} &  BLEU \\
\midrule
seq2seq	 w/o attention &  15.16   \\
seq2seq w/ attention	 & 32.28  \\
\midrule
uniSSNT+ &  28.93 \\
biSSNT+	 &  \bfseries{33.55} \\
\bottomrule
\end{tabular}
\caption[Results of French to English translation]{BLEU scores  from different models for the French to English machine translation task. }
\label{tb:fr-en-bleu}
\end{table}

\section{Alignment Quality}
\label{ssnt_analysis}

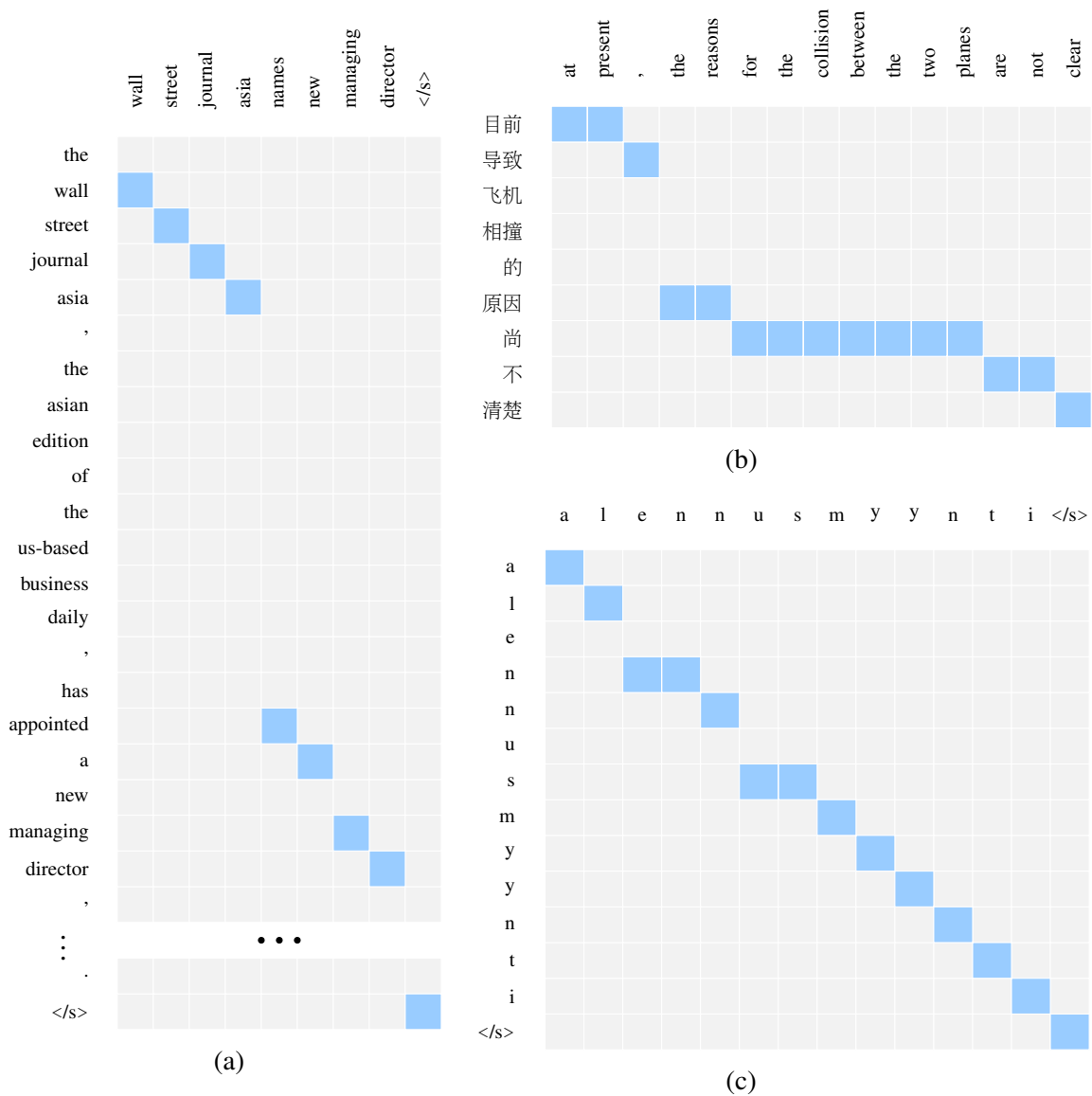
\begin{figure*}[!htb]
	\begin{minipage}{0.42\textwidth}
		\begin{subfigure}{\linewidth}
			\begin{tikzpicture}
			\def \xscale {0.5}
			\def \yscale {0.5}
			\definecolor{color2}{RGB}{153,204,255}
			
			\foreach \c [count=\x from 0] in {
				{\texttt{<}/s\texttt{>}},{.},{},{,},{director},{managing},{new},{a},{appointed},{has},{,},{daily},{business},{us-based},{the},{of},{edition},{asian},{the},{,},{asia},{journal},{street},{wall},{the} } 
			{
				\node [left] at (-\xscale,\x*\yscale) {\scriptsize \c};	
			}

			\foreach \c [count=\x from 0] in {
				{wall},{street},{journal},{asia}, {names},{new}, {managing},{director},{\texttt{<}/s\texttt{>}}}  
			{
				\node [rotate=90,right] at (\xscale*\x,25*\yscale) {\scriptsize \c};	
			} 
			
			\foreach \x in {0,...,8}
			\foreach \y in {0,1,3,4,...,24} 
			{
				\draw [white,fill=black!5]
				(\x*\xscale-0.5*\xscale,\y*\yscale-0.5*\yscale) rectangle (\x*\xscale+0.5*\xscale,\y*\yscale+0.5*\yscale);
			} 
			
			\node  at (-2*\xscale,2*\yscale) {\vdots};	
			\draw [fill]  
			(3.5*\xscale, 2*\yscale) circle [radius=0.04] 
			(4*\xscale, 2*\yscale) circle [radius=0.04]
			(4.5*\xscale, 2*\yscale) circle [radius=0.04]
			;	
			
			\draw [white,fill=color2] 
			(-0.5*\xscale,22.5*\yscale) rectangle (0.5*\xscale,23.5*\yscale)
			(0.5*\xscale,21.5*\yscale) rectangle (1.5*\xscale,22.5*\yscale)
			(1.5*\xscale,20.5*\yscale) rectangle (2.5*\xscale,21.5*\yscale)
			(2.5*\xscale,19.5*\yscale) rectangle (3.5*\xscale,20.5*\yscale)
			(3.5*\xscale,7.5*\yscale) rectangle (4.5*\xscale,8.5*\yscale)
			(4.5*\xscale,6.5*\yscale) rectangle (5.5*\xscale,7.5*\yscale)
			(5.5*\xscale,4.5*\yscale) rectangle (6.5*\xscale,5.5*\yscale)
			(6.5*\xscale,3.5*\yscale) rectangle (7.5*\xscale,4.5*\yscale)
			(7.5*\xscale,-0.5*\yscale) rectangle (8.5*\xscale,0.5*\yscale)
			;
			
			\end{tikzpicture}
			\caption{}
			\label{vis1}
		\end{subfigure}
	\end{minipage}    
	\begin{minipage}{.5\textwidth} 
		\begin{subfigure}{\linewidth}
			
			\begin{tikzpicture}
			\def \xscale {0.5}
			\def \yscale {0.5}
			\definecolor{color2}{RGB}{153,204,255}
			
			\foreach \c [count=\x from 0] in {
				\begin{CJK*}{UTF8}{gbsn}
				清楚
				\end{CJK*},
				\begin{CJK*}{UTF8}{gbsn}
				不
				\end{CJK*},
				\begin{CJK*}{UTF8}{gbsn}
				尚
				\end{CJK*},
				\begin{CJK*}{UTF8}{gbsn}
				原因
				\end{CJK*},
				\begin{CJK*}{UTF8}{gbsn}
				的
				\end{CJK*},
				\begin{CJK*}{UTF8}{gbsn}
				相撞
				\end{CJK*},
				\begin{CJK*}{UTF8}{gbsn}
				飞机
				\end{CJK*},
				\begin{CJK*}{UTF8}{gbsn}
				导致
				\end{CJK*},
				\begin{CJK*}{UTF8}{gbsn}
				目前
				\end{CJK*},
			} 
			{
				\node [left] at (-\xscale,\x*\yscale) {\scriptsize \c};	
			}

			\foreach \c [count=\x from 0] in {
				{at},{present},{,},{the}, {reasons},{for}, {the},{collision},{between},{the},{two},{planes},{are},{not},{clear}}  
			{
				\node [rotate=90,right] at (\xscale*\x,9*\yscale) {\scriptsize \c};	
			} 
			
			\foreach \x in {0,...,14}
			\foreach \y in {0,...,8} 
			{
				\draw [white,fill=black!5]
				(\x*\xscale-0.5*\xscale,\y*\yscale-0.5*\yscale) rectangle (\x*\xscale+0.5*\xscale,\y*\yscale+0.5*\yscale);
			} 
			
			
			\draw [white,fill=color2] 
			(-0.5*\xscale,7.5*\yscale) rectangle (0.5*\xscale,8.5*\yscale)
			(0.5*\xscale,7.5*\yscale) rectangle (1.5*\xscale,8.5*\yscale)
			(1.5*\xscale,6.5*\yscale) rectangle (2.5*\xscale,7.5*\yscale)
			(2.5*\xscale,2.5*\yscale) rectangle (3.5*\xscale,3.5*\yscale)
			(3.5*\xscale,2.5*\yscale) rectangle (4.5*\xscale,3.5*\yscale)
			(4.5*\xscale,1.5*\yscale) rectangle (5.5*\xscale,2.5*\yscale)
			(5.5*\xscale,1.5*\yscale) rectangle (6.5*\xscale,2.5*\yscale)
			(6.5*\xscale,1.5*\yscale) rectangle (7.5*\xscale,2.5*\yscale)
			(7.5*\xscale,1.5*\yscale) rectangle (8.5*\xscale,2.5*\yscale)
			(8.5*\xscale,1.5*\yscale) rectangle (9.5*\xscale,2.5*\yscale)
			(9.5*\xscale,1.5*\yscale) rectangle (10.5*\xscale,2.5*\yscale)
			(10.5*\xscale,1.5*\yscale) rectangle (11.5*\xscale,2.5*\yscale)
			(11.5*\xscale,0.5*\yscale) rectangle (12.5*\xscale,1.5*\yscale)
			(12.5*\xscale,0.5*\yscale) rectangle (13.5*\xscale,1.5*\yscale)
			(13.5*\xscale,-0.5*\yscale) rectangle (14.5*\xscale,0.5*\yscale)
			;
			
			\end{tikzpicture}
			\caption{}
			\label{vis2}
		\end{subfigure}\\[1ex]
		
		\begin{subfigure}{\linewidth}
			\begin{tikzpicture}
			\def \xscale {0.54}
			\def \yscale {0.5}
			\definecolor{color2}{RGB}{153,204,255}
			
			\foreach \c [count=\x from 0] in {
				{\texttt{<}/s\texttt{>}},{i},{t},{n},{y},{y},{m},{s},{u},{n}, {n},{e},{l},{a}} 
			{
				\node [left] at (-\xscale,\x*\yscale) {\scriptsize \c};	
			}

			\foreach \c [count=\x from 0] in {
				{a},{l},{e},{n}, {n},{u}, {s},{m},{y},{y},{n},{t},{i},{\texttt{<}/s\texttt{>}}}  
			{
				\node [above] at (\xscale*\x,14*\yscale) {\scriptsize \c};	
			} 
			
			\foreach \x in {0,...,13}
			\foreach \y in {0,...,13} 
			{
				\draw [white,fill=black!5]
				(\x*\xscale-0.5*\xscale,\y*\yscale-0.5*\yscale) rectangle (\x*\xscale+0.5*\xscale,\y*\yscale+0.5*\yscale);
			}

			\draw [white,fill=color2] 
			(-0.5*\xscale,12.5*\yscale) rectangle (0.5*\xscale,13.5*\yscale)
			(0.5*\xscale,11.5*\yscale) rectangle (1.5*\xscale,12.5*\yscale)
			(1.5*\xscale,9.5*\yscale) rectangle (2.5*\xscale,10.5*\yscale)
			(2.5*\xscale,9.5*\yscale) rectangle (3.5*\xscale,10.5*\yscale)
			(3.5*\xscale,8.5*\yscale) rectangle (4.5*\xscale,9.5*\yscale)
			(4.5*\xscale,6.5*\yscale) rectangle (5.5*\xscale,7.5*\yscale)
			(5.5*\xscale,6.5*\yscale) rectangle (6.5*\xscale,7.5*\yscale)
			(6.5*\xscale,5.5*\yscale) rectangle (7.5*\xscale,6.5*\yscale)
			(7.5*\xscale,4.5*\yscale) rectangle (8.5*\xscale,5.5*\yscale)
			(8.5*\xscale,3.5*\yscale) rectangle (9.5*\xscale,4.5*\yscale)
			(9.5*\xscale,2.5*\yscale) rectangle (10.5*\xscale,3.5*\yscale)
			(10.5*\xscale,1.5*\yscale) rectangle (11.5*\xscale,2.5*\yscale)
			(11.5*\xscale,0.5*\yscale) rectangle (12.5*\xscale,1.5*\yscale)
			(12.5*\xscale,-0.5*\yscale) rectangle (13.5*\xscale,0.5*\yscale)
			;
			
			\end{tikzpicture}
			\caption{}
			\label{vis3}
		\end{subfigure}
		
	\end{minipage}
	\caption[Example alignments found by the neural transduction model]{Example alignments found by our models. Figure (a) and (c) are outputs generated by biSSNT+, and Figure (b) is the output of uniSSNT+. Highlighted grid cells represent the correspondence between the input and output tokens.}
	\label{vis}
\end{figure*}

\begin{figure}
	\centering
	\includegraphics[scale=0.68]{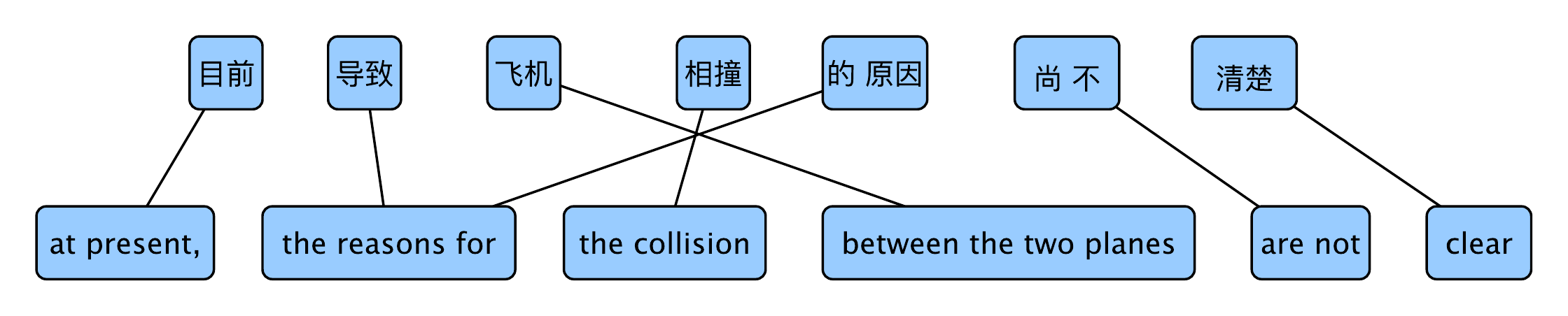}
	\caption[Gold output and alignment]{Gold output and alignment for an example in Chinese-English translation.}
	\label{fig:align1}
\end{figure}

\begin{figure*}[!htb]
	\begin{subfigure}{0.4\linewidth}
		\begin{tikzpicture}
		\def \xscale {0.5}
		\def \yscale {0.5}
		\definecolor{color2}{RGB}{153,204,255}
		
		\foreach \c [count=\x from 0] in {
			{\texttt{<}/s\texttt{>}},	
			\begin{CJK*}{UTF8}{gbsn}
			。
			\end{CJK*},
			\begin{CJK*}{UTF8}{gbsn}
			之一
			\end{CJK*},
			\begin{CJK*}{UTF8}{gbsn}
			国家
			\end{CJK*},
			\begin{CJK*}{UTF8}{gbsn}
			的
			\end{CJK*},
			\begin{CJK*}{UTF8}{gbsn}
			贫穷
			\end{CJK*},
			\begin{CJK*}{UTF8}{gbsn}
			最
			\end{CJK*},
			\begin{CJK*}{UTF8}{gbsn}
			世界
			\end{CJK*},
			\begin{CJK*}{UTF8}{gbsn}
			是
			\end{CJK*},
			\begin{CJK*}{UTF8}{gbsn}
			尼泊尔
			\end{CJK*}
		} 
		{
			\node [left] at (-\xscale,\x*\yscale) {\scriptsize \c};	
		}

		\foreach \c [count=\x from 0] in {
			{nepal},{is},{the},{most}, {poor},{country}, {.},{\texttt{<}EOS\texttt{>}}
		}  
		{
			\node [rotate=90,right] at (\xscale*\x,10*\yscale) {\scriptsize \c};	
		} 
		
		\foreach \x in {0,...,7}
		\foreach \y in {0,...,9} 
		{
			\draw [white,fill=black!5]
			(\x*\xscale-0.5*\xscale,\y*\yscale-0.5*\yscale) rectangle (\x*\xscale+0.5*\xscale,\y*\yscale+0.5*\yscale);
		} 
		
		\draw [white,fill=color2] 
		(-0.5*\xscale,7.5*\yscale) rectangle (0.5*\xscale,8.5*\yscale)
		(0.5*\xscale,7.5*\yscale) rectangle (1.5*\xscale,8.5*\yscale)
		(1.5*\xscale,5.5*\yscale) rectangle (2.5*\xscale,6.5*\yscale)
		(2.5*\xscale,5.5*\yscale) rectangle (3.5*\xscale,6.5*\yscale)
		(3.5*\xscale,4.5*\yscale) rectangle (4.5*\xscale,5.5*\yscale)
		(4.5*\xscale,1.5*\yscale) rectangle (5.5*\xscale,2.5*\yscale)
		(5.5*\xscale,-0.5*\yscale) rectangle (6.5*\xscale,0.5*\yscale)
		(6.5*\xscale,-0.5*\yscale) rectangle (7.5*\xscale,0.5*\yscale)
		;
		
		\end{tikzpicture}
		\caption{}
		\label{uni_vis}
	\end{subfigure}	
	\begin{subfigure}{0.6\linewidth}
		\begin{tikzpicture}
		\def \xscale {0.5}
		\def \yscale {0.5}
		\definecolor{color2}{RGB}{153,204,255}
		
		\foreach \c [count=\x from 0] in {
			{\texttt{<}/s\texttt{>}},	
			\begin{CJK*}{UTF8}{gbsn}
			。
			\end{CJK*},
			\begin{CJK*}{UTF8}{gbsn}
			之一
			\end{CJK*},
			\begin{CJK*}{UTF8}{gbsn}
			国家
			\end{CJK*},
			\begin{CJK*}{UTF8}{gbsn}
			的
			\end{CJK*},
			\begin{CJK*}{UTF8}{gbsn}
			贫穷
			\end{CJK*},
			\begin{CJK*}{UTF8}{gbsn}
			最
			\end{CJK*},
			\begin{CJK*}{UTF8}{gbsn}
			世界
			\end{CJK*},
			\begin{CJK*}{UTF8}{gbsn}
			是
			\end{CJK*},
			\begin{CJK*}{UTF8}{gbsn}
			尼泊尔
			\end{CJK*}
		} 
		{
			\node [left] at (-\xscale,\x*\yscale) {\scriptsize \c};	
		}

		\foreach \c [count=\x from 0] in {
			{nepal},{is},{one},{of},{the}, {poorest},{countries},{in},{the},{world}, {.},{\texttt{<}/s\texttt{>}}
		}  
		{
			\node [rotate=90,right] at (\xscale*\x,10*\yscale) {\scriptsize \c};	
		} 
		
		\foreach \x in {0,...,11}
		\foreach \y in {0,...,9} 
		{
			\draw [white,fill=black!5]
			(\x*\xscale-0.5*\xscale,\y*\yscale-0.5*\yscale) rectangle (\x*\xscale+0.5*\xscale,\y*\yscale+0.5*\yscale);
		} 
		
		\draw [white,fill=color2] 
		(-0.5*\xscale,8.5*\yscale) rectangle (0.5*\xscale,9.5*\yscale)
		(0.5*\xscale,7.5*\yscale) rectangle (1.5*\xscale,8.5*\yscale)
		(1.5*\xscale,7.5*\yscale) rectangle (2.5*\xscale,8.5*\yscale)
		(2.5*\xscale,7.5*\yscale) rectangle (3.5*\xscale,8.5*\yscale)
		(3.5*\xscale,6.5*\yscale) rectangle (4.5*\xscale,7.5*\yscale)
		(4.5*\xscale,4.5*\yscale) rectangle (5.5*\xscale,5.5*\yscale)
		(5.5*\xscale,2.5*\yscale) rectangle (6.5*\xscale,3.5*\yscale)
		(6.5*\xscale,0.5*\yscale) rectangle (7.5*\xscale,1.5*\yscale)
		(7.5*\xscale,0.5*\yscale) rectangle (8.5*\xscale,1.5*\yscale)
		(8.5*\xscale,0.5*\yscale) rectangle (9.5*\xscale,1.5*\yscale)
		(9.5*\xscale,-0.5*\yscale) rectangle (10.5*\xscale,0.5*\yscale)
		(10.5*\xscale,-0.5*\yscale) rectangle (11.5*\xscale,0.5*\yscale)
		;
		
		\end{tikzpicture}
		\caption{}
		\label{bi_vis}
	\end{subfigure}\\[1ex]
   
	\begin{subfigure}{\linewidth}
		\centering
		\includegraphics[scale=0.68]{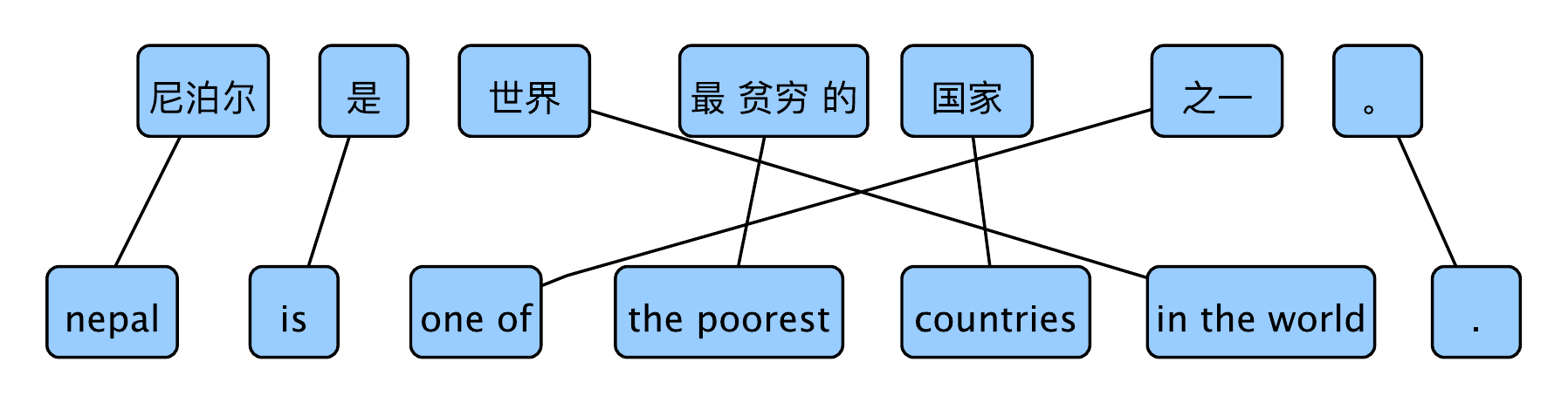}
		\caption{}
		\label{fig:align2}
	\end{subfigure}
	
	\caption[Comparison between alignments]{Comparison between the outputs generated by uniSSNT+ (a) and biSSNT+ (b). Figure (c) is the gold translation together with the true alignments between the sentence pairs.}
	\label{uni_bi_vis}
\end{figure*}
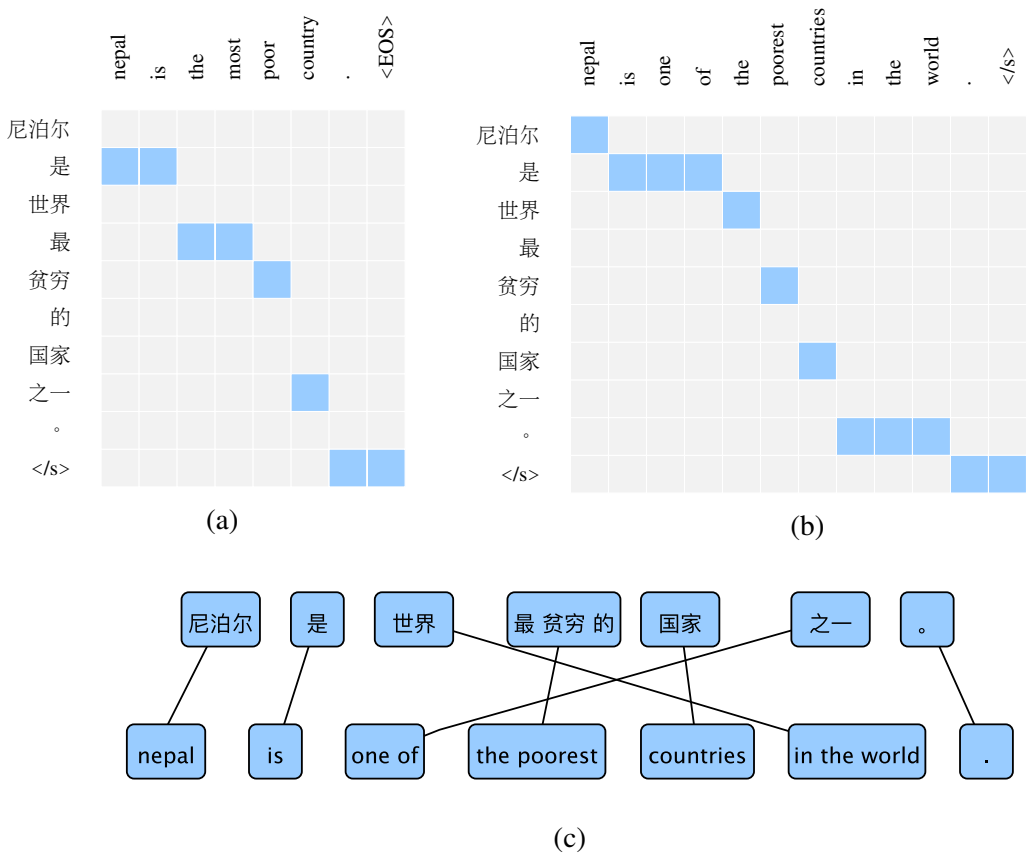

\begin{figure}
	\centering
	\includegraphics[scale=0.57]{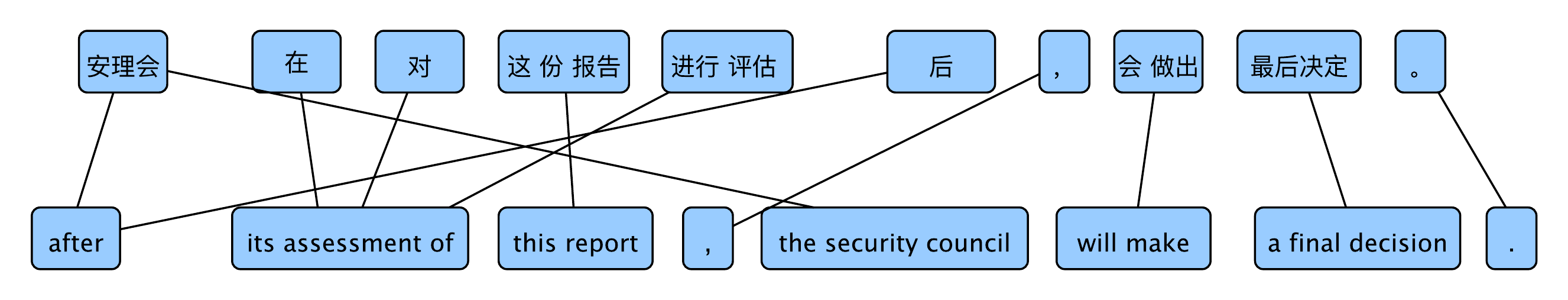}
	\caption[Example of Chinese-English translation]{Example of Chinese-English translation, in which biSSNT+ fails to generate a reasonable translation, whereas the attentional seq2seq model succeeds. }
	\label{fig:align3}
\end{figure}

\begin{figure}
	\centering
	\includegraphics[scale=0.68]{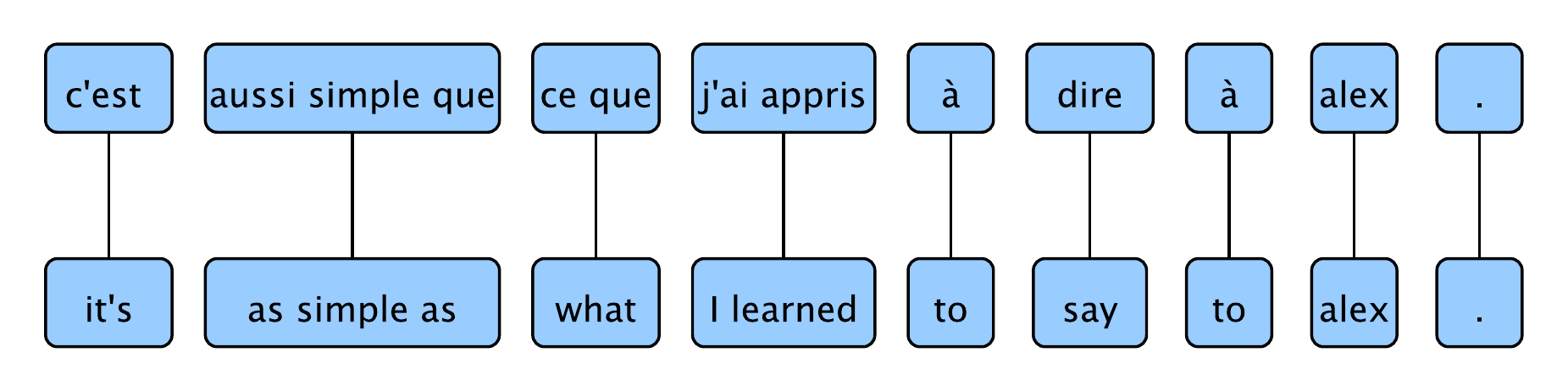}
	\caption[Example of French-English translation]{Example of French-English translation, in which biSSNT+ generates a more precise translation than the attentional seq2seq model.}
	\label{fig:align4}
\end{figure}
Figure \ref{vis} presents visualisations of segment alignments generated by our model for sample instances from the tasks of sentence summarisation, Chinese-English machine translation, and morphological inflection. We see that the model is able to learn the correct correspondences between segments of the input and output sequences. For instance, the alignment follows a nearly diagonal path for the example in Figure \ref{vis3}, where the input and output sequences are identical. We observe that the model is robust on long phrasal mappings and local word reorderings. As shown in Figure \ref{vis1}, the mapping between `the wall street journal asia, the asian edition of the us-based business daily' and `wall street journal asia' demonstrates that our model learns to ignore phrasal modifiers containing additional information. Figure \ref{vis2} presents how the models deal with word reorderings. As shown in the gold alignments of the example in Figure \ref{fig:align1}, word orders of the Chinese phrase 
\begin{CJK*}{UTF8}{gbsn}
导致\ 飞机\ 相撞\ 的\ 原因
\end{CJK*} and the English phrase `the reasons for the collision between the two planes' are largely different. In particular, the three tokens \begin{CJK*}{UTF8}{gbsn}
导致\  的\ 原因
\end{CJK*}, which are separated by two other tokens, correspond to the continuous phrase `the reasons for'. We see in Figure \ref{vis2} that when reading the input token \begin{CJK*}{UTF8}{gbsn}
导致
\end{CJK*}, the model knows that this is an incomplete phrase and it keeps reading more tokens until it sees the token \begin{CJK*}{UTF8}{gbsn}
原因
\end{CJK*} and starts emitting the translation for the entire phrase. This example also validates our hypothesis that the model learns to fall back to vanilla encoder-decoders when word orders of the sentence pairs are diverse. 

However, uniSSNT+ may fail in cases where it cannot detect whether the current phrase is finished or not without reading future tokens. For instance, in Figure \ref{uni_bi_vis} the last token in the Chinese sentence corresponds to `one of' in English. Since uniSSNT+ cannot foresee the last input token, it just keeps translating the prefixes of the input sentence, resulting in the output of `nepal is the most poor country'. This problem is avoided by biSSNT+, attribute to the encoder modelled with a bidirectional LSTM. It generates the phrase `one of' in advance at the right position.

By examining the output of biSSNT+ in comparison with the attentional seq2seq model,  we find that biSSNT+ does worse than the attentional seq2seq model in sentence pairs where word orders are dramatically inconsistent (Figure \ref{fig:align3}). Our models, however, generate more precise output when the alignments between sentences are largely monotonic. As shown in Figure \ref{fig:align4}, biSSNT+ can capture the correspondence between phrases perfectly, but the seq2seq model with attention does not generate the tokens of `say to'.   These findings are consistent with the experimental results that the attentional seq2seq model works better than SSNT on Chinese-English translation but worse on the other three tasks.

\section{Summary}
This chapter focuses on the generation tasks of seq2seq mapping. 
We have proposed a novel segment to segment neural transduction model that tackles the limitations of vanilla encoder-decoders that have to read and memorise an entire input sequence in a fixed-length context vector before producing any output. By introducing a latent segmentation that determines correspondences between prefixes of the input and output sequences, our model learns to generate and align jointly. During training, the hidden alignment is marginalised out using dynamic programming, and during decoding the best alignment path is generated alongside the predicted output sequence. By employing a unidirectional LSTM as encoder, our model is capable of doing online generation.  Experiments on four representative natural language processing tasks, abstractive sentence summarisation, morphological inflection generation, Chinese-English machine translation, and French-English machine translation showed that our model significantly outperformed encoder-decoder baselines while requiring much smaller hidden layers.  Qualitative analysis showed that the alignment paths generated by our model were highly intuitive.

Although the model performs very well in general, there are several aspects that it does not work as well as expected and some parts of model design could be improved.

 First, to enable the model to perform online generation, we have constrained the alignments to be monotone. A side effect of this monotonic assumption is that the model biases towards sequence pairs with largely monotonic alignments. We can observe the evidence from the experimental results of Chinese-English translation versus the other three tasks. 
Qualitative analysis demonstrated that when tackling reorderings the model is capable of choosing how much input to read ahead before generating each segment, especially when there is a strong indicator showing at what position the current segment is complete. However, the model cannot always find the precise segmentation, and sometimes it prefers generating an output token immediately even when it sees an incomplete segment. A possible piece of future work is to improve the model so that it can predict output incrementally without deteriorating its performance when there are dramatic reorderings.

Second, in the current model design, in order to derive exact inference, we have made several independent assumptions in terms of word generation and alignment decisions.
To realise these independences, we let the input and output RNNs update their hidden states independently and the outputs of the two RNNs are combined only for  word and alignment probability calculations (``late fusion''). It would be interesting to investigate to what extent the model's performance could be improved by relaxing these independent assumptions, e.g. by switching from ``late fusion'' to ``early fusion'' (although this implies approximate inference). 

Finally, the model is computationally expensive. The bottleneck comes from the computation of {\it softmax}es. Unlike seq2seq models with/without attention which have $J$  {\it softmax}es, our model requires to calculate $I \times J$ {\it softmax}es, where $I, J$ denote the number of tokens in the input and output sequences, respectively. Additionally, it is difficult to do dynamic programming over mini-batches in a vectorised way, and thus the implementation of inference algorithms is also slow. In order to scale up the model for bigger datasets, we should address the computational constraints both from the theoretical and practical aspects.


%% file: chapter_6.tex
\chapter{Incorporating Unpaired Data}
\label{ch:noisy_channel}

\begin{chapterabstract}
In the previous chapters, we have shown that recurrent neural network seq2seq models are excellent models for the generation tasks of seq2seq mapping, provided sufficient input--output pairs are available for estimating their parameters. However, in many domains, vastly more unpaired  examples are available than input--output pairs. In this chapter, we investigate the incorporation of additional unpaired data for solving seq2seq mapping problems.
We formulate seq2seq mapping as a noisy channel decoding problem and use recurrent neural networks to parameterise the source and channel models. 
The component models of noisy channel can be trained with not only paired training samples but also unpaired samples from the marginal output distribution. 
Using a latent variable to control how much of the conditioning sequence the channel model needs to read in order to generate a subsequent symbol, we obtain a tractable and effective beam search decoder. Experimental results on abstractive sentence summarisation, morphological inflection, and machine translation show that noisy channel models  outperform direct models, and that they significantly benefit from increased amounts of unpaired output data that direct models cannot easily use.\let\thefootnote\relax\footnote{The material in this chapter was originally presented in \cite{yu:2017}.} 
\addtocounter{footnote}{-1}\let\thefootnote\svthefootnote
\end{chapterabstract}

\section{Introduction}
Data plays an indispensable role in machine learning. In theory, the more accurately annotated data is available, the better a machine learning algorithm can learn. For example, in the task of image recognition, models will perform significantly better if they are fed with millions of images with human-annotated labels than those with thousands of labelled data will. This claim is particularly true for neural models, which are extremely data hungry. However, in practice, it is often infeasible to obtain labels for all examples in large datasets, as high-quality annotations usually involve human effort which is very expensive and time-consuming. By contrast, unlabelled examples are  easy to obtain in large quantities, and can outnumber the amount of available labelled data substantially. For example, in the seq2seq mapping problems, transcribed speech is relatively rare although non-spoken texts are abundant; Swahili--English translations are rare although English texts are abundant; etc.. It is therefore useful to investigate how to effectively integrate abundant unlabelled and limited labelled data to build better and more accurate models. There are also biological motivations for studying the process of learning from partially labelled data. Children learn language mainly by listening and imitating, with limited feedback from adults. Humans also excel at other partially labelled learning tasks, such as visual discrimination with hyperacuity.


Recurrent network seq2seq models ~\citep{kalchbrenner2013recurrent,sutskever2014sequence,bahdanau2014neural}, including the segment to segment neural transduction model (SSNT) that we described in the previous chapter, have achieved promising results in modelling $p(\text{output sequence}\ \boldsymbol{y} \mid \text{input sequence}\ \boldsymbol{x})$. However, the data used for training these models is only limited to input-output $(\boldsymbol{x}, \boldsymbol{y})$  pairs. This limitation makes them uncompetitive in domains where the available paired data is not sufficient. In this chapter, we overcome the limitation by exploiting both the paired and unpaired data (in particular unpaired output examples). 

Our strategy is to use Bayes' rule to rewrite $p(\boldsymbol{y} \mid \boldsymbol{x})$ as $p(\boldsymbol{x} \mid \boldsymbol{y}) p(\boldsymbol{y})/p(\boldsymbol{x})$, a factorisation which is called a \emph{noisy channel model}~\citep{shannon:1948}. A noisy channel model thus consists of two component models: the conditional \emph{channel model}, $p(\boldsymbol{x} \mid \boldsymbol{y})$, which characterises the \emph{reverse} transduction problem and whose parameters are estimated from the paired $(\boldsymbol{x},\boldsymbol{y})$ samples, and the unconditional \emph{source model}, $p(\boldsymbol{y})$, whose parameters are estimated from both the paired and (usually much more numerous) unpaired samples.

Despite the capability of exploiting both kinds of data, the noisy channel model also has the advantage of combining two separate models. 
By applying them together we hope to counterbalance the errors generated from each separate model. 
Furthermore, unlike direct models which can suffer from explaining-away\footnote{In a graphical model, explaining away refers to a reasoning pattern in which there are multiple causes behind a single effect and these causes interact with each other. When one of these causes of the effect is observed, the probabilities of other causes will decrease \citep[Chapter 3]{DBLP:books/daglib/0023091}. In seq2seq mapping, the prediction of the next token is conditioned on the preceding output tokens and the input sequence: $p(y_j\ |\  \boldsymbol{y}_1^{j-1}, \boldsymbol{x})$. When modelling sequence pairs, it has been observed that neural seq2seq models (e.g., encoder-decoders and those with attention) tend to depend too much on the highly predictive output prefix and ignore the input \citep{tu:2016,huang2018}. In the case of machine translation, this means that the translated sentence may be fluent but not adequate, meaning it may not convey the meaning of the input sentence. } effects during training, noisy channel models must produce outputs that explain their inputs.

In principle, the noisy channel decomposition is straightforward; however, in practice, decoding (i.e., computing $\arg \max_{\boldsymbol{y}} p(\boldsymbol{x} \mid \boldsymbol{y}) p(\boldsymbol{y})$) is a significant computational challenge, and tractability concerns impose restrictions on the form the component models can take. To illustrate, an appealing parameterisation would be to use an attentional seq2seq network \citep{bahdanau2014neural} to model the channel probability $p(\boldsymbol{x} \mid \boldsymbol{y})$. However, seq2seq models are designed under the assumption that the complete conditioning sequence is available before any prefix probabilities of the output sequence can be computed. This assumption is problematic for channel models since it means that a complete output sequence must be constructed before the channel model can be evaluated (since the channel model conditions on the output). Therefore, to be practical, the channel probability must decompose in terms of prefixes of the conditioning variable, $\boldsymbol{y}$. While the chain rule justifies decomposing output variable probabilities in terms of successive extensions of a partial prefix, no such convenience exists for conditioning variables, and approximations must be introduced.

In this work, we use our proposed SSNT (described in Chapter \ref{ch:ssnt})  which uses a latent alignment variable to enable its probabilities to factorise in terms of prefixes of both the input and output, making it an appropriate channel model~(\S\ref{sec:model}). Using this channel model, the decoding problem then becomes similar to the problem faced when decoding with direct models~(\S\ref{sec:decoding}). Experiments on abstractive summarisation, machine translation, and morphological inflection show that the noisy channel model can significantly improve performance and exploit unpaired output training samples and that models that combine the direct model and a noisy channel model offer further improvements still~(\S\ref{sec:experiments}).

\section{Background}
In this section, we first provide the background of the noisy channel model, which includes the originality of the concept, the derivation of the model based on the Bayes' rule, the advantages of the model, and its application on statistical machine translation. In order to make this chapter self-contained, we then review the main concepts of SSNT that were presented in Chapter \ref{ch:ssnt}, with emphasis on why it is suitable for a channel model. 
\subsection{The Noisy Channel Model}
\begin{figure}
	\centering
	\includegraphics[scale=0.8]{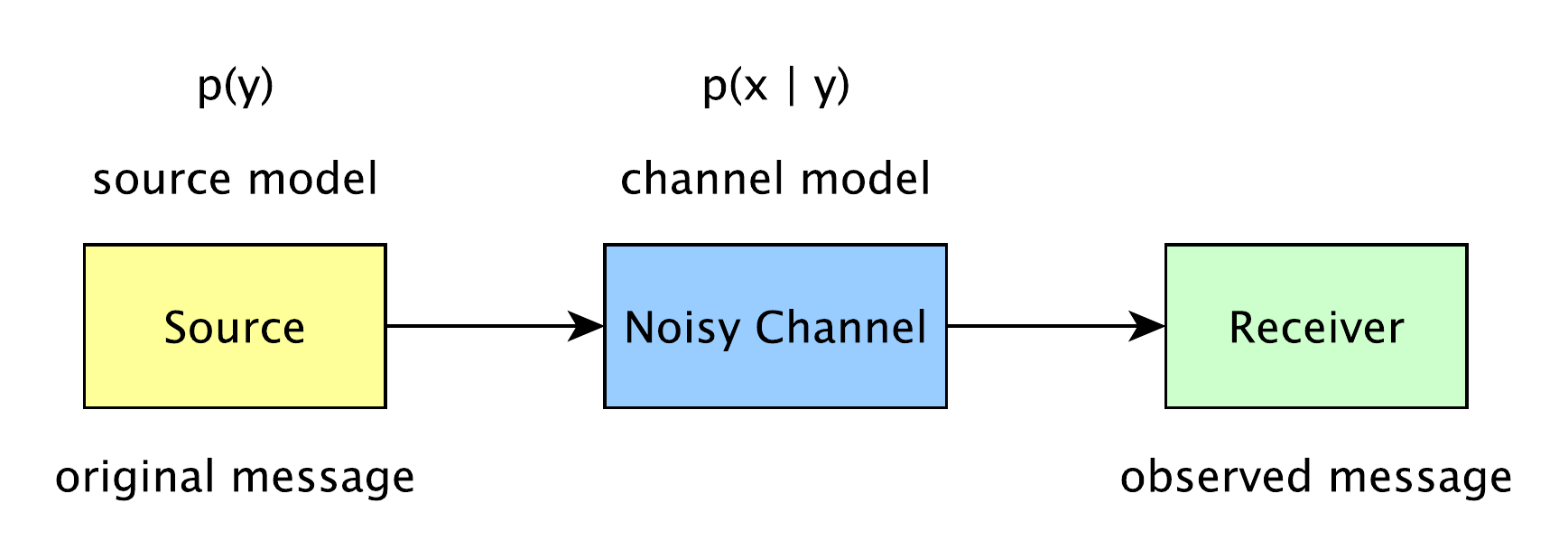}
	\caption[Noisy channel model]{The noisy channel model.}
	\label{fig:noisy_channel}
\end{figure}
The noisy channel model \citep{shannon:1948} has been the foundation of statistical models in speech recognition and natural language processing. It was originally introduced to describe a communication process in which a message emitted by a sender is corrupted by passing through a noisy communication channel resulting in a ``distorted'' version of the original message arriving at the recipient. The challenge is to recover the intended message using the knowledge about the possible source messages and the knowledge about how the message is corrupted in the channel.

The noisy channel model is essentially Bayesian inference. Let $\boldsymbol{y}$ and $\boldsymbol{x}$ denote the original message and the observed distorted message, respectively. Our goal is to identify the $\boldsymbol{y}$ such that $p(\boldsymbol{y}\ |\ \boldsymbol{x})$ is the highest:
\begin{eqnarray}
\hat{\boldsymbol{y}} = \arg\max_{\boldsymbol{y}} p(\boldsymbol{y}\ |\ \boldsymbol{x}).
\end{eqnarray}
Using the Bayes' rule, we can rewrite the above equation as:
\begin{eqnarray}
\label{eq:nc_bayes}
\hat{\boldsymbol{y}} = \arg\max_{\boldsymbol{y}} \frac{p(\boldsymbol{x}\ |\ \boldsymbol{y})p(\boldsymbol{y})}{p(\boldsymbol{x})}.
\end{eqnarray}
Since  $p(\boldsymbol{x})$ is the same for every possible $\boldsymbol{y}$, the denominator $p(\boldsymbol{x})$ in Equation \ref{eq:nc_bayes} can be dropped. Therefore, we have:
\begin{eqnarray}
\hat{\boldsymbol{y}} = \arg\max_{\boldsymbol{y}}\underbrace{p(\boldsymbol{x}\ |\ \boldsymbol{y})}_{\text{channel model}} \cdot \underbrace{p(\boldsymbol{y})}_{\text{source model}}.
\end{eqnarray}

The noisy channel model is thus formulated with two components, the channel model that characterises how the message is distorted and the source model that gives the probability of the original message (prior probability). We compute the most probable message $\boldsymbol{y}$ given the observed message $\boldsymbol{x}$ by multiplying the source model $p(\boldsymbol{y})$ and the channel model $p(\boldsymbol{x}\ |\ \boldsymbol{y})$ and selecting the message for which this product is the highest. 

The seq2seq problems can be framed in terms of the noisy channel model, where the target sequence corresponds to the original message and the source sequence corresponds to the observed message.   
Notably, the incorporation of the source model allows us to train the model with unpaired data (which is usually largely available) in addition to paired data. 

\textbf{Advantages of the Noisy Channel Model}~ Beyond their data omnivorousness, noisy channel models have other benefits. First, the two component models mean that two different aspects of the transduction problem can be addressed independently. For example, in many applications, source models are language models, and innovations in these can be leveraged to obtain improvements in any system that uses them as a component. Second, the component models can have complementary strengths, since inference is carried out in the product space; this simplifies design because a single model does not have to get everything perfectly right. Third, the noisy channel operates by selecting outputs that both are \emph{a priori} likely \emph{and} that explain the input well. This addresses a failure mode that can occur in conditional models in which inputs are ``explained away'' by highly predictive output prefixes, resulting in poor training \citep{klein:2001}. Since the noisy channel formulation requires its outputs to explain the observed input, this problem is avoided.

\textbf{Noisy Channel Model in Statistical Machine Translation}~ During the period when information theory was developed, \cite{weaver1955translation} suggested that the problem of translation could be tackled using the noisy channel model. His idea is that we assume a foreign speaker intends to utter an English sentence, but the sentence is corrupted in a noisy channel which causes a foreign sentence observed. This approach was abandoned at that time due to the lack of computational power and due to some theoretical objections. It was not until the 90s when the noisy channel model was brought back for translation \citep{brown1990statistical}. Since then the statistical methods and in particular the noisy channel model has been dominant in machine translation for decades.

 Consider a source-target pair $(\boldsymbol{x}, \boldsymbol{y})$. In machine translation the channel model that characterises the reverse transduction is called the \emph{translation model}, and the source model is called the \emph{language model}. The language model, trained on the corpora in the target language, is responsible for the fluency of the translation. The $n$-gram language models are used to be the standard language model in machine translation, and more recently recurrent language models (as reviewed in Section \ref{sec:rnnlm}) are more widely used. Compared to the research on language modelling, more work in the statistical machine translation community has been devoted to the development of translation models. The initial translation models are based on the mapping of words. They are usually parameterised by a combination of probabilities \citep[Chapter 4]{koehn2009statistical}: 
 \begin{enumerate}
 	\item \emph{fertility probability}: the probability that a source word is translated to $n$ target words;
 	\item \emph{word translation probability}: the probability that a source word is translated into some target word;
 	\item \emph{distortion probability}: the probability of word reordering.
 \end{enumerate} 
 The more successful translation models are phrase-based models, where phrases are served as atomic units. Phrase-based models typically do not strictly follow the noisy channel framework. Instead, the whole machine translation system is constructed by an interpolation of different components such as a language model, a phrase translation model, and a reordering model. We adopt this log-linear framework in our decoding algorithm, which is presented in Section \ref{sec:model_combination}. Although neural seq2seq models have achieved promising results in machine translation, statistical models are still leading the state of the art in some translation tasks, especially for  low-resource languages. For a detailed review of statistical machine translation, we refer the readers to the survey \citep{lopez2008statistical} and the book \citep{koehn2009statistical}. 

In addition to machine translation~\citep{brown:1993,koehn2007moses,dyer2010cdec}, noisy channel decompositions have also been successfully used in a variety of other problems, including speech recognition~\citep{jelinek:1998}, spelling correction~\citep{brill:2000}, document compression~\citep{daume2002noisy}, word sense disambiguation~\citep{yuret2010noisy}, and question answering~\citep{echihabi:2003}. 

\subsection{Segment to Segment Neural Transduction}
\label{sec:model}

Our model is based on the Segment to Segment Neural Transduction model (SSNT) that we proposed in Chapter \ref{ch:ssnt}. At a high level, the model alternates between encoding more of the input sequence and decoding output tokens from the encoded representation. This presentation highlights the key components of the model that enable incremental construction of the conditioning context. For a detailed description of the model, we refer the readers to the previous chapter.

SSNT models the conditional probability $p(\boldsymbol{y} \mid \boldsymbol{x})$ of an output sequence $\boldsymbol{y}$ given an input sequence $\boldsymbol{x}$. Different from other neural seq2seq models, a latent variable $\boldsymbol{z}$ is introduced to control how many input tokens to be read before predicting the next output token. Since we assume that the input is read just once from left to right, we restrict $\boldsymbol{z}$ to be a monotonically increasing alignment (i.e., $z_{j+1} \ge z_j$ is true with probability 1),
where $z_j = i$ denotes that the output token at position $j$ ($y_j$) is generated when the input sequence up through position $i$ has been read. The SSNT model is:
\begin{align}
\begin{split}
p(\boldsymbol{y} \mid \boldsymbol{x}) & = \sum_{\boldsymbol{z}} p(\boldsymbol{y}, \boldsymbol{z} \mid \boldsymbol{x}) \\
p(\boldsymbol{y}, \boldsymbol{z} \mid \boldsymbol{x}) & \approx   \prod_{j=1}^{|\boldsymbol{y}|} \underbrace{p(z_j \mid z_{j-1},
	\boldsymbol{x}_{1}^{z_j},
	\boldsymbol{y}_1^{j-1})}_{\text{alignment probability}} \underbrace{p(y_j \mid \boldsymbol{x}_{1}^{z_j},
	\boldsymbol{y}_1^{j-1})}_{\text{word probability}}. \label{eq:model}
\end{split}
\end{align}
The joint probability is factorised into two terms: the alignment probability and word probability. 
In SSNT, the input and output sequences $\boldsymbol{x}$, $\boldsymbol{y}$ are encoded with two separate LSTMs \citep{hochreiter1997long}, resulting in sequences of hidden states representing prefixes of these sequences. In our previous formulation, the input sequence encoder (i.e., the conditioning context encoder) can either be a unidirectional or bidirectional LSTM, but here we assume that it is a unidirectional LSTM, which ensures that it will function well as a channel model that can compute probabilities with incomplete conditioning contexts (this is necessary since, at decoding time, we will be constructing the conditioning context incrementally). The aligned hidden state vectors from both LSTMs are used to calculate the probability of the next output token. 

We also carefully design the parameterisation of the alignment transition so that the alignment decision between the $i$-th input token and the $j$-th output token (i.e. $p(z_j = i \mid z_{j-1},
	\boldsymbol{x}_{1}^{z_j},
	\boldsymbol{y}_1^{j-1})$) can be computed by just reading the input sequence up to the $i$-th token, rather than the entire input sequence. This is accomplished
by decomposing it into a sequence of \textsc{shift} and \textsc{emit} operations that progressively decide whether to read another input token or start generating the next output token. The inference of the model is carried out using a dynamic programming algorithm similar to the forward-backward algorithm for the Hidden Markov Models \citep{rabiner1989tutorial}.

\section{Decoding}
\label{sec:decoding}
We now turn to the problem of decoding, that is, of computing
\begin{align*}
\hat{\boldsymbol{y}} = \arg \max_{\boldsymbol{y}} p(\boldsymbol{x} \mid \boldsymbol{y}) p(\boldsymbol{y}),
\end{align*}
where we are using the SSNT model described in the previous section as the channel model and a language model that delivers prior probabilities of the output sequence in left-to-right order, i.e., $p(y_i \mid \boldsymbol{y}^{i-1}_1)$.

Marginalising the latent variable during search is computationally hard \citep{simaan:1996}, and so we approximate the search problem as  \citep{jelinek1976continuous,brown:1993}
\begin{align*}
\hat{\boldsymbol{y}} = \arg \max_{\boldsymbol{y}} \max_{\boldsymbol{z}} p(\boldsymbol{x},\boldsymbol{z} \mid \boldsymbol{y}) p(\boldsymbol{y}).
\end{align*}
However, even with this simplification, the search problem remains nontrivial. On one hand, we must search over the space of all possible outputs with a model that makes no Markovian assumptions. This is similar to the decoding problem faced in standard seq2seq transducers.  On the other hand, our model computes the probability of the given input conditional on the predicted output hypothesis. Therefore, instead of just relying on a single softmax to provide a probability for every output word type (as we conveniently can in the direct model), we must loop over each output word type, and run a softmax over the input vocabulary---a computational expense that is quadratic in the size of the vocabulary!

To reduce this computational effort, we make use of an auxiliary direct model $q(\boldsymbol{y}, \boldsymbol{z} \mid \boldsymbol{x})$ to explore probable extensions of partial hypotheses, rather than trying to perform an exhaustive search over the vocabulary each time we extend an item on the beam.

Algorithm~\ref{nc_decode} describes the decoding algorithm based on a formulation by \citet{tillmann1997dp}. The idea is to create a matrix $Q$ of partial hypotheses. Each hypothesis in cell $(i,j)$ covers the first $i$ words of the input ($\boldsymbol{x}_1^i$) and corresponds to an output hypothesis prefix of length $j$ ($\boldsymbol{y}_1^j$). The hypothesis is associated with a model score. For each cell $(i, j)$, the direct proposal model first calculates the scores of possible extensions of previous cells that could then reach $(i,j)$ by considering every token in the output vocabulary, from all previous candidate cells $(i-1,\le j)$. That gives the top $K_1$ partial output sequences. These partial output sequences are subsequently rescored by the noisy channel model, and the $K_2$ best candidates are kept in the beam and used for further extension. The beam size $K_1$ and $K_2$ are hyperparameters to be tuned in the experiments. 

\begin{algorithm*}[ht]                      
	\caption{Noisy Channel Decoding}        
	\label{nc_decode} 
	\begin{algorithmic}          
		\State \textbf{Notation: } $Q$ is the Viterbi matrix, bp is the backpointer, $W$ stores the predicted tokens, $\mathcal{V}$ refers to the vocabulary, $I=|\boldsymbol{x}|$, and $J_\text{max}$ denotes the maximum number of output tokens that can be predicted.
		\State \textbf{Input: } source sequence $\boldsymbol{x}$
		\State \textbf{Output: } best output sequence $\boldsymbol{y^*}$
		\State \textbf{Initialisation: } $Q \in \mathbb{R}^{I \times J_\text{max}\times K_1}$, bp $\in \mathbb{N}^{I \times J_\text{max}\times K_1}$,  $W \in \mathbb{N}^{I \times J_\text{max}\times K_1}$,
		
		\ \ \ \ \ \ \ \ \ \ \ \ \ \ \ \ \ \
		$Q_{temp} \in \mathbb{R}^{K_1}$, $bp_{temp} \in \mathbb{N}^{K_1}$, $W_{temp} \in \mathbb{N}^{K_1}$
		\For{$i \in [1, I]$}
		\State $Q_{temp} \gets \topk(K_1)_{y \in \mathcal{V}}q(z_1 = i) $ $q(y\ |\ \textsc{start}, z_1, \boldsymbol{x}_{1}^{z_1})$ \Comment Candidates generated by $q(\boldsymbol{y}\ |\ \boldsymbol{x})$.
		\State $bp_{temp}\gets 0$
		\State $W_{temp} \gets \argtopk(K_1)_{y \in \mathcal{V}}q(z_1 = i)$ $q(y\ |\ \textsc{start}, z_1, \boldsymbol{x}_1^{z_1})$
		
		\State $Q[i, 1] \gets \topk(K_2)_{y \in W_{temp}} O_{\boldsymbol{x}_1^i, y}$ \Comment Rerank the candidates by objective ($O$).
		\State $W[i,1] \gets \argtopk(K_2)_{y \in W_{temp}}O_{\boldsymbol{x}_1^i, y}$
		\EndFor
		\For{$j\in[2, J_\text{max}]$}
		\For{$i \in [1, I]$}
		\State $Q_{temp} \gets \topk(K_1)_{y \in \mathcal{V}, k \in [1, i]} Q[k,j-1] \cdot$ $q(z_j = i\ |\ z_{j-1} = k)q(y\ |\ \boldsymbol{y}_1^{j-1}, z_j, \boldsymbol{x}_1^{z_j})$
		\State $bp_{temp} , W_{temp} \gets \argtopk(K_1)_{y \in \mathcal{V}, k \in [1, i]} $ $Q[k,j-1]q(z_j = i\ |\ z_{j-1} = k) \cdot$
		
		\ \ \ \ \ \ \ \ \ \ \ \ \ \ \ \ \ \ \ \ \ \ \ \ \ \ \ \ \ \ \ \ \ \ \ \ \ \ \ \ \ \ \ \ \ \ \ \ \ \ \ \ \ \ \ \ \ \ \ \ \ \ \ \ \ \ \ \ \ \ \ \ \ \ \ \ \ \ \ \ \ \ \ \ \ \ \ \ \ \ \ \ \ \ \ \ \ \ \ \ \ \ \ \ \ \ \ \ \ \ \ \ \ \ $q(y\ |\ \boldsymbol{y}_1^{j-1}, z_j, \boldsymbol{x})$
		
		\State $Y \gets \candidate(bp_{temp}, W_{temp})$ \Comment Get partial candidate $\boldsymbol{y}_1^j$.
		\State $Q[i,j] \gets \topk(K_2)_{\boldsymbol{y}_j \in Y} O_{\boldsymbol{x}_1^i, \boldsymbol{y}_1^j}$ 
		\State $bp[i,j] , W[i,j] \gets 
		\argtopk(K_2)_{\boldsymbol{y}_1^j \in Y}$ $O_{\boldsymbol{x}_1^i, \boldsymbol{y}_1^j}$
		\EndFor
		\EndFor 
		\State
		\Return a sequence of words stored in $W$ by following backpointers starting from $(I,\argmax_j Q[I, j])$.
	\end{algorithmic}
\end{algorithm*}

\subsection{Model combination}
\label{sec:model_combination}
The decoder we have just described makes use of an auxiliary decoding model.
This means that, as a generalisation, it is capable of decoding under an objective that is a linear combination of the direct model, channel model, language model and a bias for the output length,\footnote{In the experiments, we did not marginalise the probability of the direct model when calculating the general search objective. We found that marginalising the probability does not give better performance and makes decoding extremely slow.}
\begin{equation}
O_{\boldsymbol{x}_1^i, \boldsymbol{y}_1^j} = \lambda_1 \log p(\boldsymbol{y}_1^j\ |\ \boldsymbol{x}_1^i) + \lambda_2 \log p(\boldsymbol{x}_1^i\ |\ \boldsymbol{y}_1^j) + \lambda_3 \log p(\boldsymbol{y}_1^j) + \lambda_4 |\boldsymbol{y}_1^j|.
\end{equation}
The bias is used to penalise the noisy channel model for generating too-short (or long) sequences. The $\lambda$'s are hyperparameters to be tuned on a small amount of held-out development data. 

\section{Experiments}
\label{sec:experiments}
We evaluate our model on three natural language processing tasks, abstractive sentence summarisation, machine translation, and morphological inflection generation. For each task, we compare the performance of the direct model, noisy channel model, and the interpolation of the two models.

\subsection{Abstractive Sentence Summarisation}
Sentence summarisation is the problem of constructing a shortened version of a sentence while preserving the majority of its meaning. In contrast to extractive summarisation, which can only copy words from the original sentence, abstractive summarisation permits arbitrary rewording of the sentence. The dataset \citep{DBLP:conf/emnlp/RushCW15} that we use is constructed by pairing the first sentence and the headline of each article from the annotated Gigaword corpus \citep{graff2003english,napoles2012annotated}. There are 3.8m, 190k and 381k sentence pairs in the training, validation and test sets, respectively. In the previous chapter, we have mentioned that we filtered the dataset by restricting the lengths of the input and output sentences to be no greater than 50 and 25 tokens, respectively. From the filtered data, we further sampled 1 million sentence pairs for training. We experiment on training the direct model and channel model with both the sampled 1 million and the full 3.8 million parallel data. The language model is trained on the target side of the parallel data, i.e. the headlines. We evaluate the generated summaries of 2000 randomly sampled sentence pairs using full length ROUGE F1. This setup is in line with the previous work on this task \citep{DBLP:conf/emnlp/RushCW15,chopra,gulcehre2016pointing}.

The same configuration is used to train the direct model and the channel model. The losses of the two models are optimised by Adam \citep{DBLP:journals/corr/KingmaB14}, with the initial learning rate of 0.001. We use LSTMs with 1 layer for both the encoder and decoders, with hidden units of 256. The mini-batch size is 32, and dropout of 0.2 is applied to the input and output of LSTMs. For the language model, we use a 2-layer LSTM with 1024 hidden units and 0.5 dropout. The learning rate is 0.0001. All the hyperparameters are optimised via grid search on the perplexity of the validation set. During decoding, beam search is employed with the number of proposals generated by the direct model $K_1 = 20$, and the number of best candidates selected by the noisy channel model $K_2 = 10$.

\begin{table}[t]\centering
	\begin{tabular}{@{}lccccc@{}}
		\toprule
		Model & Paired data & Unpaired data & RG-1 & RG-2  & RG-L \\
		\midrule
		direct (uni)$^*$ & 1.0m & - & 30.94 & 14.20 & 28.72 \\
		direct (bi) & 1.0m & - & 31.25 & 14.52 & 29.03 \\
		direct (bi) & 3.8m & - & 33.82 & 16.66 & 31.50 \\
		\midrule
		channel + LM + bias (uni)$^*$	& 1.0m & 1.0m & 31.92 & 14.75 & 29.58 \\
		channel + LM + bias (bi)	& 1.0m & 1.0m & 31.96 & 14.89 & 29.51 \\
		direct + channel + LM + bias (uni) & 1.0m & 1.0m & 33.07 & 15.21 & 30.29 \\
		direct + channel + LM + bias (bi) & 1.0m & 1.0m & 33.18 & 15.65 & 30.45 \\
		channel + LM + bias (uni)$^*$ & 1.0m & 3.8m & 32.59 & 15.05 & 30.06 \\
		channel + LM + bias (bi) & 1.0m & 3.8m & 32.65 & 14.95 & 30.23 \\
		direct + LM + bias (bi) & 1.0m & 3.8m & 31.25 & 14.52 & 29.03 \\
		direct + channel + LM + bias (uni) & 1.0m & 3.8m  & 33.16 & 15.63 & 30.53 \\
		direct + channel + LM + bias (bi) & 1.0m & 3.8m  & 33.21  & 15.65 & 30.60 \\
		chanel + LM + bias (bi) & 3.8m & 3.8m & 34.12 & 16.41 & 31.38\\
		direct + LM + bias (bi) & 3.8m & 3.8m & 33.82 & 16.66 & 31.50 \\
		direct + channel + LM + bias (bi) & 3.8m & 3.8m & \bfseries{34.41} & \bfseries{16.86} & \bfseries{31.83} \\
		\bottomrule
	\end{tabular}
	\caption[ROUGE F1 scores on the sentence summarisation test set] {ROUGE F1 scores on the sentence summarisation test set. The `uni' and `bi' in the parentheses denote the encoder for the model proposing candidates is a unidirectional LSTM or bidirectional LSTM. Those rows marked with an $*$ denote models that process their input online.}
	\label{sent_comp_result}
\end{table}

\begin{table}[t]\centering
	\begin{tabular}{@{}lccccc@{}}
		\toprule
		Model & Paired data & Unpaired data & RG-1 & RG-2  & RG-L \\
		\midrule
		ABS+ & 3.8m & - & 29.55 & 11.32 & 26.42 \\
		RAS-LSTM & 3.8m & - & 32.55 & 14.70 & 30.03 \\
		RAS-Elman & 3.8m & - & 33.78 & 15.97 & 31.15 \\
		Pointing unkown words & 3.8m & - & \bfseries{35.19} & 16.66 & \bfseries{32.51} \\
		ASC + FSC & 1.0m & 3.8m & 31.09 & 12.79 & 28.97 \\
		ASC + FSC & 3.8m & 3.8m & 34.17 & 15.94 & 31.92 \\
		\midrule
		direct + channel + LM + bias (bi) & 1.0m & 3.8m & 33.21  & 15.65 & 30.60 \\
		direct + channel + LM + bias (bi) & 3.8m & 3.8m & 34.41 & \bfseries{16.86} & 31.83 \\
		\bottomrule 
	\end{tabular}
	\caption[Overview of results on the abstractive sentence summarisation task] {Overview of results on the abstractive sentence summarisation task. ABS+ \citep{DBLP:conf/emnlp/RushCW15} is the attentive model with bag-of-words as the encoder. RAS-LSTM and RAS-Elman \citep{chopra} are the seq2seq models with attention with the RNN cell implemented as LSTMs and an Elman architecture \citep{elman1990finding}, respectively. Pointing the unknown words \citep{gulcehre2016pointing} uses pointer networks \citep{vinyals2015pointer} to select the output token from the input sequence in order to avoid generating unknown tokens. ASC + FSC \citep{miao2016} is the semi-supervised model based on a variational autoencoder. } 
	\label{prev_work}
\end{table}

Table \ref{sent_comp_result} presents the ROUGE-F1 scores of the test set from the direct model, noisy channel model (channel + LM + bias), the interpolation of the direct model and the noisy channel model (direct + channel + LM + bias), and the interpolation of the direct model and language model (direct + LM + bias) trained on different sizes of data. The noisy channel model with the language model trained on the target side of the 1 million parallel data outperforms the direct model by approximately 1 point. Such improvement indicates that the language model helps improve the quality of the output sequence when no extra unlabelled data is available. Training the language model with all the headlines in the dataset, i.e. 3.8 million sentences, gives a further boost to the ROUGE score. This is in line with our expectation that the model benefits from adding large amounts of unlabelled data. The interpolation of the direct model, channel model, language model and bias of the output length achieves the best results --- the ROUGE score is close to the direct model trained on all the parallel data. Although there is still an improvement, when the direct model is trained with more data, the gap between the direct model and the noisy channel model is smaller. No gain is observed if the language model is combined with the direct model. We find that as we increase the weight of the language model, the result is getting worse.

Table \ref{prev_work} surveys published results on this task, and places our best models in the context of the current state-of-the-art results. ABS+ \citep{DBLP:conf/emnlp/RushCW15}, RAS-LSTM and RAS-Elman \citep{chopra} are different variations of the attentive models. {\it Pointing the unknown words} uses pointer networks \citep{vinyals2015pointer} to select the output token from the input sequence in order to avoid generating unknown tokens. ASC + FSC \citep{miao2016} is a semi-supervised model based on a variational autoencoder. Trained on 1m paired samples and 3.8m unpaired samples, the noisy channel achieves comparable or better results than (direct) models trained with 3.8m paired samples. Compared to \cite{miao2016}, whose ASC + FSC models is an alternative strategy for using unpaired data, the noisy channel is significantly more effective --- 33.21 versus 31.09 in ROUGE-1.

Finally, motivated by the qualitative observation that noisy channel model outputs were quite fluent and often used reformulations of the input rather than a strict compression (which would be poorly scored by  ROUGE), we carried out a human preference evaluation whose results are summarised in Table~\ref{tab:human}. This confirms that noisy channel summaries are strongly preferred over those of the direct model.

\begin{table}[h]\centering
	\begin{tabular}{lc}
		\toprule
		Model & count \\
		\midrule
		both bad &188 \\ 
		both good &106 \\
		direct $>$ noisy channel &135 \\
		noisy channel $>$ direct & \bfseries{212} \\
		\bottomrule 
	\end{tabular}
	\caption[Results of human evaluation]{Preference ratings for 641 segments from the test set (each segment had ratings from at least 2 raters with $\ge$ 50\% agreement on the label and where one label had a plurality of the votes).} 
	\label{tab:human}
\end{table}

\subsection{Machine Translation}
We next evaluate our models on a Chinese--English machine translation task. We use the same parallel data as the experiment in \ref{sec:zh-en-mt}---184k sentence pairs from the FBIS corpus and LDC2003E14---and monolingual data with 4.3 million of English sentences (selected from the English Gigaword). The training data is preprocessed by lowercasing the English sentences, replacing digits with `\#' token, and replacing tokens appearing less than 5 times with an `UNK' token.

The models are trained using Adam \citep{DBLP:journals/corr/KingmaB14} with the initial learning rate of 0.001 for the direct model and the channel model, and 0.0001 for the language model. The LSTMs for the direct and channel models have 512 hidden units and 1 layer, and 2 layers with 1024 hidden units per layer for the language model. Dropout of 0.5 on the input and output of LSTMs is set for all the model training. The noisy channel decoding uses $K_1$ = 20 and $K_2$ = 10 as the beam sizes.

Table \ref{mt-result} lists the translation performance of different models in BLEU scores. We include the results from the vanilla and attentional seq2seq models \citep{sutskever2014sequence,bahdanau2014neural} as benchmarks. For direct models, we leverage bidirectional LSTMs as the encoder for this task. 
We can see that the noisy channel model is approximately 3 points higher in BLEU than the direct model, and the combination of noisy channel and direct model gives an extra boost. Confirming the empirical findings of prior work \citep{gulcehre:2015} (and in line with theoretical predictions), the interpolation of the direct model and language model is not effective.

\begin{table}[t]\centering
	\begin{tabular}{@{}lc@{}}
		\toprule
		Model & BLEU \\
		\midrule
		seq2seq w/o attention & 11.19 \\
		seq2seq w/ attention & 25.27\\
		direct (bi) & 23.33 \\
		\midrule
		direct + LM + bias (bi) & 23.33 \\
		channel + LM + bias (bi) & 26.28 \\
		direct + channel + LM + bias (bi) & \bfseries{26.44} \\
		\bottomrule
	\end{tabular}
	\caption[Results of Chinese to English translation]{BLEU scores from different models for the Chinese to English machine translation task.}
	\label{mt-result}
\end{table}

\subsection{Morphological Inflection Generation}
The dataset \citep{durrett2013supervised} that we use in the experiments is created from Wiktionary, including inflections for German nouns, German verbs, Spanish Verbs, Finnish noun and adjective, and Finnish verbs. We only experimented on German nouns and German verbs, as German nouns is the most difficult task,\footnote{While state-of-the-art systems can achieve 99\% accuracies on Spanish verbs and Finnish verbs, they can only get 89\% accuracy on German nouns.} and the direct model does not perform as well as other state-of-the-art systems on German verbs. The train/dev/test split for German nouns is 2364/200/200, and for German verbs is 1617/200/200. There are 8 and 27 inflection types in German nouns and German verbs, respectively. Following previous work, we learn a separate model for each type of inflection independent
of the other inflections. We report results on the average accuracy across different inflections.
Our language models were trained on word types extracted by running a morphological analysis tool on the WMT 2016 monolingual data and extracting examples of appropriately inflected word forms.\footnote{\url{http://www.statmt.org/wmt16/translation-task.html}}
After annotation, the number of instances for training the language model ranged from 300k to 3.8m for different inflection types in German nouns, and from 200 to 54k in German verbs.

The experimental setup that we use on this task is $K_1$ = 60, $K_2$ = 30,
\begin{itemize}
	\item direct and channel model: 1 layer LSTM with 128 hidden, $\eta = 0.001$, dropout = 0.5.
	\item language model: 2 layer LSTM with 512 hidden, $\eta = 0.0001$, dropout = 0.5.
\end{itemize}

Table \ref{morph-result} summarises the results from our models. On both datasets, the noisy channel model (channel + LM + bias) does not perform as well as the direct model, but the interpolation of the direct model and noisy channel model (direct + channel + LM + bias) significantly outperforms the direct model. The interpolation of the direct model and language model (direct + LM + bias) achieves better results than the direct model and the noisy channel model on German nouns, but not on German verbs. For further comparison, we also included the state-of-the-art results as benchmarks. NCK15 \citep{nicolai2015inflection} tackles the task based on the three-stage approach: (1) align the source and target word, (2) extract inflection rules, (3) apply the rule to new examples. FTND16 \citep{faruqui2015morphological} is based on neural seq2seq models. Both models (NCK15+ and FTND16+) rerank the candidate outputs by the scores predicted from n-gram language models, together with other features. 

\begin{figure}
	\begin{subfigure}{.5\textwidth}
		\centering
		\begin{tabular}{@{}lc@{}}
			\toprule
			Model & Acc. \\
			\midrule
			NCK15 & 88.60\\
			FTND16 & 88.12 \\
			NCK15+ & 89.90\\
			FTND16+ & 89.31\\
			\midrule
			direct (uni) & 82.25 \\
			direct (bi) & 87.68 \\
			\midrule
			channel + LM + bias (uni) & 78.38\\
			channel + LM + bias (bi) & 78.13 \\
			direct + LM + bias (bi) & 90.31 \\
			direct + channel + LM + bias (uni) & 88.44 \\
			direct + channel + LM + bias (bi) & \bfseries{90.94} \\ 
			\bottomrule
		\end{tabular}
		\caption{}
		\label{fig:sfig1}
	\end{subfigure}%
	\begin{subfigure}{.5\textwidth}
		\centering
		\begin{tabular}{@{}lc@{}}
			\toprule
			Model & Acc. \\
			\midrule
			NCK15 & 97.50 \\
			FTND16 & \bfseries{97.92} \\
			NCK15+ & 97.90 \\
			FTND16+ & 97.11 \\
			\midrule
			direct (uni) & 87.85\\
			direct (bi) & 94.83\\
			\midrule
			channel + LM + bias (uni) & 84.42\\
			channel + LM + bias (bi) & 92.13\\
			direct + LM + bias (bi) & 94.83 \\
			direct + channel + LM + bias (uni) & 92.20\\
			direct + channel + LM + bias (bi) & 97.15\\ 
			\bottomrule
		\end{tabular}
		\caption{}
		\label{fig:sfig2}
	\end{subfigure}
	\caption[Accuracy on morphological inflection]{Accuracy on morphological inflection of German nouns (a), and German verbs (b). NCK15 \citep{nicolai2015inflection} and FTND16 \citep{faruqui2015morphological} are previous state-of-the-art on this task, with NCK15 based on feature engineering, and FTND16 based on neural networks. NCK15+ and FTND16+ are the semi-supervised setups of these models.}
	\label{morph-result}
\end{figure}

\section{Analysis}
By observing the output generated by the direct model and noisy channel model, we find (in line with theoretical critiques of conditional models) that the direct model may leave out key information. By contrast, the noisy channel model does seem to avoid this issue. To illustrate, in Example~1 in Table~\ref{tb:nc_sum_example}, the direct model ignores the key phrase `coping with', resulting in incomplete meaning, but the noisy channel model covers it. Similarly, in Example 6 in Table~\ref{tb:nc_mt_example}, the direct model does not translate the Chinese word corresponding to `investigation'. We also observe that while the direct model mostly copies words from the source sentence, the noisy channel model prefers generating paraphrases. For instance, in Example 2, while the direct model copies the word `accelerate' in the generated output, the noisy channel model generates `speed up' instead. While one might argue that copying is a preferable compression technique than paraphrasing (as long as it produces grammatical outputs), it does show the power of these models. 

\begin{table}
	\begin{tabular}{@{}lp{12cm}@{}}
		\toprule
		\bfseries{Example 1:} & \\
		\bfseries{source:} & the european commission on health and consumers protection $-$lrb$-$ \_unk\_ $-$rrb$-$ has offered cooperation to indonesia in coping with the spread of avian influenza in the country , official news agency antara said wednesday . \\
		\bfseries{reference:} & eu offers indonesia cooperation in \underline{avian flu eradication} \\
		\bfseries{direct:} & eu offers cooperation to indonesia in \underline{avian flu} \\
		\bfseries{nc:} & eu offers cooperation to indonesia in \underline{coping with bird flu} \\
		\midrule
		\bfseries{Example 2:} & \\
		\bfseries{source:} & vietnam will \underline{accelerate} the export of industrial goods mainly by developing auxiliary industries , and helping enterprises sharpen competitive edges , according to the ministry of industry on thursday . \\
		\bfseries{reference:} & vietnam to \underline{boost} industrial goods export \\
		\bfseries{direct:} & vietnam to \underline{accelerate} export of industrial goods \\
		\bfseries{nc:} & vietnam to \underline{speed up} export of industrial goods \\
		\midrule
		\bfseries{Example 3:} & \\
		\bfseries{source:} & japan 's toyota team europe were banned from the world rally championship for one year here on friday in a crushing ruling by the world council of the international automobile federation -lrb- fia - \\
		\bfseries{reference:} &  toyota are banned for a year \\
		\bfseries{direct:} & toyota banned from world rally championship \\
		\bfseries{nc:} & toyota europe banned from world rally championship for one year \\
		\midrule
		\bfseries{Example 4:} & \\
		\bfseries{source:} & oil prices roared higher towards \#\# dollars on monday as equity markets surged on government action aimed at tackling a severe economic downturn . \\
		\bfseries{reference:} & oil prices soar towards \#\# dollars\\
		\bfseries{direct:} & oil prices jump towards \#\# dollars \\
		\bfseries{nc:} & oil prices climb towards \#\# dollars \\
		\bottomrule
	\end{tabular}
	\caption[Example outputs from models on sentence summarisation] {Example outputs on the test set from the direct model and noisy channel model for the summarisation task.}
	\label{tb:nc_sum_example}
\end{table}

\begin{table}
	\begin{tabular}{@{}lp{12cm}@{}}
		\toprule
		\bfseries{Example 5:} & \\
		\bfseries{source:} & \begin{CJK*}{UTF8}{gbsn}
			欧盟\ 和\ 美国\ 都\ 表示\ 可以\ 接受\ 这\ 一\ 妥协\ 方案\ 。
		\end{CJK*}  \\
		\bfseries{reference:} & both the eu and the us indicated that they can accept this plan for a compromise . \\
		\bfseries{direct:} & the eu and the united states indicated that it can accept this compromise .\\
		\bfseries{nc:} & the european union and the united states have said that they can accept such a compromise plan .\\
		\midrule
		\bfseries{Example 6:} & \\
		\bfseries{source:} & \begin{CJK*}{UTF8}{gbsn}
			那么\ 这些\ 这个\ 方面\ 呢\ 是\ 现在\ 警方\ 调查\ 重点\ 。
		\end{CJK*} \\
		\bfseries{reference:} & well , this is the current focus of \underline{police investigation} .  \\
		\bfseries{direct:} & these are present at the current \underline{police} . \\
		\bfseries{nc:} & then these are the key to the current \underline{police investigation} .  \\
		\midrule
		\bfseries{Example 7:} & \\
		\bfseries{source:} & \begin{CJK*}{UTF8}{gbsn}
			双方\ 有可能\ 就此\ 问题\ 在\ 下周\ 进行\ 磋商\ 。
		\end{CJK*} \\
		\bfseries{reference:} & the two sides may conduct negotiations on this issue next week .  \\
		\bfseries{direct:} & the two sides may hold consultations on next week . \\
		\bfseries{nc:} & the two sides are likely to hold consultations on this issue next week . \\
		\midrule
		\bfseries{Example 8:} & \\
		\bfseries{source:} & \begin{CJK*}{UTF8}{gbsn}
			那么\ 在\ 这个\ 问题\ 上\ ,\ 伊朗\ 现在\ 态度\ 比较\ 强硬\ ,\ 而\ 美国\ 的\ 态度\ 更为\ 强硬\ 。
		\end{CJK*} \\
		\bfseries{reference:} & well , iran 's attitude is now quite firm on this issue , while the us takes an even firmer attitude .  \\
		\bfseries{direct:} & on this issue , iran 's attitude is quite hard and the attitude of the united states is still tougher . \\
		\bfseries{nc:} & then , on this issue , iran has now taken a tougher attitude toward it . however , the attitude of the united states is even harder . \\
		\bottomrule
	\end{tabular}
	\caption[Example outputs from models on translation] {Example outputs on the test set from the direct model and noisy channel model for the machine translation.}
	\label{tb:nc_mt_example}
\end{table}

\section{Related work}
 The idea of adding language models and monolingual data in machine translation has been explored in earlier work. \cite{gulcehre:2015} propose two strategies of combining a language model with a neural seq2seq model. In shallow fusion, during decoding the seq2seq model (direct model) proposes candidate outputs and these candidates are reranked based on the scores calculated by a weighted sum of the probability of the translation model and that of the language model. In deep fusion, the language model is integrated into the decoder of the seq2seq model by concatenating their hidden state at each time step. \cite{sennrich:2016} incorporate target language unpaired training data by doing back-translation to create synthetic parallel training data. While this technique is quite effective, its practicality seems limited to problems where the inputs and outputs contain roughly the same information (such as translation). \cite{cheng:2016} leverages the abundant monolingual data by doing multitask learning with an autoencoding objective.

A number of papers have remarked on the tendency for content to get dropped (or repeated) in translation. \citet{liu:2016} propose translating in both a left-to-right and a left-to-right direction and seeking a consensus. \citet{tu:2016} propose augmenting a direct model's decoding objective with a reverse translation model (similar to our channel model except it conditions on the direct model's output RNN's hidden states rather than the words); however, that work just reranks complete translation hypotheses rather than developing a model that permits an incremental search.

\section{Summary}

We have presented and empirically validated a noisy channel transduction model that uses component models based on recurrent neural networks. This formulation lets us use unpaired outputs to estimate the parameters of the source model and input-output pairs to train the channel model. Despite the channel model's ability to condition on long sequences, we are able to maintain tractable decoding by using a latent segmentation variable that breaks the conditioning context up into a series of monotonically growing segments. Our experiments show that this model makes excellent use of unpaired training data.

A limitation of our noisy channel model is that it is computationally expensive, which makes it difficult to apply to datasets with long sequences.  Recall that in the decoding algorithm we leverage an auxiliary direct model to explore possible extensions of partial hypothesis and rerank these proposals using the noisy channel model.  The bottleneck comes from the extra computation of partial outputs reranking. Let $K, I, J$ denote the beam size, the lengths of the input and output sequences, respectively. At each cell $(i, j)$ of the matrix of partial hypothesis, the model has to calculate the probability  $p(\boldsymbol{x}_i \ |\ \boldsymbol{y}_j)p(\boldsymbol{y}_j)$ for every proposal $\boldsymbol{y}_j$.  To decode a sentence, the probability $p(\boldsymbol{x}_i \ |\ \boldsymbol{y}_j)p(\boldsymbol{y}_j)$  is required to be calculated $K \times I \times J$ times. This means that an algorithm with cubic complexity\footnote{As described in Chapter \ref{ch:ssnt}, the complexity of the forward algorithm for computing $p(\boldsymbol{x}_i \ |\ \boldsymbol{y}_j)$ is cubic.} has to run $K \times I \times J$ times. In practice, we can save computation time by caching $p(\boldsymbol{x}_{i-1} \ |\ \boldsymbol{y}_{j-1})$ that has been calculated, but the decoding process is still much slower than the direct models. We will address this limitation in future work.

%% file: conclusion.tex
\chapter{Conclusions}
\label{ch:conclusion}

The aim of this thesis is to advance the state-of-the-art in seq2seq mapping problems with neural networks. Throughout the thesis, we have demonstrated how this goal has been accomplished from three aspects, namely by investigating the role of distributed sentence models, modelling the segmentations between sequences, and by incorporating unpaired data. 

Following an overview of the basic concepts and algorithms of neural networks in Chapter \ref{ch:nn}, we attempt at answer sentence selection in Chapter \ref{ch:sentence_model} via means of distributed representations, and learn to match questions with answers by considering their semantic encoding. We showed that our simple models outperformed those more complex models reliant on feature engineering and external linguistic resources. By keeping the QA matching model consistent, we also experimented on various types of neural networks on modelling sentences. The results are in line with our expectations that distributed sentence models that are capable of capturing long range dependencies achieve superior performance over simpler models on the seq2seq mapping problem.

In Chapter \ref{ch:ssnt}, we introduced the segment to segment neural transduction model (SSNT) which addresses the drawbacks of vanilla encoder-decoders that memorise the entire input sequence in a fixed-size vector. 
We presented an efficient learning algorithm based on dynamic programming and a beam search decoder that produces not only output tokens but also alignments. SSNT is capable of capturing unbounded dependencies in both the input and output sequences. Compared to the attentive seq2seq models, our model has the strength of generating output online. This inherent online nature also permits it the flexibility to use its capacity to choose how much input to encode before decoding each segment. We found that SSNT outperformed the vanilla encoder-decoder by a significant margin and was comparable to the attentional seq2seq model on experiments in machine translation, sentence summarisation, and morphological inflection. 

Despite these aforementioned strengths, we also analysed the limitations of SSNT. When designing the parameterisation of the alignment distribution,  we impose a monotone restriction on the alignment. Experiments show that the model biases towards sequence pairs with largely monotonic alignments due to this restriction. On one hand, this bias makes the model work better than other seq2seq models on the tasks of French-English translation and sentence summarisation. On the other hand, its performance deteriorates in other tasks where substantial reorderings exist in sequence pairs. 
Theoretically, in the extreme case where the last token of the input sequence corresponds to the first token of the output sequence, the model should learn to behave like a vanilla seq2seq model that first encodes the entire input sequence and then generates the entire output sequence. However, in practice, the model tends to alternate between small segments.  Another limitation is that SSNT involves the calculation of a quadratic number of softmaxes, which not only takes more time to train but also requires more memory usage than seq2seq models.

Lastly, in Chapter \ref{ch:noisy_channel}, we discussed how SSNT can be integrated into a noisy channel formulation. We showed that the neural noisy channel model has advantages of exploiting both paired and unpaired data and avoiding the explaining-way effects during training, which the direct models may suffer from. We successfully applied the  model to several sequence generation tasks and obtained promising results. We also pointed out that while the noisy channel model has many advantages, there are still severe computational limitations that would need to be overcome for this to be used on a large scale. 

In the future, apart from addressing the limitations of our proposed models, we would also like to apply these models to other tasks of seq2seq mapping. For example, we would like to apply the neural noisy channel model to automatic speech recognition (ASR). In this case, SSNT will serve as an acoustic model which directly generates acoustic signals conditional on sequences of characters.
Seeing that noisy channel models are still the state of the art in ASR, by using a more sophisticated acoustic model based on RNNs, we expect this model will have the strengths of current noisy channel models (independent acoustic and language modelling components), but the flexibility of direct models for open vocabulary modelling.

Another direction we would like to pursue is to tackle the seq2seq mapping problem at the document level, for example to work on the task of document summarisation. This requires the neural model to process sequences of ten times longer than those used in our current experiments, and to capture the relationship between different sentences within the documents. How to build models satisfying these conditions will be an exciting research topic.

%% file: appendix_a.tex
\chapter{Gradient Calculation for Segment to Segment Neural Transduction}
\label{appendix_a}

In Chapter \ref{ch:ssnt}, we have described the segment to segment neural transduction model (SSNT). In this model, a latent variable is introduced to control the model to alternate between encoding more of the input sequence and decoding output tokens from the encoded representation. We have also provided a forward-backward algorithm that enables the loss and gradients to be calculated in polynomial time. Here we present detailed derivations of gradient calculation using $\alpha$ and $\beta$ that are obtained from the forward-backward algorithm. We present two ways to calculate gradients, indirect derivation (\ref{sec:in_drt}) and direct derivation (\ref{sec:dr_drt}).

\section{Loss Function}
As provided in Section \ref{sec:ssnt_loss}, the negative log likelihood of the training set $S$ is,
\begin{equation}
\begin{split}
\mathcal{L}(\boldsymbol{\theta}) &= - \sum_{(\boldsymbol{x}, \boldsymbol{y}) \in S} \log p(\boldsymbol{y}\ |\ \boldsymbol{x}; \boldsymbol{\theta}),\\ 
\end{split}
\end{equation}
where $\boldsymbol{x}$ and $\boldsymbol{y}$ denote the input and output sequences, respectively. The notation $\boldsymbol{\theta}$ represents all the parameters of the model.

We can rewrite the conditional probability in a number of ways:
\begin{align}
p(\boldsymbol{y}\ |\ \boldsymbol{x}) & = \sum_{\boldsymbol{z}} p (\boldsymbol{z, \boldsymbol{y}}\ |\ \boldsymbol{x}) \\
& = \sum_i p (z_j = i, \boldsymbol{y}\ |\ \boldsymbol{x}) \\
& = \sum_i \alpha(i, j) \beta(i, j) \\
& = \alpha(I, J) \label{eq:prob1}\\
& = \sum_i \alpha(i, J) \label{eq:prob2}\\
& = \sum_i \beta(i, 0)
\end{align}
Note that the steps \ref{eq:prob1} and \ref{eq:prob2} are equivalent because for $j = J, i \in [1, I-1]$, $\alpha(i, J) = 0$.

\section{Loss Gradient: Indirect Derivation}
\label{sec:in_drt}
For brevity, $p(z_j = i\ |\ z_{j-1} = k, \boldsymbol{x}^{z_j}_1, \boldsymbol{y}_1^{j-1} ; \boldsymbol{\theta})$ is abbreviated as $p(z_j = i\ |\ z_{j-1} = k)$.

\cite{berg2010painless} have proved that the gradient of $p(\boldsymbol{y}\ |\ \boldsymbol{x})$ has the same form as its corresponding lower bound in the EM algorithm. We therefore first derive the lower bound of $\log p(\boldsymbol{y}\ |\ \boldsymbol{x})$ and subsequently calculate the gradient of the lower bound.

According to the EM algorithm, we arrive at a lower bound for $\log p(\boldsymbol{y}\ |\ \boldsymbol{x}; \boldsymbol{\boldsymbol{\theta}} )$ by introducing an arbitrary distribution $q(\boldsymbol{z})$:
\begin{equation}
\mathcal{L}'(q,\boldsymbol{\theta})= \sum_{\boldsymbol{z}} q(\boldsymbol{z}) \log \left( \frac{p(\boldsymbol{z},\boldsymbol{y}\ |\ \boldsymbol{x};\boldsymbol{\theta})}{q(\boldsymbol{z})}\right).
\end{equation}
$q(\boldsymbol{z})$ reaches its maximum when $q(\boldsymbol{z}) = p(\boldsymbol{z}\ |\ \boldsymbol{y},\boldsymbol{x};\boldsymbol{\theta}^s)$, where $\boldsymbol{\theta}^s$ is the current model parameters. Therefore, the function we are maximising is 
\begin{equation}
\mathcal{Q}(\boldsymbol{\theta}, \boldsymbol{\theta}^s) = \sum_{\boldsymbol{z}} p(\boldsymbol{z}\ |\ \boldsymbol{y},\boldsymbol{x};\boldsymbol{\theta}^s) \log  p(\boldsymbol{z},\boldsymbol{y}\ |\ \boldsymbol{x};\boldsymbol{\theta}),
\end{equation}
and the corresponding gradients
\begin{eqnarray}
\frac{\partial \log p (\boldsymbol{y}\ |\ \boldsymbol{x}; \boldsymbol{\theta})}{\partial \boldsymbol{\theta}} = \frac{\partial \mathcal{Q}(\boldsymbol{\theta}, \boldsymbol{\theta}^s)}{\partial \boldsymbol{\theta}}.
\end{eqnarray}

The joint distribution can be rewritten as
\begin{equation}
\log  p(\boldsymbol{z},\boldsymbol{y}\ |\ \boldsymbol{x};\boldsymbol{\theta}) = \sum_{j = 1}^{J} \log p (y_j\ |\ \boldsymbol{y}_1^{j-1}, \boldsymbol{x}_1^{z_j}; \boldsymbol{\theta}) + \sum_{j = 1}^{J} \log p (z_j\ |\ z_{j - 1}).
\end{equation}
We can then do the following expansion:
\begin{equation}
\begin{split}
\mathcal{Q}(\boldsymbol{\theta}, \boldsymbol{\theta}^s) = & 
\sum_{\boldsymbol{z}} \sum_{j = 1}^{J} p(\boldsymbol{z}\ |\ \boldsymbol{y}, \boldsymbol{x}; \boldsymbol{\theta}^s)  \log p (y_j\ |\ \boldsymbol{y}_1^{j-1}, \boldsymbol{x}_1^{z_j}; \boldsymbol{\theta}) + \\
&\qquad \qquad \qquad \qquad  \sum_{\boldsymbol{z}}  \sum_{j = 1}^{J}  p(\boldsymbol{z}\ |\ \boldsymbol{y},\boldsymbol{x};\boldsymbol{\theta}^s) \log p (z_j\ |\ z_{j - 1}).
\end{split}
\end{equation}
We define
\begin{align}
\mathcal{Q}(\boldsymbol{\theta}, \boldsymbol{\theta}^s)_1 & = \sum_{\boldsymbol{z}} \sum_{j = 1}^{J} p(\boldsymbol{z}\ |\ \boldsymbol{y},\boldsymbol{x};\boldsymbol{\theta}^s)  \log p (y_j\ |\ \boldsymbol{y}_1^{j-1}, \boldsymbol{x}_1^{z_j}; \boldsymbol{\theta}), \\
\mathcal{Q}(\boldsymbol{\theta}, \boldsymbol{\theta}^s)_2 & =  
\sum_{\boldsymbol{z}}  \sum_{j = 1}^{J}  p(\boldsymbol{z}\ |\ \boldsymbol{y},\boldsymbol{x};\boldsymbol{\theta}^s) \log p (z_j\ |\ z_{j - 1}),
\end{align}
such that 
\begin{equation}
\frac{\partial \mathcal{Q}(\boldsymbol{\theta}, \boldsymbol{\theta}^s)}{\partial \boldsymbol{\theta}} = \frac{\partial \mathcal{Q}(\boldsymbol{\theta}, \boldsymbol{\theta}^s)_1}{\partial \boldsymbol{\theta}} + \frac{\partial \mathcal{Q}(\boldsymbol{\theta}, \boldsymbol{\theta}^s)_2}{\partial \boldsymbol{\theta}}.
\end{equation}

To simplify $\mathcal{Q}(\boldsymbol{\theta}, \boldsymbol{\theta}^s)_1$, we first introduce a new index $i$,
\begin{align}
\mathcal{Q}(\boldsymbol{\theta}, \boldsymbol{\theta}^s)_1 &=  \sum_{j = 1}^{J} \sum_{i = 1}^{I} \sum_{z_j = i; \boldsymbol{z}^{\backslash j}} p(\boldsymbol{z}\ |\ \boldsymbol{y},\boldsymbol{x};\boldsymbol{\theta}^s)  \log p (y_j\ |\ \boldsymbol{y}_1^{j-1}, \boldsymbol{x}_1^{z_j}; \boldsymbol{\theta}) \\
& =  \sum_{j = 1}^{J} \sum_{i = 1}^{I} \log p(y_j\ |\ \boldsymbol{y}_1^{j-1}, \boldsymbol{x}_1^{i}; \boldsymbol{\theta}) p(z_j = i\ |\ \boldsymbol{y},\boldsymbol{x};\boldsymbol{\theta}^s).
\end{align}
We can then rewrite $\mathcal{Q}(\boldsymbol{\theta}, \boldsymbol{\theta}^s)_1 $ using $\alpha$ and $\beta$:
\begin{align}
\mathcal{Q}(\boldsymbol{\theta}, \boldsymbol{\theta}^s)_1 
& =  \sum_{j = 1}^{J} \sum_{i = 1}^{I} \log p (y_j \ |\ \boldsymbol{y}_1^{j-1}, \boldsymbol{x}_1^{i}; \boldsymbol{\theta}) \frac{p(z_j = i, \boldsymbol{y}\ |\ \boldsymbol{x};\boldsymbol{\theta}^s)}{p (\boldsymbol{y}\ |\ \boldsymbol{x}; \boldsymbol{\theta}^s)}  \\
& = \frac{1}{\alpha(I, J)} \sum_{j = 1}^{J} \sum_{i = 1}^{I} \alpha(i, j) \beta(i, j) \log p (y_j\ |\ \boldsymbol{y}_1^{j-1}, \boldsymbol{x}_1^{i}; \boldsymbol{\theta}) .
\end{align}
The derivative of $\mathcal{Q}(\boldsymbol{\theta}, \boldsymbol{\theta}^s)_1$ is then:
\begin{equation}
\frac{\partial \mathcal{Q}(\boldsymbol{\theta}, \boldsymbol{\theta}^s)_1}{\partial \boldsymbol{\theta}} = \frac{1}{\alpha(I, J)} \sum_{j = 1}^{J} \sum_{i = 1}^{I}  \alpha(i, j) \beta(i, j)
\frac{\frac{\partial }{\partial \boldsymbol{\theta}} p (y_j\ |\ \boldsymbol{y}_1^{j-1}, \boldsymbol{x}_1^{i}; \boldsymbol{\theta})}{p (y_j\ |\ \boldsymbol{y}_1^{j-1}, \boldsymbol{x}_1^{i}; \boldsymbol{\theta})}. 
\end{equation}

To simplify $\mathcal{Q}(\boldsymbol{\theta}, \boldsymbol{\theta}^s)_2$, we introduce two extra indices $i$ and $k$,
\begin{align}
\mathcal{Q}(\boldsymbol{\theta}, \boldsymbol{\theta}^s)_2 & =  \sum_{j = 1}^{J} \sum_{i = 1}^{I} \sum_{k = 1}^{i} 
\sum_{z_{j - 1} = k; z_j = i;\boldsymbol{a}^{\backslash}j, j - 1}    p(\boldsymbol{z}\ |\ \boldsymbol{y},\boldsymbol{x};\boldsymbol{\theta}^s) \log p (z_j = i | z_{j - 1} = k) \\
& = \sum_{j = 1}^{J} \sum_{i = 1}^{I} \sum_{k = 1}^{i} p(z_j = i, z_{j-1} = k\ |\ \boldsymbol{y}, \boldsymbol{x} ; \boldsymbol{\theta}^s) \log p (z_j = i\ |\ z_{j - 1} = k) \\
& = \frac{1}{p(\boldsymbol{y}\ |\ \boldsymbol{x}; \boldsymbol{\theta}^s)}\sum_{j = 1}^{J} \sum_{i = 1}^{I} \sum_{k = 1}^{i} p(z_j = i, z_{j-1} = k, \boldsymbol{y}\ |\ \boldsymbol{x} ; \boldsymbol{\theta}^s) \log p (z_j = i | z_{j - 1} = k).
\end{align}
The term $\sum_{j = 1}^{J} \sum_{i = 1}^{I} \sum_{k = 1}^{i} p(z_j = i, z_{j-1} = k, \boldsymbol{y}\ |\ \boldsymbol{x} ; \boldsymbol{\theta}^s)$ represents the probability of predicting the output $\boldsymbol{y}$ for all the alignments that pass points $(k, j - 1)$ and $(i, j)$ given the input sequence $\boldsymbol{x}$. It can be simplified as:
\begin{equation}
\begin{split}
&\sum_{j = 1}^{J} \sum_{i = 1}^{I} \sum_{k = 1}^{i} p(z_j = i, z_{j-1} = k, \boldsymbol{y}\ |\ \boldsymbol{x} ; \boldsymbol{\theta}^s) \\ 
= &\sum_{j = 1}^{J} \sum_{i = 1}^{I} \sum_{k = 1}^{i} \alpha(k, j - 1) \beta(i, j) p(y_j\ |\ \boldsymbol{y}_{j-1}, \boldsymbol{x}_1^i; \boldsymbol{\theta}^s) p (z_j = i\ |\ z_{j - 1} = k).
\end{split}
\end{equation}
Therefore, $\mathcal{Q}(\boldsymbol{\theta}, \boldsymbol{\theta}^s)_2$ can be rewritten using $\alpha$ and $\beta$:
\begin{equation}
\begin{split}
\mathcal{Q}(\boldsymbol{\theta}, \boldsymbol{\theta}^s)_2  = &
\frac{1}{\alpha(I, J)}
\sum_{j = 1}^{J} \sum_{i = 1}^{I} \sum_{k = 1}^{i} \alpha(k, j - 1) \beta(i, j) p(y_j\ |\ \boldsymbol{y}_{j-1}, \boldsymbol{x}_1^i; \boldsymbol{\theta}^s) \cdot \\
& \qquad \qquad \qquad \qquad p (z_j = i\ |\ z_{j - 1} = k)
\log p (z_j = i\ |\ z_{j - 1} = k).
\end{split}
\end{equation}
The derivative of $\mathcal{Q}(\boldsymbol{\theta}, \boldsymbol{\theta}^s)_2$ is then:
\begin{equation}
\begin{split}
\frac{\partial \mathcal{Q}(\boldsymbol{\theta}, \boldsymbol{\theta}^s)_2}{\partial \boldsymbol{\theta}} = &
\frac{1}{\alpha(I, J)}
\sum_{j = 1}^{J} \sum_{i = 1}^{I} \sum_{k = 1}^{i} \alpha(k, j - 1) \beta(i, j) p(y_j\ |\ \boldsymbol{y}_{j-1}, \boldsymbol{x}_1^i; \boldsymbol{\theta}^s) \cdot \\
& \qquad \qquad \qquad \qquad \qquad \qquad \qquad \frac{\partial}{\partial \boldsymbol{\theta}} p (z_j = i\ |\ z_{j - 1} = k).
\end{split}
\end{equation}
The gradient of the objective function is
\begin{equation}
\begin{split}
\frac{\partial \mathcal{L}(\boldsymbol{\theta})}{\partial \boldsymbol{\theta} } = & - \sum_{(\boldsymbol{x}, \boldsymbol{y}) \in S} \frac{\partial \mathcal{Q}(\boldsymbol{\theta}, \boldsymbol{\theta}^s)_1}{\partial \boldsymbol{\theta}} + \frac{\partial \mathcal{Q}(\boldsymbol{\theta}, \boldsymbol{\theta}^s)_2}{\partial \boldsymbol{\theta}}.
\end{split}
\end{equation}
In the case that the transition probabilities are modelled by geometric distribution, $ \frac{\partial \mathcal{Q}(\boldsymbol{\theta}, \boldsymbol{\theta}^s)_2}{\partial \boldsymbol{\theta}} = 0$.

\section{Loss Gradient: Direct Derivation}
\label{sec:dr_drt}
We also derive the gradient directly. Let $\boldsymbol{\theta}_j$ denotes the neural network parameters w.r.t. the output distribution $p(y_j\ |\ \boldsymbol{y}_1^{j-1}, \boldsymbol{x})$ at position $j$. Using the chain rule, we expand the derivative as follows:
\begin{equation}
\begin{split}
\frac{\partial \log p(\boldsymbol{y}\ |\ \boldsymbol{x}; \boldsymbol{\theta})}{\partial \boldsymbol{\theta}} & = \sum_{j=1}^J \frac{\partial \log p(\boldsymbol{y}\ |\ \boldsymbol{x}; \boldsymbol{\theta})}{\partial \boldsymbol{\theta}_j} \\
& = \sum_{j=1}^J \sum_{i=1}^I \frac{\partial \log p(\boldsymbol{y}\ |\ \boldsymbol{x}; \boldsymbol{\theta})}{\partial \alpha(i, j)} \frac{\partial \alpha(i, j)}{\partial \boldsymbol{\theta}_j}.
\end{split}
\end{equation}
The derivative w.r.t. the forward weights is computed as:
\begin{equation}
\begin{split}
\frac{\partial \log p(\boldsymbol{y}\ |\ \boldsymbol{x}; \boldsymbol{\theta})}{\partial \alpha(i, j)} & = \frac{1}{p(\boldsymbol{y}\ |\ \boldsymbol{x}; \boldsymbol{\theta})} \frac{\partial}{\partial \alpha(i, j)}(p(\boldsymbol{y}\ |\ \boldsymbol{x}; \boldsymbol{\theta})) \\
& = \frac{1}{p(\boldsymbol{y}\ |\ \boldsymbol{x}; \boldsymbol{\theta})} \frac{\partial}{\partial \alpha(i, j)}\left( \sum_i \alpha(i, j)\beta(i,j) \right) \\
& = \frac{1}{p(\boldsymbol{y}\ |\ \boldsymbol{x};\boldsymbol{\theta})} \beta(i, j).
\end{split}
\end{equation}
By expanding $\alpha(i, j)$ based on its definition, we can obtain the derivative of $\alpha(i, j)$ w.r.t. $\boldsymbol{\theta}_j$,
\begin{equation}
\begin{split}
\frac{\partial \alpha(i, j)}{\partial \boldsymbol{\theta}_j} &=  \frac{\partial}{\partial \boldsymbol{\theta}_j} \left( p(y_j\ |\ \boldsymbol{x}_1^i, \boldsymbol{y}_1^{j-1}; \boldsymbol{\theta})
\left( \sum_{k=1}^i \alpha(k, j - 1) p (z_j = i\ |\ z_{j-1} = k) \right)
\right)  \\
& = \frac{\partial}{\partial \boldsymbol{\theta}_j} \left( p(y_j\ |\ \boldsymbol{x}_1^i, \boldsymbol{y}_1^{j-1}; \boldsymbol{\theta}) \right) \left( \sum_{k=1}^i \alpha(k, j - 1) p (z_j = i\ |\ z_{j-1} = k) \right) + \\
& \qquad \quad p(y_j\ |\ \boldsymbol{x}_1^i, \boldsymbol{y}_1^{j-1}; \boldsymbol{\theta}) \frac{\partial}{\partial \boldsymbol{\theta}_j}  \left( \sum_{k=1}^i \alpha(k, j - 1) p (z_j = i\ |\ z_{j-1} = k) \right).
\end{split}
\end{equation}
We first simplify the first term:
\begin{equation}
\begin{split}
& \ \ \frac{\partial}{\partial \boldsymbol{\theta}_j} \left( p(y_j\ |\ \boldsymbol{x}_1^i, \boldsymbol{y}_1^{j-1}; \boldsymbol{\theta}) \right) \left( \sum_{k=1}^i \alpha(k, j - 1) p (z_j = i\ |\ z_{j-1} = k) \right) \\
= &\ \  \frac{\alpha(i, j)}{p(y_j\ |\  \boldsymbol{y}_1^{j-1}, \boldsymbol{x}_1^i; \boldsymbol{\theta})} \frac{\partial}{\partial \boldsymbol{\theta}_j} p(y_j\ |\ \boldsymbol{y}_1^{j-1}, \boldsymbol{x}_1^i; \boldsymbol{\theta}) \\
= &\ \  \frac{\alpha(i, j)}{p(y_j\ |\  \boldsymbol{y}_1^{j-1}, \boldsymbol{x}_1^i; \boldsymbol{\theta})} \frac{\partial}{\partial \boldsymbol{\theta}} p(y_j\ |\ \boldsymbol{y}_1^{j-1}, \boldsymbol{x}_1^i; \boldsymbol{\theta}).
\end{split}
\end{equation}
For the second term, we have $\frac{\partial}{\partial \boldsymbol{\theta}_j} \alpha(k, j-1) = 0$, due to our definition of $\boldsymbol{\theta}_j$ and the fact that the $\alpha$ values are computed going forwards. Therefore, we have:
\begin{equation}
\begin{split}
& p(y_j\ |\ \boldsymbol{x}_1^i, \boldsymbol{y}_1^{j-1}; \boldsymbol{\theta}) \frac{\partial}{\partial \boldsymbol{\theta}_j}  \left( \sum_{k=1}^i \alpha(k, j - 1) p (z_j = i\ |\ z_{j-1} = k) \right) \\
=\ \ & p(y_j\ |\ \boldsymbol{x}_1^i, \boldsymbol{y}_1^{j-1}; \boldsymbol{\theta}) \sum_{k =1}^i p (z_j = i\ |\ z_{j-1} = k) \frac{\partial}{\partial \boldsymbol{\theta}_j}   \alpha(k, j-1) + \\
& \qquad \qquad p(y_j\ |\ \boldsymbol{x}_1^i, \boldsymbol{y}_1^{j-1}; \boldsymbol{\theta}) \sum_{k =1}^i 
\alpha(k, j-1) \frac{\partial}{\partial \boldsymbol{\theta}_j} p (z_j = i\ |\ z_{j-1} = k)   \\
= \ \ & p(y_j\ |\ \boldsymbol{x}_1^i, \boldsymbol{y}_1^{j-1}; \boldsymbol{\theta}) \sum_{k =1}^i 
\alpha(k, j-1) \frac{\partial}{\partial \boldsymbol{\theta}} p (z_j = i\ |\ z_{j-1} = k). 
\end{split}
\end{equation}
By substituting the derivatives of the two terms back, we have the derivative of $\log p(\boldsymbol{y}\ |\ \boldsymbol{x}; \boldsymbol{\theta})$:
\begin{equation}
\begin{split}
\frac{\partial \log p(\boldsymbol{y}\ |\ \boldsymbol{x}; \boldsymbol{\theta})}{\partial \boldsymbol{\theta}}
& =  \sum_{j=1}^J \sum_{i=1}^I \frac{\partial \log p(\boldsymbol{y}\ |\ \boldsymbol{x}; \boldsymbol{\theta})}{\partial \alpha(i, j)} \frac{\partial \alpha(i, j)}{\partial \boldsymbol{\theta}_j} \\
& =  \frac{1}{\alpha(I, J)} \sum_{j=1}^{J} \sum_{i=1}^{I} \alpha(i, j) \beta(i, j)
\frac{\frac{\partial}{\partial \boldsymbol{\theta}} p(y_j\ |\ \boldsymbol{y}_1^{j-1}, \boldsymbol{x}_1^i; \boldsymbol{\theta})}{p(y_j\ |\ \boldsymbol{y}_1^{j-1}, \boldsymbol{x}_1^i, \boldsymbol{\theta})}  + \\
& \quad \ \frac{1}{\alpha(I, J)}
\sum_{j = 1}^{J} \sum_{i = 1}^{I} \sum_{k = 1}^{i} \alpha(k, j - 1) \beta(i, j) p(y_j\ |\ \boldsymbol{y}_1^{j-1}, \boldsymbol{x}_1^i; \boldsymbol{\theta}) \cdot \\
& \qquad \qquad \qquad \qquad \qquad \qquad \qquad \qquad \frac{\partial}{\partial \boldsymbol{\theta}} p (z_j = i\ |\ z_{j - 1} = k).
\end{split}
\end{equation}
This result the same as the gradient computed using the indirect approach.